\documentclass[10pt,journal,compsoc]{IEEEtran}

\usepackage{epsfig}
\usepackage{subfigure}
\usepackage{makecell}
\usepackage{diagbox}
\usepackage{siunitx}

\usepackage{booktabs}
\usepackage{enumitem}
\usepackage{multirow}
\usepackage{algorithm,algorithmicx,algpseudocode}
\usepackage{hyperref}
\usepackage{graphicx}
\usepackage{amsmath}
\usepackage{float}
\usepackage{nicefrac}
\usepackage{mathrsfs}
\usepackage{xcolor}
\usepackage{colortbl}
\usepackage{marvosym}
\usepackage[export]{adjustbox}
\usepackage{pifont}
\usepackage{cleveref}
\usepackage{amssymb}
\usepackage{ifpdf}
\ifpdf
  
\else
\fi
\usepackage{color}
\definecolor{cyan}{cmyk}{1,0,0,0}
\definecolor{darkgreen}{rgb}{0,0.5,0}
\definecolor{orange}{rgb}{1,0.5,0}
\definecolor{magenta}{cmyk}{0,1,0,0}
\definecolor{darkyellow}{cmyk}{0,0,0.75,0}
\definecolor{gray}{rgb}{0.8,0.8,0.8}
\expandafter\let\csname Cross\endcsname\relax
\usepackage{bbding}

\usepackage{amsmath,amsfonts}
\usepackage{array}
\usepackage{textcomp}
\usepackage{subfigure}
\usepackage{url}
\usepackage{verbatim}
\ifCLASSOPTIONcompsoc
\usepackage[nocompress]{cite}
\else
\usepackage{cite}
\fi

\ifCLASSINFOpdf
\else
\fi

\def\etal{{\em et al.}}
\crefname{equation}{Eq.}{Eqs.}
\begin{document}

% \title{Stimulating the Diffusion Model for Image Denoising via Adaptive Embedding and Ensembling}
\title{Stimulating Diffusion Model for Image Denoising via Adaptive Embedding and Ensembling}
\author{
Tong Li, 
Hansen Feng, 
Lizhi Wang,~\IEEEmembership{Member,~IEEE}, 
Lin Zhu,~\IEEEmembership{Member,~IEEE},\\ 
Zhiwei Xiong,~\IEEEmembership{Member,~IEEE} and Hua Huang,~\IEEEmembership{Senior Member,~IEEE}

\IEEEcompsocitemizethanks{
\IEEEcompsocthanksitem Tong Li, Hansen Feng and Lin Zhu are with the School of Computer Science and Technology, Beijing Institute of Technology, Beijing, 100081, China. Email: $\{$litong, fenghansen, zhulin$\}$@bit.edu.cn\protect

\IEEEcompsocthanksitem Zhiwei Xiong is with the Department of Electronic Engineering and Information Science, University of Science and Technology of China, Hefei, 230027, China. Email: zwxiong@ustc.edu.cn\protect

\IEEEcompsocthanksitem Lizhi Wang and Hua Huang are with the School of Artificial Intelligence, Beijing Normal University, Beijing, 100875, China. Email: wanglizhi@ustc.edu.cn; huahuang@bnu.edu.cn\protect

\IEEEcompsocthanksitem Corresponding author: Lizhi Wang\protect

\IEEEcompsocthanksitem Tong Li and
Hansen Feng contributed equally to this work\protect
% \IEEEcompsocthanksitem This work is supported by National Natural Science Foundation of China (62322204, 62072038, 62131003).\protect
% \IEEEcompsocthanksitem Part of this work is done during the internship of Hansen Feng at Megvii Technology.
}% <-this % stops an unwanted space

}

% The paper headers
\markboth{Journal of \LaTeX\ Class Files,~Vol.~14, No.~8, August~2015}%
	{Shell \MakeLowercase{\textit{et al.}}: Bare Demo of IEEEtran.cls for Computer Society Journals}

\IEEEtitleabstractindextext{
\begin{abstract}
		Image denoising is a fundamental problem in computational photography, where achieving high perception with low distortion is highly demanding. Current methods either struggle with perceptual quality or suffer from significant distortion. Recently, the emerging diffusion model has achieved state-of-the-art performance in various tasks and demonstrates great potential for image denoising. However, stimulating diffusion models for image denoising is not straightforward and requires solving several critical problems. For one thing, the input inconsistency hinders the connection between diffusion models and image denoising. For another, the content inconsistency between the generated image and the desired denoised image introduces distortion. To tackle these problems, we present a novel strategy called the Diffusion Model for Image Denoising (DMID) by understanding and rethinking the diffusion model from a denoising perspective. Our DMID strategy includes an adaptive embedding method that embeds the noisy image into a pre-trained unconditional diffusion model and an adaptive ensembling method that reduces distortion in the denoised image. Our DMID strategy achieves state-of-the-art performance on both distortion-based and perception-based metrics, for both Gaussian and real-world image denoising.
        The code is available at \href{https://github.com/Li-Tong-621/DMID}{https://github.com/Li-Tong-621/DMID}.
	\end{abstract}
	
	%————————————————————————————————————————————————————————
}
\maketitle
\IEEEdisplaynontitleabstractindextext
\IEEEpeerreviewmaketitle
%———————————————————————————————————————————————————————

	%\setcitestyle{numbers,sort&compress}

	%\keywords{image denoising, diffusion model, iterative denoising, distortion-perception}

 \IEEEraisesectionheading{
 \section{Introduction}\label{sec:Introduction}
 }

	% \IEEEPARstart{A}{s} smartphones have become ubiquitous, capturing high-quality images has become increasingly demanding. However, capturing images under extreme conditions, such as low-light environments, can lead to significant information loss due to noise. As a result, image denoising with low distortion and high perception has become a crucial and active research area.

    \IEEEPARstart{A}{s} smartphones have become ubiquitous, the pursuit of capturing high-quality images has become notably more demanding. Yet, when capturing images under challenging conditions, such as low-light environments, substantial information would be lost due to imaging noise. Consequently, the field of image denoising, with a focus on achieving low distortion and high perception, has been as a vital and thriving area of research.
	
	Traditional methods~\cite{KSVD,lowrank,smooth1} rely on image priors to guide the denoising process, but their effectiveness is limited under extreme conditions. With the development of deep learning, discriminative methods have become the mainstream method, which are usually trained by pixel-level losses~\cite{liang2021swinir,zamir2022restormer,2021Plug,2016Beyond,2017FFDNet}. Actually,
    the pixel-level losses tend to predict the median (or average) of all possible values rather than the realistic images~\cite{mathieu2016deep,Ledig_2017_CVPR}. Thus, these discriminative methods often struggle with perceptual quality, particularly under extreme noise levels.
	
	To improve perceptual quality, other image restoration tasks usually employ generative methods. Current state-of-the-art solutions usually rely on Generative Adversarial Networks (GANs)~\cite{2014Generative,Ledig_2017_CVPR,kupyn2018deblurgan}. However, little work has been done to improve the perceptual quality of image denoising, even though generative approaches have been introduced to address real noise modeling~\cite{maleky2022noise2noiseflow,jang2021c2n}. Moreover, GAN-based methods are widely blamed for artifacts and inconsistency, leading to significant distortion, especially for extreme degradation. 
	\begin{figure}[!t]
%\vspace{-1em}
\centering
 \includegraphics[width=.49\textwidth]{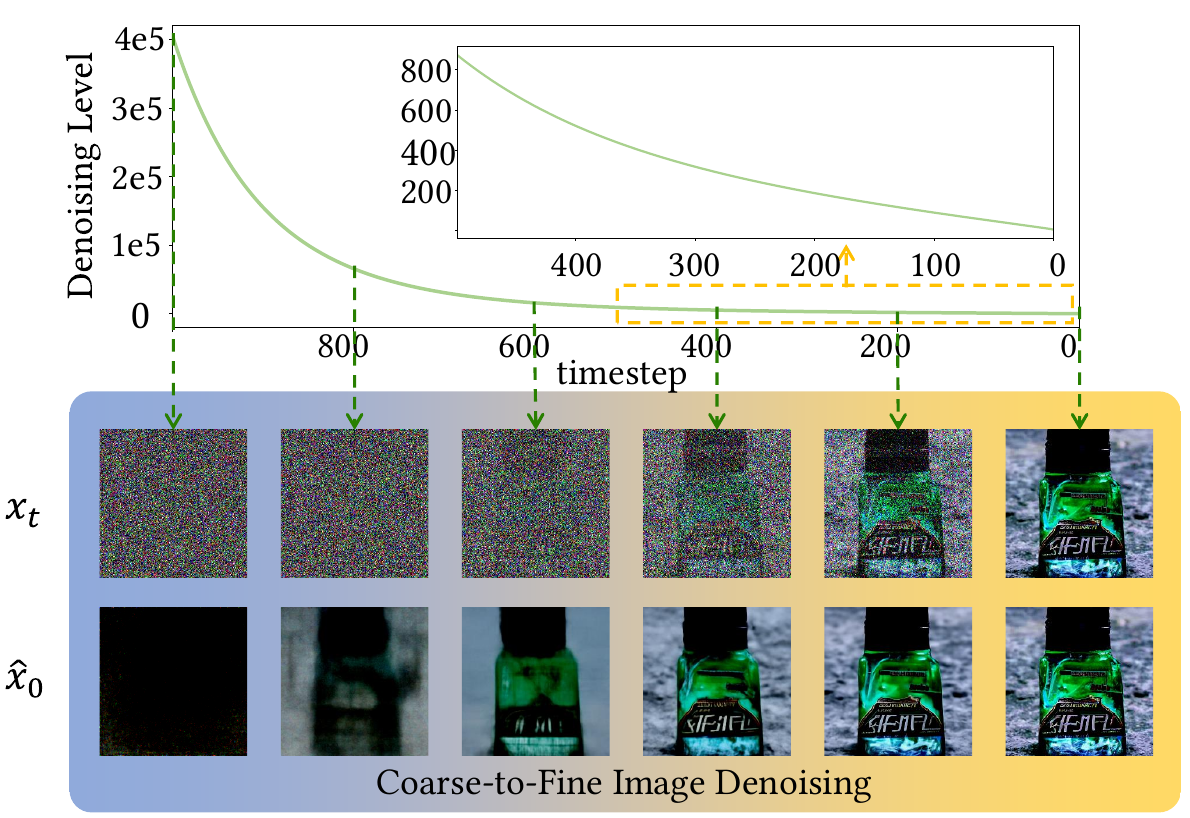}
 \caption{
 % Reverse process of a diffusion model. It can be viewed as a coarse-to-fine iterative denoising process.
 From the denoising perspective, the reverse process of a diffusion model can be viewed as a coarse-to-fine iterative denoising process. 
 {
 % \color{red} 
 The noise level corresponds to the standard deviation of noise on an 8-bit image, where the maximum signal for a clean image is 255.}
 }
 
 \label{fig:diffusion model-iterative denoising}
 % \vspace{-1em}
\end{figure}

	Recently, the diffusion model~\cite{ho2020denoising,sohl2015deep,song2020score} has achieved state-of-the-art (SOTA) performance in various tasks, such as image super-resolution~\cite{li2022srdiff,saharia2022image}, image inpainting~\cite{lugmayr2022repaint,dpnpr}, image deblurring~\cite{whang2022deblurring}. The methods~\cite{li2022srdiff,lugmayr2022repaint,dpnpr,whang2022deblurring} that utilize diffusion models exhibit higher perceptual quality and fewer artifacts. Therefore, we believe that the diffusion model holds significant potential for image denoising. 
    However, stimulating the diffusion model for image denoising still remains several critical problems. Specifically, the diffusion model is designed to receive 
    standard Gaussian noise as input, which is different from the noisy image input required for image denoising. 
	Furthermore, the content difference between the generated image and the desired denoised image introduces distortion. 
	The input inconsistency problem and the content inconsistency problem are crucial to be tackled to stimulate the diffusion model for image denoising.
	
	\begin{figure}[t]
%\vspace{1em}
 \includegraphics[width=0.48\textwidth]{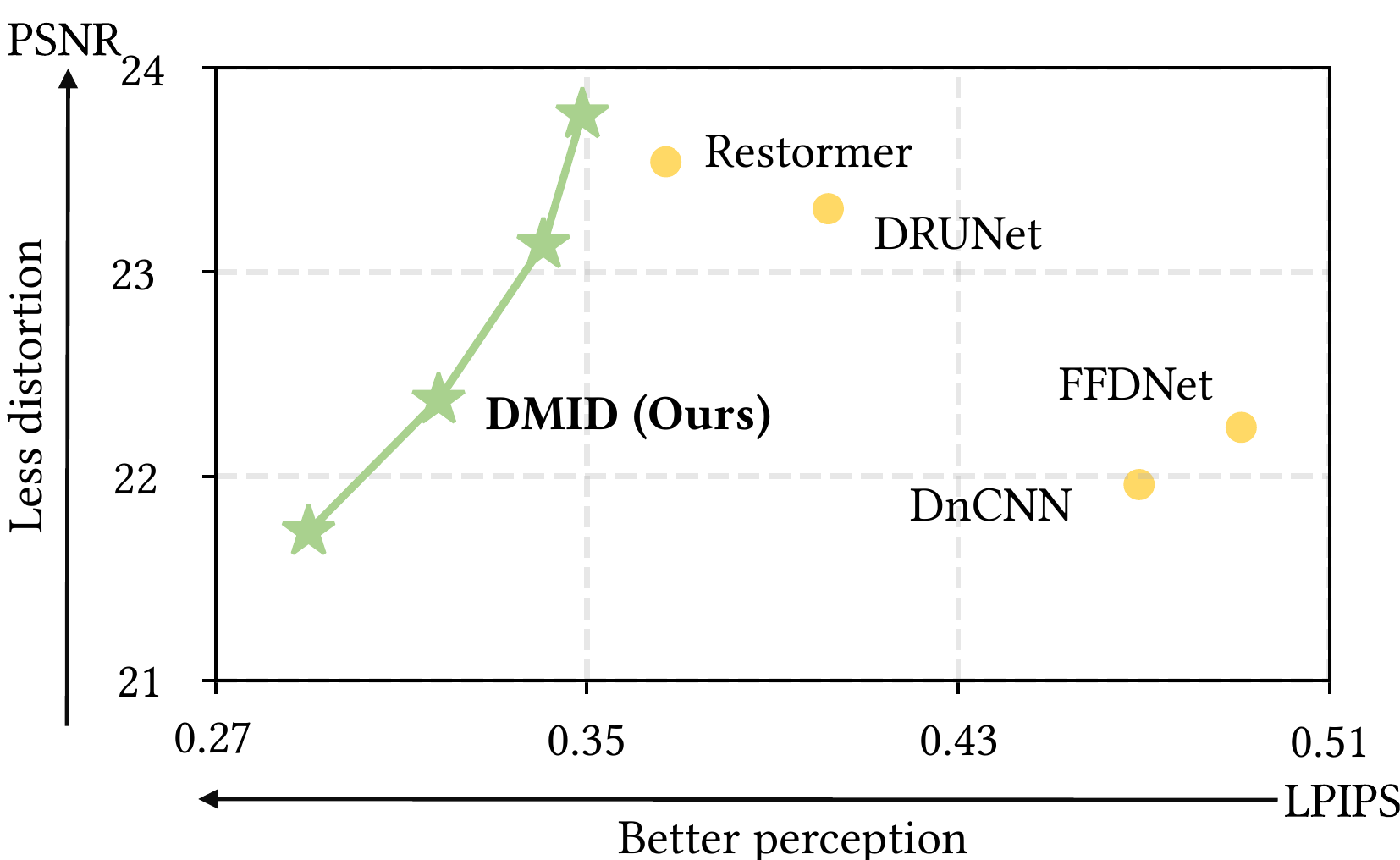}
 \caption{Perception-distortion trade-off of different methods. Our method traverses through the perception-distortion curve and achieves SOTA performance.}
 \label{fig:perception-distortion curve}
% \vspace{-1em}
\end{figure}

	In this paper, we propose a novel strategy to stimulate the Diffusion Model for Image Denoising~(DMID), consisting of an adaptive embedding method and an adaptive ensembling method. Our insight comes from the denoising perspective towards the diffusion model.  
    % %%%%%
    From the denoising perspective, the diffusion model and the iterative denoising methods share the same structure and similar functionalities. 
    As illustrated in Figure~\ref{fig:diffusion model-iterative denoising}, the diffusion model can be viewed as an iterative denoising method.
    %%%%%
    % From the denoising perspective, the diffusion model and the iterative denoising methods share the same structure. It is widely acknowledged that the functionality is determined by the underlying structure for deep learning methods. 
    % Therefore, the diffusion model and the iterative denoising share similar functionalities.
    % As illustrated in Figure~\ref{fig:diffusion model-iterative denoising}, the diffusion model can also be viewed as an iterative denoising method.
    %
    This new perspective provides us with huge potential to employ a pre-trained unconditional diffusion model to tackle the aforementioned problems, which also eliminates the resource-intensive and time-consuming demands associated with training a diffusion model~\cite{dhariwal2021diffusion}.

	Firstly, 
	%from the denoising perspective, 
	the problem of input inconsistency is essentially an embedding problem. Here the noisy image and iterative denoising process correspond to the intermediate state and the reverse process subsequence of the pre-trained unconditional diffusion model, respectively. Therefore, we propose an adaptive embedding method that embeds the noisy image into an intermediate state of the pre-trained unconditional diffusion model. Our embedding method enables to perform Gaussian and real-world image denoising.
	
	Secondly,
	the problem of content inconsistency fundamentally constitutes a stochasticity constraint problem. Since
	% from the denoising perspective,
	the distortion of the diffusion model mainly comes from the stochasticity inherent in the iterative process. To tackle this problem, we propose an adaptive ensembling method that  
    can adjust distortion and perception without additional training.
    Our ensembling method enables to traverse through the perception-distortion curve~\cite{8578750} as shown in Figure~\ref{fig:perception-distortion curve}, which achieves SOTA performance on both distortion-based and perception-based metrics.
	
	%Secondly,
	%% from the denoising perspective,
	%the distortion of the diffusion model mainly comes from the stochasticity inherent in the iterative process. Therefore, the challenge of content inconsistency fundamentally constitutes a stochasticity constraint problem. To tackle this problem, we propose an adaptive ensembling method that does not require additional training. Our ensembling method enables us to traverse through the perception-distortion curve~\cite{8578750} as shown in Figure ~\ref{fig:perception-distortion curve}, which achieves SOTA performance on both distortion-based and perception-based metrics.

	In summary, our contributions are as follows:
	
	\begin{itemize}%[leftmargin=2.5em]
		
		\item From the denoising perspective, we contribute a novel understanding and rethinking of the diffusion model and stimulate the diffusion model for image denoising.
		
		\item We propose an adaptive embedding method that embeds the noisy image into a pre-trained unconditional diffusion model, which enables us to perform Gaussian and real-world image denoising.
		
		\item We propose an adaptive ensembling method that constrains the distortion of the diffusion model, which enables us to traverse through the perception-distortion curve.
		
		\item Our method achieves SOTA performance on both distortion-based and perception-based metrics, and our advantages increase as the noise level becomes larger. 
		%Specifically, we achieve an advantage of over 0.5dB in PSNR and an over $50 \%$ improvement in FID under extreme conditions compared to Restormer~\cite{zamir2022restormer}.

	\end{itemize}   
	
	The organization of this paper is as follows. 
    Firstly, we provide a brief introduction to the development of image denoising and the diffusion model in Section \ref{sec:RelatedWorks}. 
    Next, we offer a rethinking and understanding of the diffusion model from the denoising perspective in Section \ref{sec:rethinking and understanding}. 
    Then, we propose our DMID strategy to tackle the existing problems in Section \ref{sec:strategy}. 
    After that, we demonstrate the performance and evaluate the effectiveness of our strategy in Section \ref{sec:experiments}. 
    % Finally, we conclude this paper and point out the future directions in Section \ref{sec:conclusion}.
    Finally, we discuss the advantages and disadvantages in Section \ref{sec:final_discussion} and point out future directions in Section \ref{sec:conclusion}.
	%%%%%%%%%%%%%%%%%%%%%%%%%%%%%%%%%%%%%%%%%%%%%%%%%%%%%%%%%%%%%%%%%%%%%%%%%%%%%%%%%
	\section{Related Works}\label{sec:RelatedWorks}
	In this section, we briefly introduce the development of image denoising and the diffusion model.
	
	\subsection{Image Denoising}
	Image denoising is an essential task in computer vision. It aims at restoring a clean image from its noisy counterpart. %This is a fundamental problem in low-level computer vision. %This is a fundamental and challenging problem in low-level computer vision.%, as the ill-posed nature of the problem is proportional to the noise level. %We divide image denoising methods into distortion-based methods and perception-based methods.  
	However, achieving high perception quality and low distortion in image denoising remains a challenging task. To tackle this problem, image denoising methods can be broadly divided into two categories: distortion-based methods and perception-based methods.
	
	Distortion-based methods consist of traditional denoising methods and discriminative methods. Traditional image denoising methods focus on the usage of image priors such as sparsity~\cite{KSVD,elad2006image}, low rank~\cite{lowrank}, self-similarity~\cite{2005A,2007Image}, and smoothness~\cite{smooth1,smooth2}. However, the ability of these priors in extreme conditions is ultimately limited, making it difficult to adequately denoise the corrupted images. With the development of deep learning, discriminative models~\cite{liang2021swinir,zamir2022restormer,wang2022uformer, chen2021pre, 2021Plug,2016Beyond,2017FFDNet,cheng2021nbnet,zamir2021multi}, have become the mainstream methods. The main idea of discriminative methods is to use neural networks to learn mappings from noisy images to clean images through paired noisy-clean data or just noisy data~\cite{noise2void,noise2self,lee2022apbsn,wang2023lgbpn,Recorrupted2Recorrupted}. $L_1$ loss and $L_2$ loss are most widely adopted during training. However, it is well known that these pixel-level losses tend to predict the median (or average) of all possible values rather than the realistic images~\cite{mathieu2016deep,Ledig_2017_CVPR}, resulting in over-smooth images with poor perceptual quality.

	In addition, some distortion-based methods~\cite{osher2005iterative,dong2011sparsity,8481558,KSVD} employ an iterative framework to optimize denoising performance. The iterative denoising methods usually predict a rough image and refine it through multiple iterations. The iterative denoising framework serves as a regularization and constraint. In addition, the iterative denoising framework decomposes image denoising into a series of sub-problems, making denoising easier. However, these methods still struggle with perceptual quality.
	
	In contrast to distortion-based methods, perception-based methods typically exhibit superior perceptual quality but have significant distortion. Perception-based methods usually refer to generative methods. Current perception-based approaches with competitive performance for other image restoration usually rely on Generative Adversarial Networks (GAN)~\cite{2014Generative,Ledig_2017_CVPR,kupyn2018deblurgan}. However, GAN-based methods often introduce artifacts and inconsistent details that are not present in the original clean image. The artifacts usually lead to significant distortion, especially for extreme degradation.
	%, such as $\times$16 super-resolution. 
	As for image denoising, many methods employ GAN and flow~\cite{rezende2015variational} to solve real noise modeling problems~\cite{maleky2022noise2noiseflow,marras2020reconstructing,jang2021c2n,abdelhamed2019noise,chang2020learning,pngan,2019Real} to get paired data easily. It is really a shame that little work has attempted to improve the perceptual quality of image denoising. 
	
	\subsection{Diffusion Model}
	\label{sec:related work diffusion model}
	The diffusion model is a burgeoning likelihood-based generative model and has demonstrated remarkable success over other models in various tasks~\cite{dhariwal2021diffusion,ho2020denoising,rombach2022high}. 
	
     The diffusion model is originally proposed separately from diffusion-based~\cite{sohl2015deep} and score-matching-based~\cite{song2019generative} perspectives. Later 
     Ho~\etal~\cite{ho2020denoising} demonstrate the enormous potential of unconditional diffusion model for image synthesis.
     % Recently, Dhariwal~\etal~\cite{dhariwal2021diffusion} improves unconditional image synthesis quality by finding a better architecture and firstly proposes a kind of conditional diffusion model. This work successfully brings diffusion models back to the public view.
     Recently Dhariwal~\etal propose a kind of conditional diffusion model and bring diffusion models back to the public view.
	After that, conditional diffusion models adapt to various tasks and have achieved brilliant performances in a series of image restoration tasks, such as image super-resolution~\cite{li2022srdiff,saharia2022image}, image inpainting~\cite{lugmayr2022repaint,dpnpr}, image deblurring~\cite{whang2022deblurring}. 
    Some methods also exhibit the capability to address multiple tasks.
    For instance, the novel method~\cite{kulikov2023sinddm} proposes a framework for training a diffusion model on a single image and is applicable in various tasks, including style transfer and harmonization. 
    Another advanced method~\cite{dpnpr} exhibits proficiency in tasks such as super-resolution, deblurring, and inpainting.

	Recently, some methods~\cite{kawar2022denoising,wang2022zero,dpnpr} have exclusively employed pre-trained unconditional diffusion models, originally trained for image synthesis, to address linear image restoration problems. These methods typically assume that the observed image $y$ is degraded as $y = Hx + n$, where $H$ represents the degradation matrix, $n$ denotes noise, and $x$ signifies the desired clean image. 
    %In the context of these methods~\cite{kawar2022denoising,wang2022zero}, during each sampling step of the diffusion model, certain information is replaced while preserving other information by decomposing the degradation matrix $H$, often using techniques like Singular Value Decomposition, Range-Null Space Decomposition. 
    The existing methods usually replace some content with the original degraded image and preserve the others during each sampling step of the diffusion model. These methods achieve this by decomposing the degradation matrix $H$ to identify the boundary of the reserved information area. The popular techniques are Singular Value Decomposition and Range-Null Space Decomposition. 
    
     However, 
     it should be emphasized
     that all the existing methods that employ pre-trained unconditional diffusion models to address restoration problems have been centered around image super-resolution or deblurring tasks, overlooking the specific task of image denoising. 
    %This trend can be attributed to the heavy reliance of current methods on the decomposition of the degradation matrix $H$. However, in the case of image denoising, the degradation matrix $H$ is simply an identity matrix, which renders matrix decomposition ineffective in providing additional valuable information to identify the boundary of the reserved information area. Therefore, the adaptation of pre-trained unconditional diffusion models, originally trained for image synthesis, for the purpose of image denoising requires further refinement that aligns with the unique demands of denoising. From the denoising perspective, we conduct a comparative analysis between the diffusion model and iterative denoising methods, highlighting the two inconsistency (input inconsistency and content inconsistency) problems and proposing solutions to address these issues.
    Applying the existing diffusion-based image restoration methods to image denoising still encounters several issues.
    Firstly, existing methods fall short in the input inconsistency. Existing methods perform restoration by decomposing the degradation matrix $H$ to identify the boundary of the reserved information area. Therefore these methods circumvent the input inconsistency for other restoration tasks. However, when it comes to image denoising, the degradation matrix $H$ is simply an identity matrix, which renders matrix decomposition ineffective in providing additional valuable information to identify the boundary of the reserved information area. Therefore such decomposition-based methods fall short in the input inconsistency, making it ill-suited for image denoising.  
    Secondly, existing methods overlook the content inconsistency. These methods often blindly pursue high perceptual quality, leading to a divergence between the denoised image and the desired clean image. Moreover, existing methods neglect the factors that impact distortion-based and perception-based performance, thus failing to meet the high requirements for both distortion-based and perception-based quality in image denoising.

    More recently, a method~\cite{xie2023diffusion} attempts to customize different diffusion models for each noise type. However, to customize distinct diffusion models, the method has to simplify noise types to standard Gaussian or standard Poisson noise, making it less effective for real-world image denoising.

	%\section{Preliminaries}
	\label{sec:diffusion model}

	%%%%%%%%%%%%%%%%%%%%%%%%%%%%%%%%%%%%%%%%%%%%%%%%%%%%%%%%%%
	
	\section{Rethinking and Understanding}
	\label{sec:rethinking and understanding}
	In this section, we first introduce the pipeline of the diffusion model in Section \ref{sec:Diffusion Models Pipeline}. Next, we present the iterative denoising framework in Section \ref{sec:framework}. 
	%Then, 
	Finally,
	we integrate the diffusion model with the iterative denoising framework in Section \ref{sec:discussion}, providing a new understanding of the diffusion model from the denoising perspective.

	\subsection{Diffusion Model Pipeline}
	\label{sec:Diffusion Models Pipeline}
	
	In this section, we give a rough introduction to the diffusion model. 
	
	Diffusion models are composed of a T-timestep forward process and a T-timestep backward process. 
    % The forward process, also known as the diffusion process, gradually adds \textit{i.i.d.} Gaussian noise to the initial data $x_0$ and converges to the standard Gaussian distribution. The reverse process is the inference process, which starts with {
    % % \color{red}
    % standard Gaussian noise $x_T \sim \mathcal{N}(0,1)$ and eventually obtains the expected high-quality image $x_0$.
    The forward process gradually adds Gaussian noise to the initial data $x_0$ and converges to the standard Gaussian distribution. The reverse process starts with standard Gaussian noise $x_T \sim \mathcal{N}(0,1)$ and eventually obtains the expected high-quality image $x_0$.
	
	The forward process is changed from the previous state $x_{t-1}$ follows the Markov chain to generate current state $x_t$:
	\begin{alignat}{2}
		q(x_t | x_{t-1}) & = \mathcal{N}(x_t; \sqrt{1-\beta_t} x_{t-1}, \beta_t \mathbf{I}) \ ,
		\label{eq:singlediffusion}
	\end{alignat}
	where $t$ is an intermediate timestep, $x_t$ represents the data (such as a noisy image) of the state at timestep $t$, $\beta_t$ can be predefined constants as hyperparameters or learned by reparameterization\cite{kingma2013auto}, and $\mathcal{N}$ represents Gaussian distribution. Using the reparameterization technique and following \cref{eq:singlediffusion}, current state $x_t$ can also be expressed as:
	
	\begin{alignat}{2}
		q(x_t | x_0) & = \mathcal{N}(x_t; \sqrt{\bar{\alpha}_t} x_0, (1-\bar{\alpha}_t) \mathbf{I})  \ ,
		\label{eq:wholediffusion}
	\end{alignat}
	where $\alpha_t = 1 - \beta_t$, $\bar{\alpha_t} = \prod_{i=1}^t (1 - \beta_i)$.
	
	% The reverse process generates an image by a series of sampling processes \cref{eq:siglereverse} from $x_t$ to $x_{t-1}$, modeled by the neural network. This process gradually transforms the Gaussian distribution $x_T$ into the expected data distribution $x_0$.
	
	% \begin{equation}
	% 	\label{eq:siglereverse}
	% 	p_{\theta}(x_{t-1}|x_t) = \mathcal{N}(x_{t-1}; \mu_{\theta}(x_t, t), \sigma_{\theta}^2) \ .
	% \end{equation}
    % The reverse process generates an image by a series of sampling processes from $x_t$ to $x_{t-1}$, and we refer the number of sampling times as $S_t$:
    The reverse process generates an image by a series of sampling processes, and we refer to the number of sampling times as $S_t$. For instance, from the initial state $x_{1000}$ to get the intermediate state $x_{500}$ and from the intermediate state $x_{500}$ to obtain the final state $x_0$, the sampling times are $S_t=2$. The sampling process from $x_t$ to $x_{t-1}$ is expressed as:
	
	\begin{equation}
		\label{eq:siglereverse}
		p_{\theta}(x_{t-1}|x_t) = \mathcal{N}(x_{t-1}; \mu_{\theta}(x_t, t), \sigma_{\theta}^2) \ ,
	\end{equation}
    % The sampling process is modeled by the neural network. 
    where $\mu_{\theta}(x_t, t)$ is modeled by the neural network. The reverse process gradually transforms the Gaussian distribution $x_T$ into the expected data distribution $x_0$.

	% Different sampling strategies of \cref{eq:siglereverse} have been proposed from DDPM~\cite{ho2020denoising} to DDIM~\cite{song2020denoising}. These strategies 
    Different sampling strategies~\cite{ho2020denoising,song2020denoising} of \cref{eq:siglereverse} can be summarized as a predicted item and an additional noise item:
	\begin{multline}
		\label{eq:reverse_ddim}
		x_{t-1} =\sqrt{\bar{\alpha}_{t-1}} \underbrace{\left(\frac{x_t - \sqrt{1-\bar{\alpha}_t} \epsilon_{\theta}(x_t, t) }{{\sqrt{\bar{\alpha}_t}}} \right)}_{predict ed \, \hat{x}_0} \\
		+ \underbrace{\sqrt{1-\bar{\alpha}_{t-1}-\sigma_t^2}\epsilon_{\theta}(x_t, t)
			+\sigma_t \epsilon_t}_{additional \, noise} \ ,
	\end{multline}
	where $\sigma_t$ is an arbitrary constant and $\epsilon_{\theta}(x_t, t)$ is modeled by neural network. 
    In addition, the neural network is trained by optimizing the variational bound on negative data log likelihood $\mathbb{E}_{q}[{-\log p_\theta(x_0)}] \leq \mathcal{L} $ to remove standard Gaussian noise. 
    % In addition, the neural network is trained to remove standard Gaussian noise. 
    In each sampling step, the current image $x_t$ is subtracted from the network-estimated noise $\epsilon_{\theta}(x_t, t)$ to obtain the predicted clean image $\hat{x}_0$. Subsequently, based on this predicted clean image $\hat{x}_0$, some network-estimated noise $\epsilon_{\theta}(x_t, t)$ and random Gaussian noise are added to produce the next image $x_{t-1}$. 

    %%%%%%%%%%%%%%%%%%%%%%%%%%%%%%%%%%%%%%%%%%%%%%%%%%%%%%%%%%
%伪代码
\begin{figure}[tb]
%\vspace{-4mm}
% \vspace{-1em}
\begin{algorithm}[H]
  \caption{Iterative Denoising Framework.} \label{alg:iterative framework}
  \small
  \begin{algorithmic}[1]
    \vspace{.04in}
    \State $x_N = y$ 
    \For{$t=N, \dotsc, 1$}
    
          \State $\hat{x}_0 =  Denoiser(x_t)$ 
          \State $x_{t-1}= \hat{x}_0 + \gamma_t (y-\hat{x}_0)$ 
          
    \EndFor
    \State \textbf{return} $\hat{x}_0$
    
    \vspace{.04in}
  \end{algorithmic}
\end{algorithm}
% \vspace{-4mm}
\vspace{-2em}
% \vspace{-4em}
\end{figure}

%%%%%%%%%%%%%%%%%%%%%%%%%%%%%%%%%%%%%%%%%%%%%%%%%%%%%%%%%%
	\subsection{Iterative Denoising Framework}
	\label{sec:framework}

	In this section, we point out the iterative framework for image denoising.
	
	Some distortion-based methods~\cite{osher2005iterative,dong2011sparsity,8481558,KSVD} employ an iterative framework to optimize denoising performance. These methods usually predict a rough image and refine it through many iterations. Such an iterative framework breaks down the problem into a series of sub-problems, leading to a coarse-to-fine denoising process. 
	The iterative denoising framework can be developed as Algorithm \ref{alg:iterative framework}.

	Specifically, the inference of this framework includes several iterations. In each iteration, a rough denoising result $\hat{x}_0 $ is firstly estimated from the current image $x_t$. After that, a weighted original noisy image $y$ is introduced to obtain a corrected version $x_{t-1}$ with lower noise than the previous image $x_t$, as: 
	\begin{alignat}{2}
		x_{t-1}= \hat{x}_0 + \gamma_t (y-\hat{x}_0)  \ .
		\label{eq:iterative-renoise}
	\end{alignat}
	This process is repeated iteratively to refine the final result. 
	The iterative denoising framework transforms denoising into an alternating solution process of prior and recovery terms, which is a special case of plug-and-play methods~\cite{8481558,2021Plug}.

\begin{figure}[tb]
%\vspace{-4mm}

% \begin{algorithm}[H]
%   \caption{Iterative Denoising Framework.} \label{alg:iterative framework}
%   \small
%   \begin{algorithmic}[1]
%     \vspace{.04in}
%     \State $x_N = y$ 
%     \For{$t=N, \dotsc, 1$}
    
%           \State $\hat{x}_0 =  Denoiser(x_t)$ 
%           \State $x_{t-1}= \hat{x}_0 + \gamma_t (y-\hat{x}_0)$ 
          
%     \EndFor
%     \State \textbf{return} $\hat{x}_0$
    
%     \vspace{.04in}
%   \end{algorithmic}
% \end{algorithm}
% \vspace{-4mm}

\begin{algorithm}[H]
% \vspace{-1em}
  \caption{Our Denoising Strategy.} \label{alg:Inference}
  \small
  \begin{algorithmic}[1]
    \vspace{.04in}
    %\For{$r=R_t, \dotsc, 1$}
    
    \State $x_N = \sqrt{\bar{\alpha}_N} Transform(y)$ %\tcp{Jump}
	\State $k = \text{floor}({N/S_t})$
	% \State $\hat{x}=\textbf{0}$

	%\For{$t=N, \dotsc, 1$}
	\For{$i=1, \dotsc, R_t$}
	    \For{$t$ in reversed(range$(0,N,k)$)}
	
	          %第一个版本
	          %\State $\hat{x}_{0}=  \frac{1}{\alpha_t}\left(x_t - \sigma_t \epsilon_\theta(x_t, t)\right)$ %\tcp{Restore} 
	          %\State $\epsilon \sim \mathcal{P}$
	          %\State $x_{t-1}=\alpha_{t-1} \hat{x}_0 + \sigma_{t-1} ( \gamma \epsilon_\theta(x_t, t) + \sqrt{1-\gamma} \epsilon )$ %\tcp{Renoise}
	          
	          %第二个版本
	          %\State $x_{t-1} \sim \mathcal{N}(x_{t-1}; \mu_{\theta}(x_t, t), \sigma_{\theta}^2)$
	
	          %第三个版本
	          %\State $\hat{x}_{0}= \mu_{\theta}(x_t, t)$
	          \State $\epsilon \sim \mathcal{N}(0,1)$
	          \State $\hat{x}_{0}= \frac{1}{{\sqrt{\bar{\alpha}_t}}}{(x_t - \sqrt{1-\bar{\alpha}_t} \epsilon_{\theta}(x_t, t) )}$
	          % \State $x_{t-1} =\sqrt{\bar{\alpha}_{t-1}} \hat{x}_0 + {\sqrt{1-\bar{\alpha}_{t-1}-\sigma_t^2}\epsilon_{\theta}(x_t, t) +\sigma_t \epsilon} $
           \State $x_{t-k} =\sqrt{\bar{\alpha}_{t-k}} \hat{x}_0 + {\sqrt{1-\bar{\alpha}_{t-k}-\sigma_t^2}\epsilon_{\theta}(x_t, t) +\sigma_t \epsilon} $
	          
	    \EndFor
	    % \State $\hat{x}=\hat{x}+\hat{x}_0$
        %mmse版本
        \State $c_i=\hat{x}_0$
	\EndFor
	% \State $\hat{x}=\hat{x}/R_t$
    %mmse
    % \State $\hat{x}=\frac{\sum_{i=1}^{R_t} c_i p(y|c_i)p(c_i)}{\sum_{i=1}^{R_t} p(y|c_i)p(c_i)}$
    % \State $\hat{x}=\sum_{i=1}^{R_t} c_i p(y|c_i)p(c_i)/\sum_{i=1}^{R_t} p(y|c_i)p(c_i)$
    \State $\hat{x}=\sum_{i=1}^{R_t} c_i p(y|c_i)/\sum_{i=1}^{R_t} p(y|c_i)$
    
    \State Inverse Transform $\hat{x}$
    \State \textbf{return} $\hat{x}$
    \vspace{.04in}
  \end{algorithmic}
\end{algorithm}
\vspace{-2em}
\end{figure}

%%%%%%%%%%%%%%%%%%%%%%%%%%%%%%%%%%%%%%%%%%%%%%%%%%%%%%%%%%
	
	\subsection{Discussion}
	\label{sec:discussion}

    {
    % \color{red}
    \noindent \textbf{Denoising Perspective Understanding.}
    % 结构follows功能，这是一个广为人知的法则，而相似的结构可以支撑相似的功能。
	% diffusion model和迭代去噪框架在结构上的高度相似性启发我们rethink他们的相关性from a denoising perspective。	
     Form follows function~\cite{sullivan1922tall}, which is a well-known law, indicating that similar forms can support similar functions. The high formal similarity between the diffusion model and the iterative denoising framework inspires us to explore their relevance.

    % 扩散模型和迭代去噪框架从形式到细节上都是相似性。
	% 从形式上看，这两种技术都构建了一种逐渐生成高质量图像的过程，通过产生了一个粗略估计的图像，并引入了加权的原始带噪图像 y 来获得低噪声的修正版本。
	The diffusion model and the iterative denoising framework are similar from structures to details. 
	In terms of structures, both techniques build a coarse-to-fine iterative process to generate high-quality images. Each step produces a better iteration result by the weighted fusion of coarse denoised image $\hat{x}_0$ and noisy image $y$.}
    In terms of details, both the sampling process \cref{eq:reverse_ddim} and iteration \cref{eq:iterative-renoise} can be decoupled into a prediction item $\hat{x}_0$ and an additional noise item. 
    This additional noise is mainly a residual noise as $\epsilon_{\theta}(x_t,t)$ in \cref{eq:reverse_ddim} and $y-\hat{x}_0$ in \cref{eq:iterative-renoise}. 
    {
    % \color{red}
    Therefore, we declare that the diffusion model and the iterative denoising framework are indeed similar in form, where one sampling step of the former corresponds to one iteration of the latter.
    }

    {
    % \color{red}
    Based on the above observations, we can rewrite the sampling process of the diffusion model from a denoising perspective. 
    % Suppose that we start sampling from a given noisy image $y$. The generated image $x_0$ after $N$ steps of reverse process can be obtained by accumulating \cref{eq:reverse_ddim}:
    Suppose that we start sampling from the intermediate state $x_N$ at timestep $N$. The generated image $x_0$ after $N$ steps of reverse process can be obtained by accumulating \cref{eq:reverse_ddim}:
    }
	\begin{multline}
		\label{eq:x0_accumulate_5}
		x_0 = Denoiser(y,\frac{\sqrt{1-\bar{\alpha}_{N}}}{\sqrt{\bar{\alpha}_{N}}})+\sum_{t=1}^{N-1}{\frac{\sigma_{t+1}}{\sqrt{\bar{\alpha}_{t}}} (\epsilon_{t+1}-\epsilon_{\theta}(x_t, t))}\\ 
		+\sum_{t=1}^{N-1}{\frac{\sqrt{1-\bar{\alpha}_{t}-\sigma_{t+1}^2}}{\sqrt{\bar{\alpha}_{t}}} (\epsilon_{\theta}(x_{t+1},t+1)-\epsilon_{\theta}(x_t, t))}   \ ,
	\end{multline}
	where $y=\frac{x_N}{\sqrt{\bar{\alpha}_{N}}}=x+\frac{\sqrt{1-\bar{\alpha}_{N}}}{\sqrt{\bar{\alpha}_{N}}} \epsilon$ and  $\epsilon \sim \mathcal{N}(0,1)$. Here $x$ represents the desired clean image, and $y$ denotes its noisy observation. The derivation of \cref{eq:x0_accumulate_5} is given in the supplementary material.
	
	% The difference between denoising and \cref{eq:x0_accumulate_5} lies in the cumulative items. Here $\epsilon_{\theta}(x_t, t)$ is the neural network trained to remove the noise added in $x_t$, and during reverse process such a  noise is the additional noise described in \cref{eq:reverse_ddim}. Therefore the cumulative items represent the additional noise added and removed during the reverse process. Thus the cumulative items are assumed to be 0 theoretically. In fact, the additional noise does not affect the level of denoising, but it will play a role on the perception. We will further illustrate its effect in Section \ref{sec:strategy}.

	The difference between a simple denoiser and \cref{eq:x0_accumulate_5} lies in the cumulative terms. 
    Here the cumulative terms represent the additional noise added and removed during the reverse process.
    During the reverse process, the network-estimated noise $\epsilon_{\theta}(x_t, t)$ is subtracted to 
    obtain the predicted clean image $\hat{x}_0$, and is added back to produce the next image $x_{t-1}$. Theoretically, the cumulative terms should be zero and not affect the level of denoising. In fact, the iterative denoising framework also has a similar cumulative term.

    {
    % \color{red}
    Overall, we introduce a novel denoising perspective for understanding the diffusion model. The diffusion model can be classified as a variant of the iterative denoising framework. The neural network of the diffusion model can be 
    % regarded as a conditional denoiser, where timestep $t$ as a condition variable 
    regarded as a denoiser, where timestep $t$ as a variable 
    corresponds to the preset noise level ${\sqrt{1-\bar{\alpha}_{t}}} / {\sqrt{\bar{\alpha}_{t}}}$.}
    
    Figure \ref{fig:diffusion model-iterative denoising} illustrates the evolution of the intermediate state $x_t$, the estimated image $\hat{x}_0$, and the corresponding noise level during the reverse process.	
	% Figure \ref{fig:diffusion model-iterative denoising} illustrates the evolution of the noisy images $x_t$, the estimated clean image $\hat{x}_0$, and corresponding noise level during the reverse process.
    In the early stages of the reverse process, the noise level is high and the denoising effect is limited. So only low-frequency information can be determined. In the later stages of the reverse process, the remaining noise is minimal, and then high-frequency details are generated with iterations. 
		
	Denoising perspective understanding will intuitively and clearly guide us to stimulate the diffusion model for image denoising. To the best of our knowledge, this is the first time that the diffusion model is comprehended in such a concise denoising perspective.

	% \begin{figure*}[t]
% \centering
%  \includegraphics[width=0.99\textwidth]{Images/method_l.pdf}
%  \caption{Our Diffusion Model for Image Denoising (DMID) strategy. Our adaptive embedding method connects diffusion model and image denoising, while our adaptive ensemble method reduces distortion in the final denoised image.}
%  \label{fig:whole framework}

% \end{figure*}
% \begin{figure}[t]
% \vspace{1em}
%  \includegraphics[width=0.49\textwidth]{Images/method_s.pdf}
%  \caption{The details of sampling process.}
%  \label{fig:sampling process}
% \end{figure}

% \begin{figure*}[ht]
% \centering
% \begin{tabular}[b]{c c}
% \subfloat[whole framework]{\includegraphics[width=0.605\linewidth]{Images/method_l.pdf}
%  }
% \subfloat[sampling process]{\includegraphics[width=0.375\linewidth]{Images/method_s.pdf}
% }
% \\
% \end{tabular}

% \caption{Our Diffusion Model for Image Denoising (DMID) strategy. Our adaptive embedding method connects diffusion model and image denoising, while our adaptive ensemble method reduces distortion in the final denoised image. (a) is the whole framework, (b) is the procedure of the sampling process.
% }
% \label{fig:method}
% \end{figure*}

\begin{figure*}[ht]
    \begin{center}
        \subfigure[whole framework]{ \label{a}
            \includegraphics[width=0.59\linewidth]{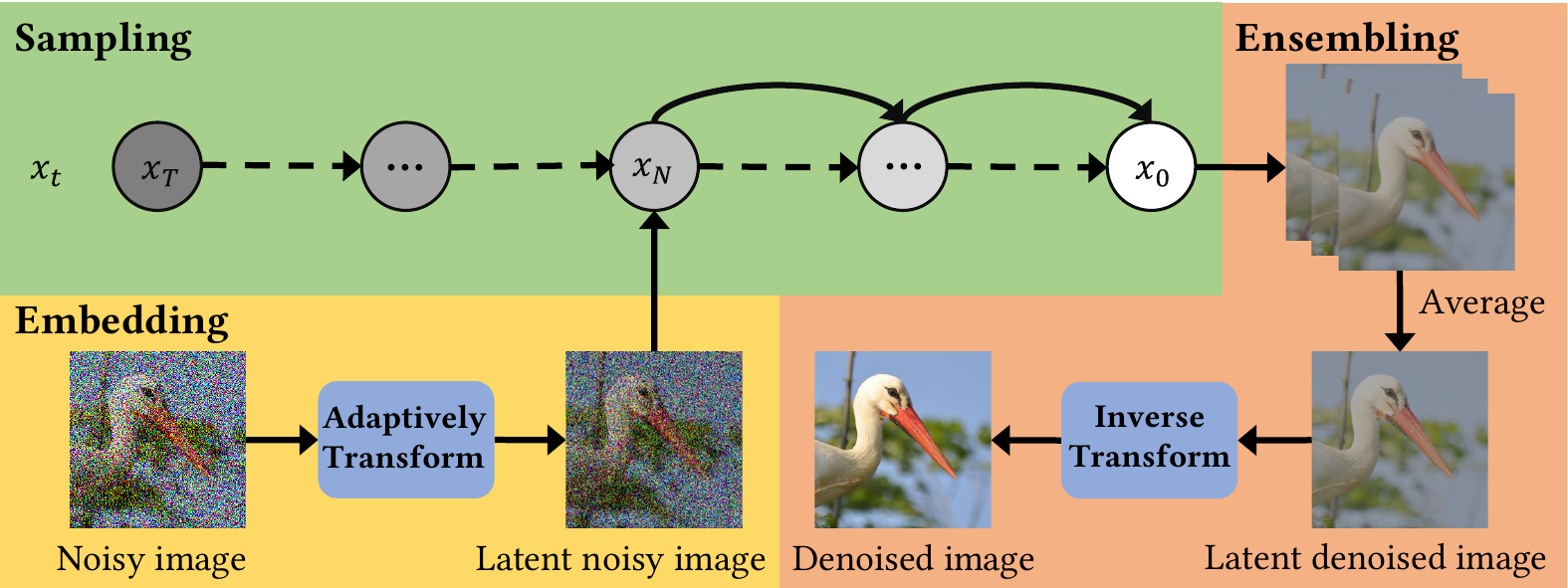}
        }
        \subfigure[sampling process]{ \label{b}
            \includegraphics[width=0.37\linewidth]{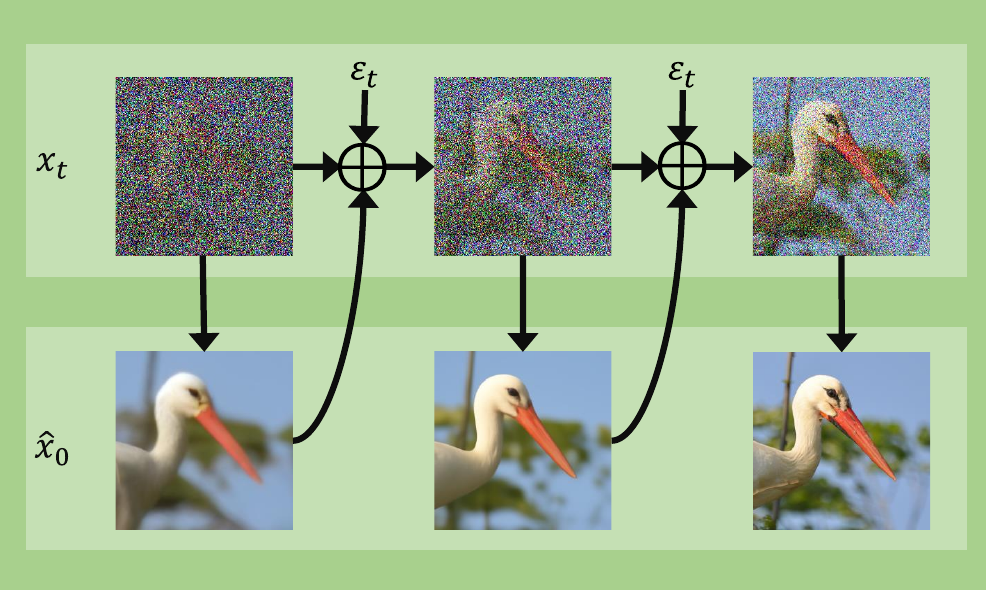}
        }
    \end{center}
        \vspace{-1em}
    \caption{Our Diffusion Model for Image Denoising (DMID) strategy. Our adaptive embedding method connects diffusion model and image denoising, while our adaptive ensemble method reduces distortion in the final denoised image.}
    \label{fig:method}
     \vspace{-1em}
\end{figure*}

    {
    % \color{red}
    \noindent \textbf{Denoising Perspective Rethinking.}
    % Referring to extensive existing research in image denoising, the denoising perspective can help us rethink some practical details of the diffusion model in an intuitive manner. For instance, explaining the sampling process through Markov chains makes it challenging to justify why $p_{\theta}(x_{t-1}|x_t)$ is generally expressed as Eq.~\eqref{eq:siglereverse} in practice rather than being directly approximated by neural networks~\cite{ho2020denoising}. From a denoising perspective, this is a classic learnability problem~\cite{NN89/learnability, PMN} of neural networks, involving the well-known deep image prior~\cite{DIP} in image denoising. Since $x_{t-1}$ and $x_t$ are noisy images with similar noise levels, $p_{\theta}(x_{t-1}|x_t)$ is much more difficult to approximate by neural networks than $p_{\theta}(x_0|x_t)$.
    % 
    % In this paper, we focus on stimulating the diffusion model for image denoising.
    The denoising perspective of the diffusion model is the starting point for us to rethink and stimulate the diffusion model for image denoising. 
	The input of diffusion models is standard Gaussian noise while the input of image denoising is a noisy image, which constitutes input inconsistency.
	The content difference between the generated image and the desired denoised image introduces distortion, which constitutes content inconsistency.
	Guided by the denoising perspective, the approach to addressing input inconsistency and content inconsistency naturally emerges.
    }

	{
    % \color{red}
    For input inconsistency, the solution lies in the comparison between the input of the diffusion model and the iterative denoising framework.
	From a denoising perspective, the diffusion model is a conditional AWGN denoiser, iteratively denoising according to a schedule where the expected noise level of the intermediate state $x_t$ gradually decreases.
	%Due to the wide coverage of noise level schedules during training, as shown in Figure \ref{fig:diffusion model-iterative denoising}, the pre-trained diffusion model effectively functions as an iterative denoiser capable of covering any noise level.
	%The iterative denoiser has never placed any limit on the noise level of the input.
	Therefore, by proposing a method to embed the input into the corresponding intermediate state of the pre-trained diffusion model based on the noise level, we naturally address input inconsistency.

    For content inconsistency, the solution lies in the comparison between the process of the diffusion model and the iterative denoising framework.
	The diffusion model excels in preserving high perceptual quality, while the classical iterative denoising framework primarily aims to reduce distortion.
	Based on our observations, the difference is mainly attributed to the introduction of random noise $\epsilon$ besides the original noisy image $y$ during the sampling process of the diffusion model.
	The excessive stochasticity of random noise $\epsilon$ allows the diffusion model to potentially generate content beyond the original clean image $x_0$.
	From the denoising perspective, this phenomenon corresponds to the local content flickering caused by the stochasticity of i.i.d. noise in low-light video denoising~\cite{2019DRV}.
	% Inspired by the late fusion in video processing, 
    Taking cues from the late fusion in video processing, 
    % Therefore, 
    % Inspired by the Monte Carlo method,
    we introduce an ensembling method to effectively adjust distortion and perception in the diffusion model, thereby addressing content inconsistency.
	
	It is worth highlighting that our method is insightful compared to existing diffusion-based image denoising methods~\cite{kawar2022denoising,wang2022zero}. Existing methods are not specifically designed for image denoising, thus they typically start from pure Gaussian noise and 
    % require additional conditions besides timestep $t$
    recommend extremely high sampling times for one inference
    , which is time-consuming and inefficient. The denoising perspective helps us rethink the diffusion model and addresses the problems of the input inconsistency and the content inconsistency in a concise and efficient manner.
    }

	\section{Method}
	\label{sec:strategy}
	%\subsection{Denoising Strategy based on Diffusion Models}
	%\subsection{Our Method}
	%\subsection{Adaptive Embedding and Ensembling}
	
	In this section, we first give an overview of our DMID strategy in Section \ref{sec:overview}, which includes an embedding method and an ensembling method. Then we describe the procedure for the embedding method and ensembling method in Section \ref{sec:embedding} and Section \ref{sec:ensembling}, respectively.
	
	\subsection{Overview}
	\label{sec:overview}
	
	In this section, we propose a novel strategy to stimulate the diffusion model for image denoising (DMID), which mainly consists of an adaptive embedding method and an adaptive ensembling method. In summary, we first perform the embedding method to construct the initial sampling state. Subsequently, we generate multiple denoised images using the pre-trained diffusion model through multiple inferences. Finally, we employ the ensembling method to reduce distortion.

    {
    % \color{red}
    The embedding method first transforms the input noisy image $y$ into a latent space where the Gaussian noise assumption is valid. Subsequently, the embedding method normalizes the image range to match the data range expected by diffusion models. Ultimately, the embedding method converts the latent image into the intermediate $x_N$ state of the diffusion model. The embedding method can be summarised as:
        \begin{equation}
        x_N=\sqrt{\bar{\alpha}_N}\ Transform(y)  \ ,
        \label{eq:jump}
    \end{equation}
    where $0 \leq N <T$ is an intermediate timestep.
    The function $Transform(y)$ serves as both a noise transformation technique and a data normalizer.
    % Initially, the function $Transform(y)$ transforms noisy images into a latent space where the Gaussian noise assumption is valid. Subsequently, it adjusts the range of the latent noisy image to align with the data range of the diffusion model, which typically spans from -1 to 1.
    Following the $Transform(y)$ function, we multiply the latent noisy image by $\sqrt{\bar{\alpha}_N}$ to obtain the intermediate $x_N$ state of the diffusion model. The value of $N$ should be chosen carefully according to the noise level of the latent noisy image to achieve optimal denoising. After that, we conclude the embedding method. 
    }

    Furthermore, we then generate a desirable denoised image by directly sampling from timestep $N$. The sampling strategy can be any existing approach, such as  DDIM~\cite{song2020denoising}, DDRM~\cite{kawar2022denoising}, or DDNM~\cite{wang2022zero} among others. In our experiments, we simply employ DDIM. The denoising process, achieved through direct sampling from timestep $N$, can be mathematically described as:
	\begin{equation}
		p(x_0|y) = \prod_{t=1}^N p_{\theta}(x_{t-1}|x_t)  \ ,
		\label{eq:renoise}
	\end{equation}
	where $x_0$ is the denoised output image. 
    DMID performs denoising within the latent space and, thus needs to inverse transform the latent denoised image, which is originally within the data range of the diffusion model, back to the original data range.

	In addition, we employ our ensembling method to further reduce distortion. Specifically, we adjust the sampling times (referred to $S_t$) in one inference process and repeat the inference process several times (referred to $R_t$). Each inference on the same image corresponds to a Monte Carlo sampling, yielding $R_t$ denoised images. To mitigate stochasticity and distortion, we conduct a Minimum Mean Square Error (MMSE) averaging of all denoised images, ultimately obtaining the final result.

	The whole procedure for the inference of our method is described in Figure \ref{fig:method} and Algorithm \ref{alg:Inference}. We will delve into the procedures of the embedding method and ensembling method in the following sections.

 % \begin{figure}[t]
%     \begin{center}
%      \includegraphics[width=1\linewidth]{Images/Embedding_method.pdf} 
%     \end{center}
%     \caption{The procedure for the embedding method. The embedding method first transforms the noise into Gaussian noise and subsequently normalizes the latent noisy image. After that, we search all the timestep $t$ to find a timestep $N$ to guarantee $\frac{\sqrt{1-\bar{\alpha}_{N}}}{\sqrt{\bar{\alpha}_{N}}}$ closest to $\sigma$. Ultimately, we multiply the noisy image by $\sqrt{\bar{\alpha}_N}$ to convert the image to the intermediate state $x_N$.}
%     \label{fig:embedding_method}
% \end{figure}
\begin{figure*}[t]
    \begin{center}
     \includegraphics[width=1\linewidth]{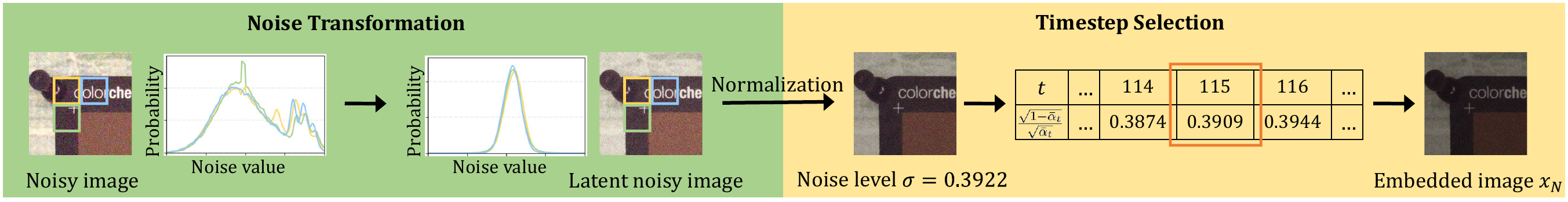}
    \end{center}
    \vspace{-1em}
    \caption{The procedure for the embedding method. The embedding method first transforms the noise into Gaussian noise and subsequently normalizes the latent noisy image. After that, we search all the timestep $t$ to find a timestep $N$ to guarantee $\frac{\sqrt{1-\bar{\alpha}_{N}}}{\sqrt{\bar{\alpha}_{N}}}$ closest to $\sigma$. Ultimately, we multiply the noisy image by $\sqrt{\bar{\alpha}_N}$ to convert the image to the intermediate state $x_N$.}
    \label{fig:embedding_method}
    \vspace{-1em}
\end{figure*}
 
	%\paragraph{\textbf{The procedure of adaptive embedding}} 
	\subsection{Adaptive Embedding}
	\label{sec:embedding}
	
	In this section, we will describe the procedure for the embedding method. 
    As discussed before, the diffusion model is reframed as a Gaussian denoiser. The intermediate states within the diffusion model are adept at handling Gaussian noise with distinct levels. Consequently, our embedding method is tasked with transforming the noise within a noisy image into Gaussian noise and converting the image into a suitable intermediate state based on the noise level. The procedure of the embedding method is shown in Figure \ref{fig:embedding_method}.

	{
		The noise transformation technique involves converting different types of noise into Gaussian noise. 
        The primary goal of our approach 
        is to address general image denoising, allowing it to handle various forms of image noise. Although we have successfully treated the diffusion model as a Gaussian denoiser, it is crucial to account for other noise types, such as real-world noise, which does not adhere to a Gaussian distribution. 
        % To surmount this challenge, our method employs a noise transformation technique NN~\cite{zheng2021unsupervised} to convert diverse noise types into Gaussian noise. 
        To surmount this challenge, we improve a noise transformation technique NN~\cite{zheng2021unsupervised} to convert diverse noise types into Gaussian noise. 
        In essence, noise transformation entails finding a latent image $z$ that exhibits correlation with the input noisy image $y$, while the noise in $z$ conforms to the assumption of additive white Gaussian noise (AWGN). NN~\cite{zheng2021unsupervised} achieves this by training an encoder-decoder neural network using a single input noisy image for autoregressive modeling. The latent image $z$ is derived from the encoder and is expected to adhere to the AWGN assumption. We designate the encoder network of the VAE with parameters $\theta_1$ as $G_{\theta_1}$, the decoder network parameterized by $\theta_2$ as $F_{\theta_2}$, and the latent image as $z=G_{\theta_1}(y) \sim \mathcal{N}(x,\sigma^2I)$. The loss function of the noise transformation technique is presented as:
		%\begin{small}
		\begin{multline}
			\mathcal{L}(x,\theta_1,\theta_2)= 
			\frac{1}{2} E_{\epsilon}{\| F_{\theta_2}(G_{\theta_1}(y)+\epsilon)-y \|}^2 
			\\
			+ \frac{1}{2\sigma^2} {\|G_{\theta_1}(y)-x\|}^2 
			+ \lambda R(x)\ 
			,
			\label{eq:VAEloss}
		\end{multline}
		%\end{small}
		where $x$ represents the clean image, $R(x)$ is a regularization function, and $\epsilon \sim \mathcal{N}(0,1)$. 
        The derivation of \cref{eq:VAEloss} is given in the supplementary material.
		%In our specific approach, we approximate the likelihood term in the maximum a posteriori (MAP) framework by utilizing the evidence lower bound (ELBO) of a neural network. We introduce a regularization term $R(x)$ to construct a traditional MAP architecture. For unsupervised training using only the input image, we employ an iterative optimization method called ADMM to optimize \cref{eq:VAEloss}:
	%In the specific approach, 
        %the evidence lower bound (ELBO) of a neural network approximates the likelihood term within the maximum a posteriori (MAP) framework. The regularization term $R(x)$ and the likelihood term together form a traditional MAP architecture. 
        % the first term and the second term guarantee the data fidelity, the final term serves for regularization and the three terms work together to prevent zero mapping and identity mapping.
         % \OLD
         {In the specific approach, three terms in the \cref{eq:VAEloss} work in synergy. The first and second terms ensure data fidelity, while the final term serves as a regularization component. Together, these three terms prevent zero mapping and identity mapping.}
        To facilitate unsupervised training using only the input noisy image $y$ without the clean image $x$, \cref{eq:VAEloss} is optimized through ADMM~\cite{admm}:
		\begin{multline}
			\mathcal{L}_{\rho}(\bar{x},\theta_1,\theta_2,p,q)
			= \mathcal{L}(\bar{x},\theta_1,\theta_2)
			- \lambda R(\bar{x})    
			+ R(p) 
			\\
			+\frac{\rho}{2}{\| \bar{x}-p+\frac{q}{\rho}\|^2}- \frac{\rho}{2}{\| \frac{q}{\rho} \|^2} \ ,
			\label{eq:Lagrangian}
		\end{multline}
		\begin{equation}
			\bar{x}^{k+1},\theta^{k+1}_1,\theta^{k+1}_2 = 
			\mathop{\arg\min}\limits_{\bar{x},\theta_1,\theta_2}
			\mathcal{L}_{\rho}(\bar{x},\theta_1,\theta_2,p^{k},q^{k})
			\ ,
			\label{eq:alternating1}
		\end{equation}
		\begin{equation}
			p^{k+1} = 
			\mathop{\arg\min}\limits_{p}
			\mathcal{L}_{\rho}(\bar{x}^{k+1},\theta_1^{k+1},\theta_2^{k+1},p,q^{k})
			\ ,
			\label{eq:alternating_p}
		\end{equation}
		\begin{equation}
			q^{k+1} = 
			q^{k}+\rho( \bar{x}^{k+1}-p^{k+1} )
			\ ,
			\label{eq:alternating_q}
		\end{equation}
		where $\bar{x}$ is an estimation of the clean image $x$, $p$ is an auxiliary variable, $q$ is the dual variable and $\rho > 0$ is a chosen constant. The subproblems are solved by alternating minimization since the clean image $x$ is not accessible during unsupervised training. Specifically \cref{eq:alternating1} is solved by updating the clean image estimation $\bar{x}$ and the network parameters $\theta_1$ and $\theta_2$ alternatively. In addition, \cref{eq:alternating_p} is solved through the plug-and-play idea:
		\begin{equation}
			p^{k+1} = D( \bar{x}^{k+1}+q^{k}/\rho )
			\ ,
			\label{eq:alternating_q_closedform}
		\end{equation}
		where $D$ is any existing Gaussian denoiser. We choose BM3D~\cite{2007Image}. Instead of directly using the latent image $z=G_{\theta_1}(y)$, NN employs a linear combination of the noisy image $y$ and the latent image $z$ as the final result after transformation.

        However, determining when to stop the optimization is a critical challenge. The original NN method required clean images to calculate PSNR and determine the stopping time. We improve this process by using SURE (Stein's Unbiased Risk Estimator)~\cite{sure,suredenoise}, which eliminates the need for clean images to determine when to stop. SURE provides an unbiased estimate of MSE and represents the quality of the image:
        \begin{equation}
			SURE(z) = \frac{\| z-D(z) \|^2}{K}-\sigma^2+\frac{2\sigma^2}{K}\sum_{i=1}^{K}\frac{\partial D_i(z)}{\partial z_i}
			\ ,
			\label{eq:sure}
	\end{equation}
        where $K$ is the image size and $z_i$ is the $i$th element of $z$.
        Additionally, we employ the Monte-Carlo (MC) approximation~\cite{ramani2008monte} of the divergence term in \cref{eq:sure} as follows:
         \begin{equation}
			\sum_{i=1}^{K}\frac{\partial D_i(z)}{\partial z_i}
            \approx \frac{1}{\mu}\epsilon^{T}(D(z+\mu \epsilon)-D(z)
			\ ,
			\label{eq:sure-mc}
	\end{equation}
        where $T$ is the transpose operator and $\mu$ is a small positive value.
        % Every 500 iterations, we calculate SURE of the latent image $z$ and the denoised image of the latent image.
        % Smaller SURE values indicate higher image quality, and we stop the iterations when SURE begins to increase as illustrated in Figure \ref{fig:psnr-sure}.
        % It merits mentioning that
        % our improved transformation technique requires no pre-training on external datasets or the clean image $x$, as the optimization process solely relies on the single input noisy image $y$. 
        Every 500 iterations, we calculate the SURE of the latent image $z$. Smaller SURE values indicate higher image quality, and we stop the iterations when SURE begins to increase.
        % For instance, Figure \ref{fig:psnr-sure} illustrates the changes in PSNR and -SURE during the noise transformation process for four images. The noise transformation processes of the four images stop after 1500 iterations, with the results from the 1000$th$ iteration chosen as the final outcome.
        It merits mentioning that our improved transformation technique requires no pre-training on external datasets or the use of the clean image $x$. The noise transformation process solely relies on the single input noisy image $y$.
	}

	The value of $N$ should be chosen carefully to achieve optimal denoising performance based on the noise level of the latent noisy image. 
    For the latent noisy image with noise level $\sigma$, we search all the timestep $t$ to find a timestep $N$ to guarantee $\frac{\sqrt{1-\bar{\alpha}_{N}}}{\sqrt{\bar{\alpha}_{N}}}$ closest to $\sigma$. After that, we multiply the noisy image by $\sqrt{\bar{\alpha}_N}$ to convert the image to the intermediate state $x_N$.
    % Some explanations are as follows.
    % To attain optimal denoising outcomes, we should align the denoising ability of the diffusion model at timestep $N$ with the noise level $\sigma$ of the latent noisy image. 
    The reason is that we should align the denoising ability of the diffusion model at timestep $N$ with the noise level $\sigma$ of the latent noisy image to attain optimal denoising outcomes. And we consider the diffusion model as a denoiser, whose denoising ability at timestep $t$ is ${\sqrt{1-\bar{\alpha}_{t}}} / {\sqrt{\bar{\alpha}_{t}}}$, as discussed in Section \ref{sec:discussion}. 
    % In practice, we calculate the corresponding noise level for each timestep in advance. 
    % For a latent noisy image with noise level $\sigma_1$, we readily identify an $N$ that corresponds to the noise level $\sigma_2$ closest to $\sigma_1$. 
    % For tasks such as Gaussian denoising, the noise level of the latent noisy image is known. In cases of unknown noise levels, established noise estimation techniques~\cite{1640848,2008Practical} can be applied to address the issue.
    In practice, we calculate the corresponding level $\frac{\sqrt{1-\bar{\alpha}_{t}}}{\sqrt{\bar{\alpha}_{t}}}$ for each timestep $t$ in advance. 
    For a latent noisy image with noise level $\sigma$, we readily identify an $N$ that corresponds to the level $\frac{\sqrt{1-\bar{\alpha}_{N}}}{\sqrt{\bar{\alpha}_{N}}}$ closest to $\sigma$. 
    For instance, we choose $N=115$ for noise level $\sigma =\frac{2*50}{255}\approx0.3922$, as ${\sqrt{1-\bar{\alpha}_{t}}} / {\sqrt{\bar{\alpha}_{t}}}$ closest to $\sigma$ at $t=115$ among all timestep $t$, as shown in Figure \ref{fig:embedding_method}.
	 For tasks such as Gaussian denoising, the noise level of the latent noisy image is known. In cases of unknown noise levels, established noise estimation techniques~\cite{1640848,2008Practical} can be applied to address the issue.
	
	\subsection{Adaptive Ensembling}
	%\paragraph{\textbf{The procedure of adaptive ensembling}}
	\label{sec:ensembling}
	
	In this section, we will describe the procedure for the ensembling method. 
    % The primary objective of the ensembling method is to reduce distortion. In essence, our embedding method is tasked with adaptively adjusting distortion and perception based on specific requirements.
    Our embedding method is tasked with adjusting distortion and perception based on specific requirements.
	The primary factor influencing distortion and perception is the stochasticity brought by the Gaussian noise in the additional noise item. 
	In practice, excessive stochasticity can lead to significant distortion in the final denoised image.
	In theory, moderate stochasticity can facilitate convergence to better results~\cite{song2019generative}. 
	% Thus our ensembling strategy involves adjusting the sampling times to control stochasticity in one inference process and repeating the inference process multiple times to constrain stochasticity and converge to higher-quality images.
    Thus our ensembling strategy involves adjusting the sampling times to control stochasticity in one inference process and averaging the images obtained from multiple inferences to converge to higher quality.
	
	Firstly we constrain the stochasticity in one inference. 
    %The stochasticity is brought by the Gaussian noise in the additional noise item. And it is determined by the number of sampling times, as the additional noise item appers in each sampling process. 
    The level of stochasticity is primarily determined by the number of sampling times since the additional noise item appears in each sampling process.
    A higher number of sampling times{ 
    % \color{red}
    $S_t$} allows for more refinement and introduces more stochasticity, resulting in a more detailed image with improved perceptual quality. However, this also leads to a smaller weighted noisy image $y$ at each iteration, which reduces its controllability at the same time. Therefore, larger sampling times generally yield better perceptual quality but also result in larger distortion. To adjust distortion and perception, we recommend adjusting the number of sampling times according to the desired outcome. To reduce distortion, setting sampling times to be less than 10 is typically sufficient.

\begin{table*}[t]
\begin{center}
\caption{Classical Gaussian image denoising. The top row is training a separate model for specific noise level, the bottom row is the models that designed to deal with various noise levels. The best and second-best methods are in {\color{red}red} and {\color{blue}blue}.}
\label{table:1-calssical}
% \vspace{0.5mm}
\renewcommand{\arraystretch}{1.2}
%\vspace{-1em}
\setlength{\tabcolsep}{8pt}
\scalebox{0.99}{
\begin{tabular}{l | c c c | c c c | c c c | c c c }
\toprule
   & \multicolumn{3}{c|}{{CBSD68}~\cite{martin2001database}} & \multicolumn{3}{c|}{{Kodak24}~\cite{franzen1999kodak}} & \multicolumn{3}{c|}{{McMaster}~\cite{zhang2011color}} & \multicolumn{3}{c}{{Urban100}~\cite{urban100}}\\
 \cline{2-13}
   \multirow{-2}{*}{Method} & $\sigma$$=$$15$ & $\sigma$$=$$25$ & $\sigma$$=$$50$ & $\sigma$$=$$15$ & $\sigma$$=$$25$ & $\sigma$$=$$50$ & $\sigma$$=$$15$ & $\sigma$$=$$25$ & $\sigma$$=$$50$ & $\sigma$$=$$15$ & $\sigma$$=$$25$ & $\sigma$$=$$50$ \\

\midrule
BRDNet~\cite{tian2020image}  &34.10 & 31.43 & 28.16 & 34.88 & 32.41 & 29.22 & 35.08 & 32.75 & 29.52  &34.42 & 31.99 &28.56\\
RNAN~\cite{zhang2019residual}  &-&-&28.27&-&-&29.58&-&-&29.72&-&-&29.08\\
RDN~\cite{zhang2020residual}  &-&-&28.31&-&-&29.66&-&-&-&-&-&29.38\\
IPT~\cite{chen2021pre}  &-&-&28.39&-&-&29.64&-&-&29.98&-&-&29.71\\
SwinIR~\cite{liang2021swinir} & \color{blue}{34.42} & 31.78 & 28.56 & 35.34 & 32.89 & 29.79 & \color{blue}{35.61} & 33.20 & 30.22 & \color{blue}{35.13} &32.90 &29.82 \\
Restormer~\cite{zamir2022restormer} & 34.40 & \color{blue}{31.79}& \color{blue}{28.60}& \color{blue}{35.47} & \color{blue}{33.04}& \color{blue}{30.01}& \color{blue}{35.61}& \color{blue}{33.34}& \color{blue}{30.30} & \color{blue}{35.13}& \color{blue}{32.96}& \color{blue}{30.02}\\

CODE~\cite{zhao2023comprehensive} &34.33 &31.69&28.47 &35.32&32.88& 29.82&35.38&33.11 &30.03&-&-&-\\
   
\midrule

\midrule

CBM3D~\cite{2007Image}  &33.52 & 30.71 & 27.38 & 34.28 & 32.15 & 28.46 & 34.06 & 31.66 & 28.51 & 32.35& 29.70&25.95  \\
DnCNN~\cite{2016Beyond}  &33.90 & 31.24 & 27.95 & 34.60 & 32.14 & 28.95 & 33.45 & 31.52 & 28.62 &32.98 &30.81 &27.59 \\
FFDNet~\cite{2017FFDNet}  &33.87 & 31.21 & 27.96 & 34.63 & 32.13 & 28.98 & 34.66 & 32.35 & 29.18 &33.83 &31.40 &28.05 \\
DSNet~\cite{peng2019dilated}  & 33.91 & 31.28 & 28.05 & 34.63 & 32.16 & 29.05 & 34.67 & 32.40 & 29.28 &-&-&-\\
DRUNet~\cite{2021Plug}  & 34.30 & 31.69 & 28.51 & 35.31 & 32.89 & 29.86 & 35.40 & 33.14 & 30.08 &34.81 &32.60 &29.61 \\
Restormer~\cite{zamir2022restormer} & 34.39 & 31.78 & 28.59 & 35.44 & 33.02 & 30.00 & 35.55 & 33.31 & 30.29 &35.06 &32.91 &\color{blue}{30.02}\\ 

\textbf{DMID-d (Ours)} & \color{red}{34.45} & \color{red}{31.86} & \color{red}{28.72} & \color{red}{35.51} & \color{red}{33.12} & \color{red}{30.14} & \color{red}{35.72} & \color{red}{33.49} & \color{red}{30.50} & \color{red}{35.26} & \color{red}{33.11} & \color{red}{30.28} \\ 
\bottomrule
\end{tabular}
}
\end{center}
\vspace{-1em}
\end{table*}

	Furthermore, we enhance the denoised images through the stochasticity in multiple inferences. The Gaussian noise in the additional noise item introduces stochasticity into the sampling process. This enables generating multiple clean images for the same noisy image when the sampling times are greater than one. 
    {
    % \color{red}
    % Thus, we propose an ensembling approach based on the Monte Carlo method to utilize the stochasticity of multiple inferences.
    % Specifically, we repeat the inference process multiple times to generate multiple candidate denoised images for the same noisy input.
    % We refer to the repetition times of inference as $R_t$, and the candidate denoised image as $c_i$. Additionally, we incorporate the Minimum Mean Square Error (MMSE) averaging \cite{guo2005mutual} to further enhance the ensembling method. 
    % As we reframe the diffusion model as a Gaussian denoiser and transform all the noise to Gaussian noise, we can utilize the existing noisy images \(y\) and the Gaussian noise model \(p(y|x)\) to calculate the observation likelihood \(p(y|c_i)\) for every candidate restored image $c_i$. This likelihood signifies the probability of how well the candidate restored image \(c_i\) corresponds to the denoised outcome of the noisy image \(y\). Thus, by weighting our predicted denoised results with the corresponding observation likelihood, we can further minimize stochasticity. The final denoised image is expressed as 
    % \begin{equation}
    % x^{\text{MMSE}} = \frac{\sum_{i=1}^{R_t} c_i p(y|c_i)p(c_i)}{\sum_{i=1}^{R_t} p(y|c_i)p(c_i)}
    % \ ,
    % \label{eq:mmse_inference}
    % \end{equation}
    % where \(p(y|c_i) \sim \mathcal{N}(c_i,\sigma^2\mathbf{I})\) is calculated by the Gaussian noise model, \(\sigma\) is the noise level of the noisy image, and \(\mathcal{N}\) represents the Gaussian distribution.
    Thus, we propose an ensembling approach based on the Monte Carlo method to utilize the stochasticity of multiple inferences.
    Specifically, we repeat the inference process multiple times to generate multiple candidate denoised images for the same noisy input.
    We refer to the repetition times of inference as $R_t$, and the candidate denoised image as $c_i$. Additionally, we incorporate the Minimum Mean Square Error (MMSE) averaging \cite{guo2005mutual} to further enhance the ensembling method. 
     The final denoised image is expressed as 
    \begin{equation}
    % x^{\text{MMSE}} = \frac{\sum_{i=1}^{R_t} c_i p(y|c_i)p(c_i)}{\sum_{i=1}^{R_t} p(y|c_i)p(c_i)}
    x^{\text{MMSE}} = \frac{\sum_{i=1}^{R_t} c_i p(y|c_i)}{\sum_{i=1}^{R_t} p(y|c_i)}
    \ ,
    \label{eq:mmse_inference}
    \end{equation}
    where \(p(y|c_i) \sim \mathcal{N}(c_i,\sigma^2\mathbf{I})\) is calculated by the Gaussian noise model, \(\sigma\) is the noise level of the noisy image, and \(\mathcal{N}\) represents the Gaussian distribution. 
    The derivation of \cref{eq:mmse_inference} is given in the supplementary material.
    Inspired by our denoising prespective understanding, we reframe the diffusion model as a Gaussian denoiser and transform all noise to Gaussian noise. Thus we can leverage existing noisy images \(y\) and the Gaussian noise model \(p(y|x)\) to compute the observation likelihood \(p(y|c_i)\) for each candidate restored image \(c_i\). This likelihood indicates how well the candidate restored image \(c_i\) aligns with the denoised outcome of the noisy image \(y\). Therefore, by weighting our predicted denoised results with the corresponding observation likelihood, we can further minimize stochasticity. According to the Law of Large Numbers, a higher number of repetition times $R_t$ yield better distortion-based quality.
    }

	%结果图-classial
	
	%一张图片两行分别是不同的位置截取

\begin{figure*}[ht]
\begin{center}
\footnotesize
\renewcommand{\arraystretch}{1.2}
\setlength{\tabcolsep}{1.5pt}
\scalebox{0.97}{
\begin{tabular}[b]{c c c c c c c c}
% \hspace{-4mm}

%481 321   160
\multirow{3}{*}{\includegraphics[trim={  110 0 50 0
 },clip, width=.22\textwidth,valign=t]{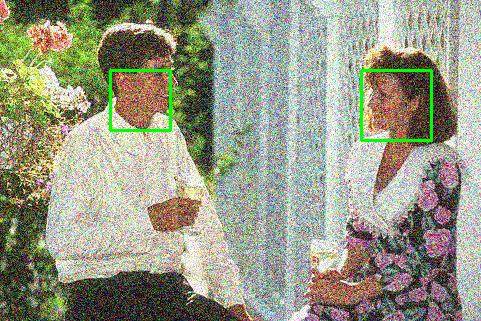}} &   
%left bottom right top     
\includegraphics[trim={  361 181 50 70
 },clip,width=.105\textwidth,valign=t]{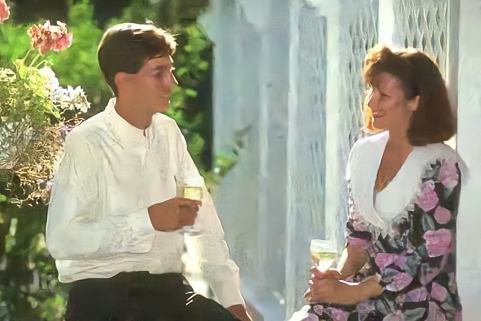} &   
\includegraphics[trim={  361 181 50 70
 },clip,width=.105\textwidth,valign=t]{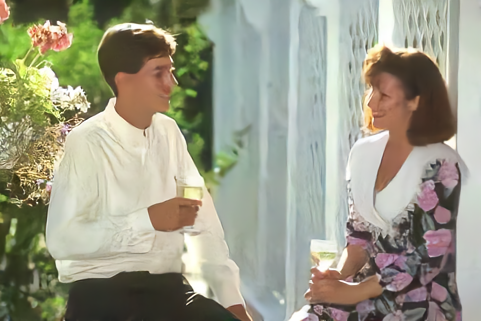} &   
\includegraphics[trim={ 361 181 50 70
 },clip,width=.105\textwidth,valign=t]{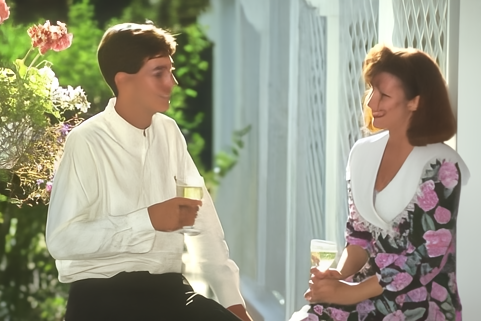} &
\includegraphics[trim={  361 181 50 70
 },clip,width=.105\textwidth,valign=t]{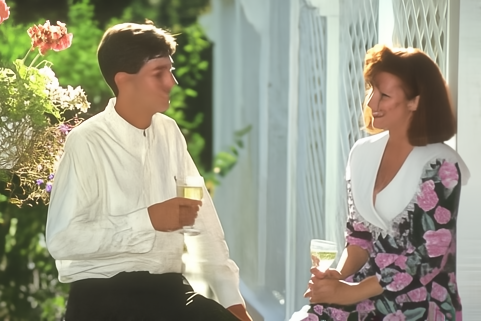} &  
\includegraphics[trim={  361 181 50 70
 },clip,width=.105\textwidth,valign=t]{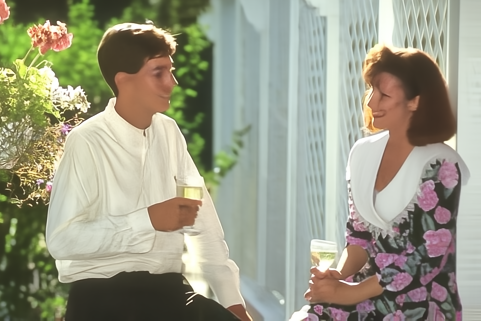} &   
\includegraphics[trim={  361 181 50 70
 },clip,width=.105\textwidth,valign=t]{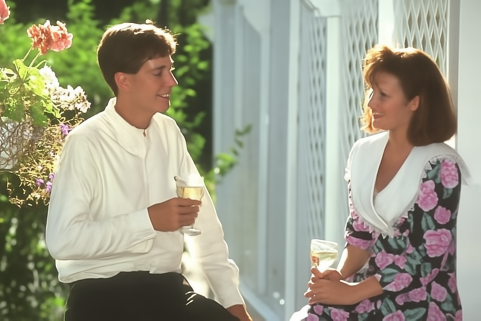} &
 \includegraphics[trim={  361 181 50 70
 },clip,width=.105\textwidth,valign=t]{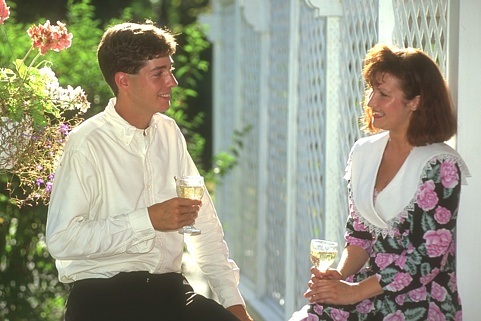} 
\\

% \small~ 14.94 dB  & \small~27.42 dB &  \small~27.47 dB & \small~28.08 dB & \small~\color{blue}{28.17 dB} & \small~\color{blue}{28.17 dB} & \small~\color{red}{28.39 dB} & \small~PSNR
% & 27.42 dB &  27.47 dB & 28.08 dB & \color{blue}{28.17 dB} & \color{blue}{28.17 dB} & \color{red}{28.39 dB} & PSNR
\addlinespace[2pt]
% &DnCNN%~\cite{2016Beyond} 
% & FFDNet%~\cite{2017FFDNet} 
% & DRUNet%~\cite{2021Plug} 
% & SwinIR%~\cite{liang2021swinir} 
% %& \small{Restormer~\cite{zamir2022restormer}}
% &Restormer
% %& \textbf{DMID-d (Ours)} 
% & \textbf{DMID-d} 
% & Clean
% \\
&   
%left bottom right top     
\includegraphics[trim={  110 191 311 70
 },clip,width=.105\textwidth,valign=t]{Images/classical/CBSD68/dncnn-157055_27.423282115199534.png} &   
\includegraphics[trim={  110 191 311 70
 },clip,width=.105\textwidth,valign=t]{Images/classical/CBSD68/ffdnet-157055_27.47242332695828.png} &   
\includegraphics[trim={ 110 191 311 70
 },clip,width=.105\textwidth,valign=t]{Images/classical/CBSD68/drunet-157055_28.080646953608248.png} &
\includegraphics[trim={  110 191 311 70
 },clip,width=.105\textwidth,valign=t]{Images/classical/CBSD68/157055_SwinIR_28.17} &  
\includegraphics[trim={  110 191 311 70
 },clip,width=.105\textwidth,valign=t]{Images/classical/CBSD68/Restormer-0028_28.17.png} &   
\includegraphics[trim={  110 191 311 70
 },clip,width=.105\textwidth,valign=t]{Images/classical/CBSD68/renoise-38_denoise.png} &
 \includegraphics[trim={  110 191 311 70
 },clip,width=.105\textwidth,valign=t]{Images/classical/CBSD68/157055.png}
\\
14.94 dB & 27.42 dB &  27.47 dB & 28.08 dB & \color{blue}{28.17 dB} & \color{blue}{28.17 dB} & \color{red}{28.39 dB} & PSNR
\\
Noisy 
& DnCNN%~\cite{2016Beyond} 
& FFDNet%~\cite{2017FFDNet} 
& DRUNet%~\cite{2021Plug} 
& SwinIR%~\cite{liang2021swinir} 
% & \small{Restormer~\cite{zamir2022restormer}}
&Restormer
%& \textbf{DMID-d (Ours)} 
& \textbf{DMID-d} 
%& \textbf{Ours}
& Clean 
 
\\

\multirow{3}{*}{\includegraphics[trim={ 134 6 134 6  },clip,width=.22\textwidth,valign=t]{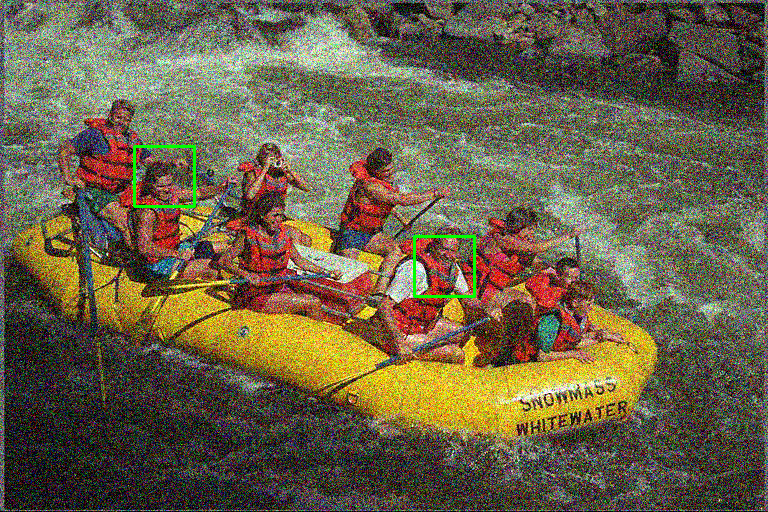}}&  
\includegraphics[trim={  0 300 440 140  },clip,width=.105\textwidth,valign=t]{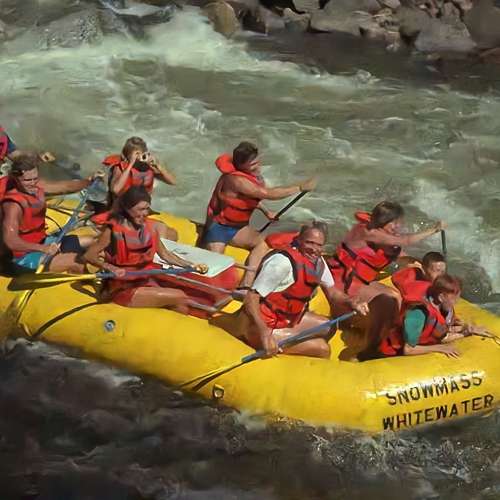} &
\includegraphics[trim={  0 300 440 140  },clip,width=.105\textwidth,valign=t]{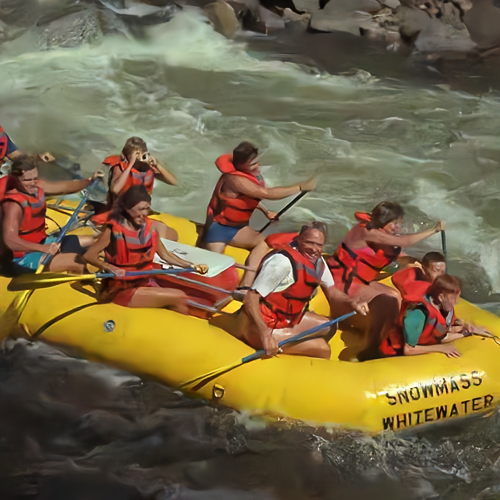} &
\includegraphics[trim={  0 300 440 140  },clip,width=.105\textwidth,valign=t]{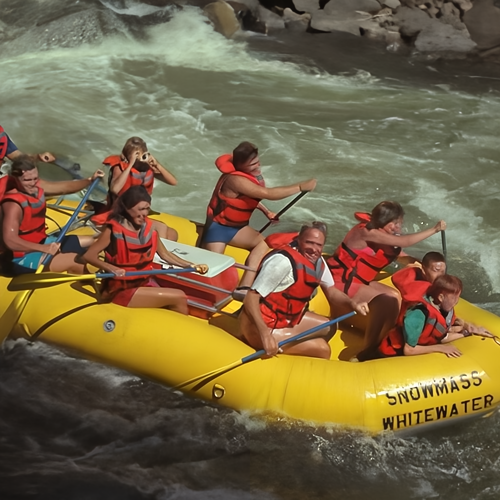} &
\includegraphics[trim={  0 300 440 140  },clip,width=.105\textwidth,valign=t]{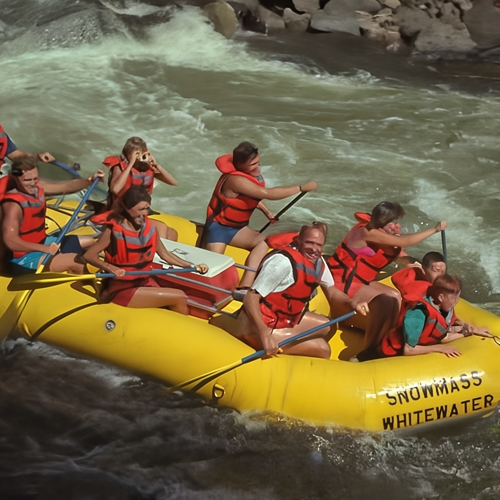} &
%134 6 134 6
\includegraphics[trim={  134 306 574 146 },clip,width=.105\textwidth,valign=t]{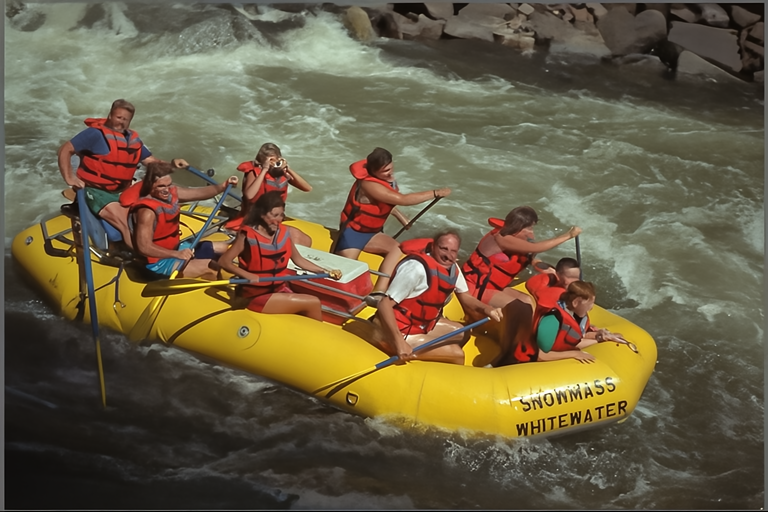} &
\includegraphics[trim={  134 306 574 146 },clip,width=.105\textwidth,valign=t]{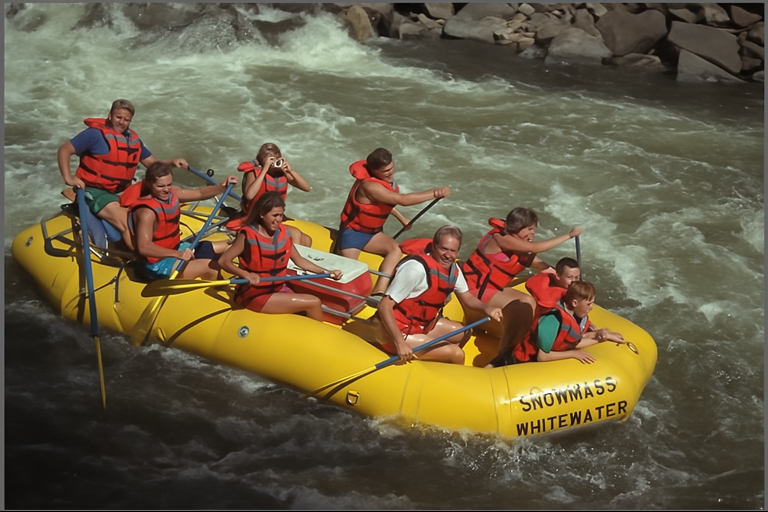} &
\includegraphics[trim={  0 300 440 140 },clip,width=.105\textwidth,valign=t]{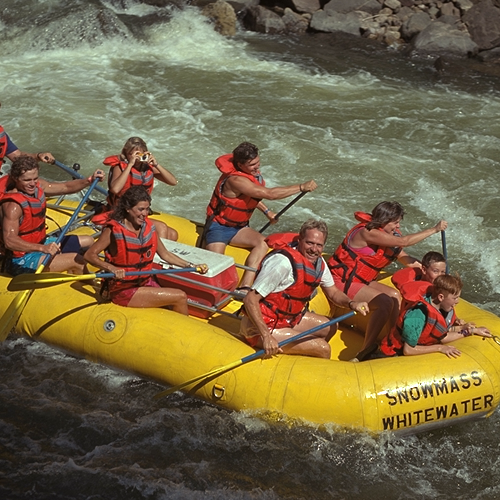} 
\\
\addlinespace[2pt]

% & DnCNN%~\cite{2016Beyond} 
% & FFDNet%~\cite{2017FFDNet} 
% & DRUNet%~\cite{2021Plug} 
% & SwinIR%~\cite{liang2021swinir} 
% % & \small{Restormer~\cite{zamir2022restormer}}
% &Restormer
% %& \textbf{DMID-d (Ours)} 
% & \textbf{DMID-d} 
% & Clean
% \\
&
%left bottom right top   
\includegraphics[trim={  280 210 160 230  },clip,width=.105\textwidth,valign=t]{Images/classical/Koak24/dncnn-kodim14_26.925091187867235.png} &
\includegraphics[trim={  280 210 160 230  },clip,width=.105\textwidth,valign=t]{Images/classical/Koak24/ffdnet-kodim14_26.99954384516321.png} &
\includegraphics[trim={  280 210 160 230  },clip,width=.105\textwidth,valign=t]{Images/classical/Koak24/drunet-kodim14_27.573876162581136.png} &
\includegraphics[trim={  280 210 160 230  },clip,width=.105\textwidth,valign=t]{Images/classical/Koak24/kodim14_SwinIR.png} &
%134 6 134 6
\includegraphics[trim={  414 216 294 236 },clip,width=.105\textwidth,valign=t]{Images/classical/Koak24/restormer-14.png} &
\includegraphics[trim={  414 216 294 236 },clip,width=.105\textwidth,valign=t]{Images/classical/Koak24/19_denoise.png} &
\includegraphics[trim={  280 210 160 230 },clip,width=.105\textwidth,valign=t]{Images/classical/Koak24/kodim14.png} 
\\
14.92 dB  & 26.93 dB &  27.00 dB & 27.57 dB & 27.65 dB & \color{blue}{28.12 dB} & \color{red}{28.26 dB} &PSNR
\\
Noisy 
& DnCNN%~\cite{2016Beyond} 
& FFDNet%~\cite{2017FFDNet} 
& DRUNet%~\cite{2021Plug} 
& SwinIR%~\cite{liang2021swinir} 
% & \small{Restormer~\cite{zamir2022restormer}}
&Restormer
%& \textbf{DMID-d (Ours)} 
& \textbf{DMID-d} 
& Clean
\\

\end{tabular}}
\end{center}
\vspace{-1em}
\caption{Visual results on classical Gaussian image denoising.
The images restored by our model exhibit more details and realism.  
	%Our method produces natural and realistic images.
}
\label{fig:SOTA}
%\vspace{-0.5em}

\end{figure*}

    It is imperative to highlight that our contribution is pioneering in multiple facets. 
    Firstly, we reframe the diffusion model as a Gaussian denoiser. This allows us to denoise in a single step or potentially optimize through multiple iterations. 
    Secondly, we elucidate the embedding method and the noise-correlated sampling starting point for image denoising. This not only reduces the required sampling times in one inference but also enhances the quality of the results.
    Finally, we present the elucidation of how sampling times $S_t$ and repetition times $R_t$ impact distortion-based and perception-based quality. This helps reduce distortion and lays the foundation for further research in image denoising area.

	%%%%%%%%%%%%%%%%%%%%%%%%%%%%%%%%%%%%%%%%%%%%%%%%%%%%%%%%%%

	\section{Experiments}
	%结果图-arbitrary

	\label{sec:experiments}
	% Our experiments are generally divided into three parts. First of all, we demonstrate our method achieves SOTA performance on distortion-based and perception-based metrics on Gaussian noise in Section \ref{sec:Gaussian denoising} and real-world noise in Section \ref{sec:realworld denoising}. In addition, we conduct ablation studies in Section \ref{sec:ablation_study}. Finally, we engage in a comparative analysis and extension to other diffusion-based image restoration approaches for image denoising in Section \ref{sec:compare and extend}.
     Our experiments are generally divided into three parts. 
     First of all, we demonstrate the performance of our method on Gaussian noise in Section \ref{sec:Gaussian denoising} and real-world noise in Section \ref{sec:realworld denoising}. 
     In addition, we conduct detailed ablation studies to fully evaluate our embedding and ensembling method in Section \ref{sec:ablation_study}. 
     Finally, we engage in a comparative analysis and extension to other diffusion-based methods in Section \ref{sec:compare and extend}.

	%测试的detail参数
	\subsection{Implementation Details}
	\label{sec:details}

	We employ a pre-trained model from~\cite{dhariwal2021diffusion} which is trained on 256$\times$256 images from ImageNet~\cite{russakovsky2015imagenet}, with full timesteps $T=1000$. Following \cite{dhariwal2021diffusion}, $\beta_t$ is defined increasing linearly from $0.0001$ to $0.02$ for $t$ from $1$ to $1000$, $\sigma_t$ is set to be $\gamma \sqrt{\frac{1-\bar{\alpha}_{t-1}}{1-\bar{\alpha}_{t}}}\sqrt{1-\frac{\bar{\alpha}_{t}}{\bar{\alpha}_{t-1}}}$, and $\gamma=0.85$. 
	
	% For various levels of noise, we first pre-calculate the value of $N$ and conduct inference. 
    Since our method can traverse through the perception-distortion
	curve~\cite{8578750,NEURIPS2019_6c29793a}, 
    % we present two variants of our method and at least one of them achieves SOTA results compared to previous methods in both distortion-based and perception-based metrics. 
    we present two variants of our method.
    The first variant, named “DMID-d”, achieves the least distortion and satisfactory perceptual quality. The second variant, named “DMID-p”, achieves the best perceptual quality and tolerable distortion. To achieve this, we need to determine the number of sampling times (referred to $S_t$) in an inference process and how many times will we repeat and get an average (referred to $R_t$). 
    % To ensure fairness, we set the full sampling times to be 1000 following \cite{ho2020denoising,lugmayr2022repaint}, that is $S_t*R_t=1000$ for the first variant.
    For the first variant, we set the full sampling times $S_t*R_t$ to be 1000 following \cite{ho2020denoising,lugmayr2022repaint} to ensure fairness.
    For the second variant, we set $R_t=1$ and the value of $S_t$ varies from 2 to 200 for different datasets and various noise levels. 
    As described in Section \ref{sec:ensembling}, decreasing the number of sampling times $S_t$ and increasing the number of repetition times $R_t$ will reduce distortion while conversely improving perceptual quality. This is why we introduced the two variations. Further explanations can be found in Section \ref{sec:ablation_study}.

	\subsection{Gaussian Image Denoising}
	\label{sec:Gaussian denoising}
	In this section, we conduct Gaussian image denoising experiments on synthetic benchmark datasets. 
	
	% Image denoising is generally evaluated on the distortion-based PSNR metric with noise levels $\sigma=15,25,50$, while our method can deal with larger noise levels on both perceptual and distortion-based metrics. Therefore we conduct classical Gaussian denoising experiments and robust Gaussian denoising experiments. 
	Image denoising is commonly evaluated using the distortion-based PSNR metric with noise levels $\sigma=15,25,50$. However, our method is robust to much higher noise levels, excelling in both distortion-based and perception-based metrics. Hence, we perform classical Gaussian denoising experiments as well as robust Gaussian denoising experiments.
 
	% For the classical Gaussian denoising, we evaluate our method on the distortion-based PSNR metric with noise levels $\sigma=15,25,50$ following previous classical comparision~\cite{zamir2022restormer,zhang2023accurate,zhao2023comprehensive}. 
	For classical Gaussian denoising, we evaluate our method using the distortion-based PSNR metric with noise levels $\sigma=15,25,50$, consistent with previous classical comparisons~\cite{zamir2022restormer,zhang2023accurate,zhao2023comprehensive}. 
	Table \ref{table:1-calssical} presents the PSNR scores achieved by various SOTA approaches on the synthetic benchmark datasets (CBSD68~\cite{martin2001database}, Kodak24~\cite{franzen1999kodak}, McMaster~\cite{zhang2011color}, Urban100~\cite{urban100}). All the results are reported from \cite{zamir2022restormer,zhang2023accurate}. The top row in Table \ref{table:1-calssical} is training a separate model for a specific noise level, the bottom row is the models that are designed to deal with various noise levels. The methods (such as Restormer~\cite{zamir2022restormer} and SwinIR~\cite{liang2021swinir}) that can only handle one noise level still perform worse than us. 
	%As shown in Figure \ref{fig:SOTA}, the two images are the hard cases for previous methods while our method can restore natural and realistic images. 
	As shown in Figure \ref{fig:SOTA}, these two images presented significant challenges for previous methods, with their denoised results showing highly unrealistic facial features. In contrast, our method successfully denoises these images to a natural and realistic quality, highlighting the effectiveness of our approach.

    For the robust Gaussian denoising, we evaluate our method using both distortion-based and perception-based metrics with larger noise levels. We compare our method with DnCNN~\cite{2016Beyond}, DRUNet~\cite{2021Plug}, Restormer~\cite{zamir2022restormer} and ART~\cite{zhang2023accurate}. Specifically, we evaluate different methods on ImageNet 1K (crop the center 256×256 image as input)~\cite{russakovsky2015imagenet}, CBSD68~\cite{martin2001database}, Kodak24~\cite{franzen1999kodak}, and McMaster~\cite{zhang2011color} datasets. In addition, we evaluate different methods on two representative distortion-based metrics (PSNR and SSIM~\cite{ssim}) and perception-based metric (LPIPS~\cite{zhang2018perceptual}). 
    Remarkably, our method is effective even when the standard deviation greatly exceeds 255, a previously unexplored capability that sets our method apart from other models. To ensure a fair comparison with other models, we evaluate different methods %on representative noise levels spanning from 0 to 255.
    across a series of representative noise levels, spanning from 0 to 255.

    %Next, we will provide a detailed explanation of the comparative method. 
    Subsequently, we will provide a detailed explanation of our comparative methods.
    DnCNN~\cite{2016Beyond} and DRUNet~\cite{2021Plug} are considered classical methods, while Restormer~\cite{zamir2022restormer} and ART~\cite{zhang2023accurate} represent the current state-of-the-art approaches. For DnCNN~\cite{2016Beyond} and DRUNet~\cite{2021Plug}, we retrain them following their specified training details. %Retraining %Restormer~\cite{zamir2022restormer} and ART~\cite{zhang2023accurate} is known to be costly and difficult, but their released pre-trained models can not deal with the noise level whose standard deviation is over 50. So following the approach of \cite{2021Plug}, we simply multiply by a constant to ensure that the standard deviation is 50.
    However, it is worth noting that retraining Restormer~\cite{zamir2022restormer} and ART~\cite{zhang2023accurate} is a resource-intensive process. 
    %Given that their publicly available pre-trained models are unable to handle noise levels with standard deviations over 50, we have opted to adopt an approach inspired by \cite{2021Plug}. 
    Their publicly available pre-trained models are not designed to handle noise levels with standard deviations exceeding 50. Therefore, 
    following the approach of \cite{2021Plug}, we multiply by a constant to ensure that the standard deviation is 50.
\begin{table*}[t!]
	\caption{Robust Gaussian image denoising. We report the results of method labeled "Ours-d" with least distortion, and a second method labeled “Ours-p” with greatest perceptual quality. Our method achieves SOTA performance on all metrics (PSNR↑ / SSIM↑ / LPIPS↓) and on all noise levels.}
	\label{table:2-arbitrary}
	\renewcommand{\arraystretch}{1.2}
	\setlength{\tabcolsep}{2pt}
	\scalebox{0.90}{
		\begin{tabular}{l| c| c| c | c |c |c |c}
			\toprule
			{\makebox[0.050\textwidth][c]{}} & 
			{\makebox[0.040\textwidth][c]{Noise}} & 
			{\makebox[0.085\textwidth][c]{DnCNN~\cite{2016Beyond}}} & 
%			{\makebox[0.085\textwidth][c]{FFDNet~\cite{2017FFDNet}}} & 
			{\makebox[0.085\textwidth][c]{DRUNet~\cite{2021Plug}}} & 
			{\makebox[0.085\textwidth][c]{Restormer~\cite{zamir2022restormer}}} & {\makebox[0.085\textwidth][c]{ART~\cite{zhang2023accurate}}} 
			&{\makebox[0.085\textwidth][c]{\textbf{DMID-d (Ours)}}} 
			&{\makebox[0.085\textwidth][c]{\textbf{DMID-p (Ours)}}}
			
			\\ 
			\cline{3-8} 
			\multirow{-2}{*}{Dataset} & Level 
			    & PSNR / SSIM / LPIPS
			    & PSNR / SSIM / LPIPS
			    & PSNR / SSIM / LPIPS
			    & PSNR / SSIM / LPIPS
			    & PSNR / SSIM / LPIPS
			    & PSNR / SSIM / LPIPS
			\\ 
			    
 			\midrule

			& $\sigma=50$
			& {\color{black}28.21} / {\color{black}0.8806} / {\color{black}0.179}
%			& {\color{black}28.74} / {\color{black}0.8941} / {\color{black}0.182}
			& {\color{black}29.46} / {\color{black}0.9081} / {\color{black}0.145}
			& {\color{black}29.61} / {\color{blue}0.9110} / {\color{black}0.136}
			& {\color{blue}29.62} / {\color{black}0.9105} / {\color{black}0.132}
			& {\color{red}29.90} / {\color{red}0.9157} / {\color{blue}0.114}
			&27.59 / 0.8722 / {\color{red}0.087}
			\\
			
			& $\sigma=100$
			& {\color{black}25.34} / {\color{black}0.8025} / {\color{black}0.293}
%			& {\color{black}25.75} / {\color{black}0.8228} / {\color{black}0.301}
			& {\color{black}26.48} / {\color{black}0.8482} / {\color{black}0.252}
			& {\color{blue}26.61} / {\color{blue}0.8532} / {\color{black}0.234}
			& {\color{black}26.57} / {\color{black}0.8501} / {\color{black}0.231}
			& {\color{red}27.00} / {\color{red}0.8626} / {\color{blue}0.201}
			&24.61 / 0.7987 / {\color{red}0.156}
			\\
			
			& $\sigma=150$
			& {\color{black}23.69} / {\color{black}0.7421} / {\color{black}0.370}
%			& {\color{black}24.06} / {\color{black}0.7667} / {\color{black}0.384}
			& {\color{black}24.81} / {\color{black}0.8032} / {\color{black}0.331}
			& {\color{blue}24.93} / {\color{blue}0.8103} / {\color{black}0.306}
			& {\color{black}24.88} / {\color{black}0.8043} / {\color{black}0.307}
			& {\color{red}25.39} / {\color{red}0.8236} / {\color{blue}0.263}
			&22.94 / 0.7465 / {\color{red}0.210}
			\\
			
			& $\sigma=200$
			& {\color{black}22.52} / {\color{black}0.6910} / {\color{black}0.432}
%			& {\color{black}22.87} / {\color{black}0.7177} / {\color{black}0.449}
			& {\color{black}23.66} / {\color{black}0.7673} / {\color{black}0.393}
			& {\color{blue}23.78} / {\color{blue}0.7762} / {\color{black}0.363}
			& {\color{black}23.72} / {\color{black}0.7674} / {\color{black}0.368}
			& {\color{red}24.25} / {\color{red}0.7915} / {\color{blue}0.295}
			&21.56 / 0.6932 / {\color{red}0.259}
			\\
			
			\multirow{-5}{*}{ImageNet}
			& $\sigma=250$
			& {\color{black}21.62} / {\color{black}0.6448} / {\color{black}0.478}
%			& {\color{black}21.94} / {\color{black}0.6742} / {\color{black}0.504}
			& {\color{black}22.80} / {\color{black}0.7374} / {\color{black}0.445}
			& {\color{blue}22.92} / {\color{blue}0.7479} / {\color{black}0.409}
			& {\color{black}22.84} / {\color{black}0.7363} / {\color{black}0.417}
			& {\color{red}23.44} / {\color{red}0.7675} / {\color{blue}0.346}
			&20.87 / 0.6701 / {\color{red}0.289}
			\\
			\midrule
			& $\sigma=50$
			& {\color{black}27.84} / {\color{black}0.8844} / {\color{black}0.226}
%			& {\color{black}27.97} / {\color{black}0.8873} / {\color{black}0.235}
			& {\color{black}28.51} / {\color{black}0.8991} / {\color{black}0.183}
			& {\color{black}28.59} / {\color{black}0.9011} / {\color{black}0.177}
			& {\color{blue}28.63} / {\color{blue}0.9015} / {\color{black}0.173}
			& {\color{red}28.69} / {\color{red}0.9029} / {\color{blue}0.162}
			&26.63 / 0.8605 / {\color{red}0.122}
			\\
			
			& $\sigma=100$
			& {\color{black}25.06} / {\color{black}0.8090} / {\color{black}0.365}
%			& {\color{black}25.19} / {\color{black}0.8146} / {\color{black}0.380}
			& {\color{black}25.76} / {\color{black}0.8357} / {\color{black}0.308}
			& {\color{blue}25.84} / {\color{blue}0.8389} / {\color{black}0.291}
			& {\color{black}25.86} / {\color{black}0.8376} / {\color{black}0.295}
			& {\color{red}25.96} / {\color{red}0.8413} / {\color{blue}0.283}
			&23.94 / 0.7840 / {\color{red}0.208}
			\\
			
			& $\sigma=150$
			& {\color{black}23.58} / {\color{black}0.7555} / {\color{black}0.449}
%			& {\color{black}23.71} / {\color{black}0.7629} / {\color{black}0.474}
			& {\color{black}24.32} / {\color{black}0.7932} / {\color{black}0.395}
			& {\color{blue}24.41} / {\color{blue}0.7980} / {\color{black}0.367}
			& {\color{black}24.41} / {\color{black}0.7945} / {\color{black}0.380}
			& {\color{red}24.54} / {\color{red}0.8001} / {\color{blue}0.361}
			&22.47 / 0.7314 / {\color{red}0.264}
			\\
			
			& $\sigma=200$
			& {\color{black}22.56} / {\color{black}0.7131} / {\color{black}0.514}
%			& {\color{black}22.71} / {\color{black}0.7211} / {\color{black}0.543}
			& {\color{black}23.36} / {\color{black}0.7616} / {\color{black}0.464}
			& {\color{blue}23.47} / {\color{blue}0.7682} / {\color{black}0.426}
			& {\color{black}23.44} / {\color{black}0.7623} / {\color{black}0.447}
			& {\color{red}23.57} / {\color{red}0.7686} / {\color{blue}0.392}
			&21.37 / 0.6842 / {\color{red}0.312}
			\\
			
			\multirow{-5}{*}{CBSD68}
			& $\sigma=250$
			& {\color{black}21.80} / {\color{black}0.6773} / {\color{black}0.558}
%			& {\color{black}21.93} / {\color{black}0.6834} / {\color{black}0.598}
			& {\color{black}22.64} / {\color{black}0.7368} / {\color{black}0.519}
			& {\color{blue}22.77} / {\color{blue}0.7451} / {\color{black}0.475}
			& {\color{black}22.72} / {\color{black}0.7364} / {\color{black}0.500}
			& {\color{red}22.88} / {\color{red}0.7462} / {\color{blue}0.455}
			&20.77 / 0.6610 / {\color{red}0.352}
			\\
			\midrule
			& $\sigma=50$
			& {\color{black}28.84} / {\color{black}0.8921} / {\color{black}0.247}
%			& {\color{black}29.10} / {\color{black}0.8980} / {\color{black}0.248}
			& {\color{black}29.86} / {\color{black}0.9132} / {\color{black}0.188}
			& {\color{black}30.00} / {\color{blue}0.9153} / {\color{black}0.185}
			& {\color{blue}30.02} / {\color{black}0.9152} / {\color{black}0.181}
			& {\color{red}30.13} / {\color{red}0.9174} / {\color{blue}0.172}
			&27.90 / 0.8770  / {\color{red}0.131}
			\\
			
			& $\sigma=100$
			& {\color{black}26.00} / {\color{black}0.8206} / {\color{black}0.387}
%			& {\color{black}26.28} / {\color{black}0.8305} / {\color{black}0.394}
			& {\color{black}27.16} / {\color{black}0.8609} / {\color{black}0.297}
			& {\color{blue}27.30} / {\color{blue}0.8642} / {\color{black}0.287}
			& {\color{black}27.27} / {\color{black}0.8616} / {\color{black}0.289}
			& {\color{red}27.50} / {\color{red}0.8682} / {\color{blue}0.274}
			&25.31 / 0.8107  / {\color{red}0.211}
			\\
			
			& $\sigma=150$
			& {\color{black}24.41} / {\color{black}0.7670} / {\color{black}0.475}
%			& {\color{black}24.67} / {\color{black}0.7792} / {\color{black}0.496}
			& {\color{black}25.69} / {\color{black}0.8233} / {\color{black}0.381}
			& {\color{blue}25.85} / {\color{blue}0.8281} / {\color{black}0.363}
			& {\color{black}25.78} / {\color{black}0.8219} / {\color{black}0.372}
			& {\color{red}26.08} / {\color{red}0.8336} / {\color{blue}0.348}
			&24.06 / 0.7707  / {\color{red}0.275}
			\\
			
			& $\sigma=200$
			& {\color{black}23.33} / {\color{black}0.7228} / {\color{black}0.543}
%			& {\color{black}23.59} / {\color{black}0.7394} / {\color{black}0.572}
			& {\color{black}24.65} / {\color{black}0.7930} / {\color{black}0.450}
			& {\color{blue}24.84} / {\color{blue}0.7999} / {\color{black}0.422}
			& {\color{black}24.74} / {\color{black}0.7901} / {\color{black}0.439}
			& {\color{red}25.08} / {\color{red}0.8054} / {\color{blue}0.382}
			&22.99 / 0.7311  / {\color{red}0.318}
			\\
			
			\multirow{-5}{*}{Kodak24}
			& $\sigma=250$
			& {\color{black}22.51} / {\color{black}0.6830} / {\color{black}0.589}
%			& {\color{black}22.79} / {\color{black}0.7067} / {\color{black}0.630}
			& {\color{black}23.89} / {\color{black}0.7686} / {\color{black}0.504}
			& {\color{blue}24.09} / {\color{blue}0.7772} / {\color{black}0.469}
			& {\color{black}23.94} / {\color{black}0.7631} / {\color{black}0.491}
			& {\color{red}24.40} / {\color{red}0.7853} / {\color{blue}0.446}
			&22.44 / 0.7133  / {\color{red}0.356}
			\\
			\midrule
			& $\sigma=50$
			& {\color{black}28.35} / {\color{black}0.9078} / {\color{black}0.180}
%			& {\color{black}29.15} / {\color{black}0.9222} / {\color{black}0.178}
			& {\color{black}30.04} / {\color{black}0.9350} / {\color{black}0.140}
			& {\color{black}30.29} / {\color{blue}0.9378} / {\color{black}0.134}
			& {\color{blue}30.31} / {\color{blue}0.9378} / {\color{black}0.132}
			& {\color{red}30.51} / {\color{red}0.9396} / {\color{blue}0.124}
			&28.34 / 0.9112 / {\color{red}0.092}
			\\
			
			& $\sigma=100$
			& {\color{black}25.59} / {\color{black}0.8499} / {\color{black}0.283}
%			& {\color{black}26.14} / {\color{black}0.8674} / {\color{black}0.283}
			& {\color{black}27.05} / {\color{black}0.8914} / {\color{black}0.233}
			& {\color{blue}27.25} / {\color{blue}0.8962} / {\color{black}0.220}
			& {\color{black}27.22} / {\color{black}0.8945} / {\color{black}0.218}
			& {\color{red}27.57} / {\color{red}0.8990} / {\color{blue}0.206}
			&25.37 / 0.8560 / {\color{red}0.158}
			\\
			
			& $\sigma=150$
			& {\color{black}23.98} / {\color{black}0.8041} / {\color{black}0.360}
%			& {\color{black}24.41} / {\color{black}0.8235} / {\color{black}0.365}
			& {\color{black}25.36} / {\color{black}0.8570} / {\color{black}0.300}
			& {\color{blue}25.57} / {\color{blue}0.8645} / {\color{black}0.279}
			& {\color{black}25.51} / {\color{black}0.8604} / {\color{black}0.281}
			& {\color{red}25.89} / {\color{red}0.8672} / {\color{blue}0.267}
			&23.72 / 0.8148 / {\color{red}0.209}
			\\
			
			& $\sigma=200$
			& {\color{black}22.85} / {\color{black}0.7646} / {\color{black}0.423}
%			& {\color{black}23.20} / {\color{black}0.7828} / {\color{black}0.433}
			& {\color{black}24.20} / {\color{black}0.8279} / {\color{black}0.354}
			& {\color{blue}24.41} / {\color{blue}0.8384} / {\color{black}0.327}
			& {\color{black}24.33} / {\color{black}0.8321} / {\color{black}0.334}
			& {\color{red}24.68} / {\color{red}0.8392} / {\color{blue}0.303}
			&22.51 / 0.7816 / {\color{red}0.252}
			\\
			
			\multirow{-5}{*}{McMaster}
			& $\sigma=250$
			& {\color{black}21.96} / {\color{black}0.7279} / {\color{black}0.469}
%			& {\color{black}22.24} / {\color{black}0.7449} / {\color{black}0.491}
			& {\color{black}23.31} / {\color{black}0.8031} / {\color{black}0.402}
			& {\color{blue}23.54} / {\color{blue}0.8161} / {\color{black}0.367}
			& {\color{black}23.43} / {\color{black}0.8077} / {\color{black}0.380}
			& {\color{red}23.81} / {\color{red}0.8174} / {\color{blue}0.359}
			&21.73 / 0.7553 / {\color{red}0.290}
			\\	
			
			\bottomrule %[0.1em]
			
		\end{tabular}%
		
	}
\end{table*}

	\begin{figure*}[!t]
% \begin{figure*}[H]
\footnotesize
\renewcommand{\arraystretch}{1.2}
\begin{center}
\setlength{\tabcolsep}{1.5pt}
\scalebox{0.97}{
\begin{tabular}[b]{c c c c c c c c}
% \hspace{-4mm}

Noisy  
& DnCNN%~\cite{2016Beyond}  
& DRUNet%~\cite{2021Plug}  
& Restormer%~\cite{zamir2022restormer} 
& ART%~\cite{zhang2023accurate}
% & \textbf{DMID-d (ours)} 
% & \textbf{DMID-p (ours)}
& \textbf{DMID-d} 
& \textbf{DMID-p} 
&Clean\\

\includegraphics[width=.122\textwidth,valign=t]{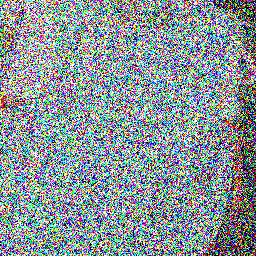} &   
\includegraphics[width=.122\textwidth,valign=t]{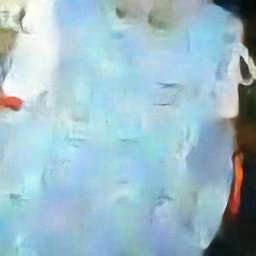} &

\includegraphics[width=.122\textwidth,valign=t]{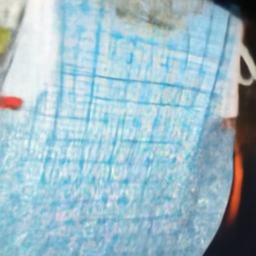} &  
\includegraphics[width=.122\textwidth,valign=t]{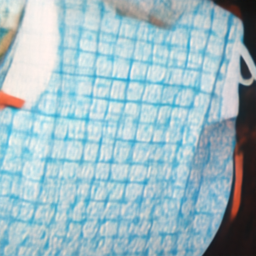} &  
\includegraphics[width=.122\textwidth,valign=t]{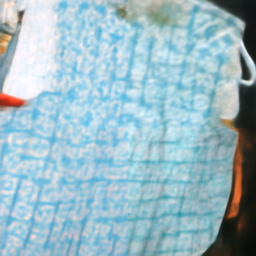} & 
\includegraphics[width=.122\textwidth,valign=t]{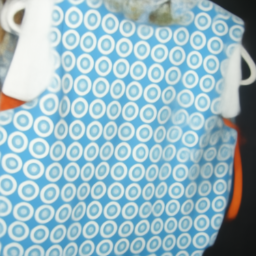} &  
\includegraphics[width=.122\textwidth,valign=t]{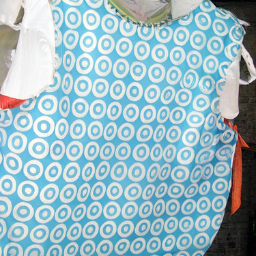}&
\includegraphics[width=.122\textwidth,valign=t]{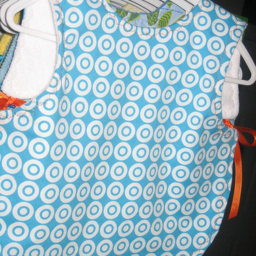} 
\\

% \small~  6.78 / 0.9750  & \small~17.77 / 0.7224 & \small~17.98 / 0.7261 &  \small~18.89 \color{black}{/ 0.4943} & \small~19.21 / 0.4621 & \small~\color{red}{22.26} \color{black}{/} \color{blue}{0.2623} & \small~ \color{blue}{20.25} \color{black}{/} \color{red}{0.1303} & \small~PSNR↑ / LPIPS↓
6.78 / 0.9750  & 17.77 / 0.7224  & 18.89 \color{black}{/ 0.4943} & 19.21 / 0.4621 & 19.04 / 0.4553 & \color{red}{22.26} \color{black}{/} \color{blue}{0.2623} &  \color{blue}{20.25} \color{black}{/} \color{red}{0.1303} & PSNR↑ / LPIPS↓  
\\

% \includegraphics[width=.122\textwidth,valign=t]{Images/arbitrary/2_noise.png} &   
% \includegraphics[width=.122\textwidth,valign=t]{Images/arbitrary/2_clean.png} & 
% \includegraphics[width=.122\textwidth,valign=t]{Images/arbitrary/dncnn_2_23.49060016174143.png} &  

% \includegraphics[width=.122\textwidth,valign=t]{Images/arbitrary/ffdnet_2_24.059444969462277.png} &     

% \includegraphics[width=.122\textwidth,valign=t]{Images/arbitrary/drunet_2_24.486905284656654.png} &  
% \includegraphics[width=.122\textwidth,valign=t]{Images/arbitrary/restormer_2_24.44.png} &  

% \includegraphics[width=.122\textwidth,valign=t]{Images/arbitrary/2_denoise_25.01.png} &  
% \includegraphics[width=.122\textwidth,valign=t]{Images/arbitrary/arbitrary2_denoise.png}  
% \\

% \small~ 7.04 / 1.4766 & \small~PSNR↑ / LPIPS↓  & \small~23.49 / 0.5094 & \small~24.06 / 0.5538 &  \small~\color{blue}{24.49} \color{black}{/ 0.5987} & \small~24.44 / 0.5582 & \small~\color{red}{25.01} \color{black}{/} \color{blue}{0.4916} & \small~ 24.27 / \color{red}{0.2685}
% \\

 %left bottom right top
\includegraphics[width=.122\textwidth,valign=t]{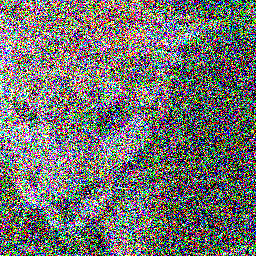}  & 
\includegraphics[width=.122\textwidth,valign=t]{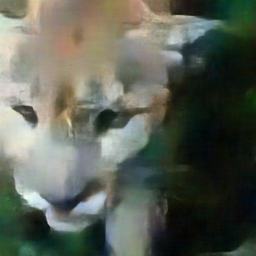} &  

\includegraphics[width=.122\textwidth,valign=t]{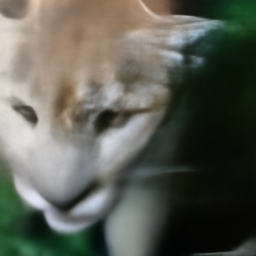} &  
\includegraphics[width=.122\textwidth,valign=t]{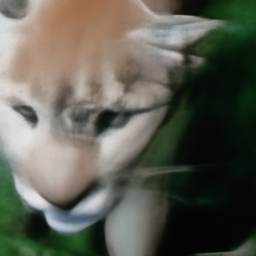} &  
\includegraphics[width=.122\textwidth,valign=t]{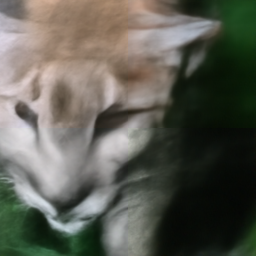} &

\includegraphics[width=.122\textwidth,valign=t]{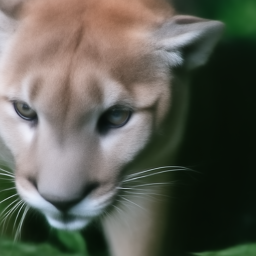} &  
\includegraphics[width=.122\textwidth,valign=t]{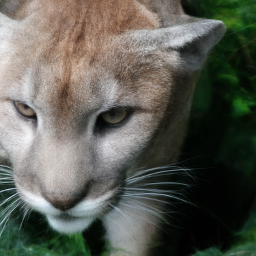}  
&   
\includegraphics[width=.122\textwidth,valign=t]{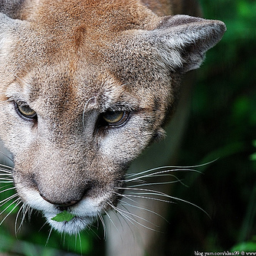}
\\
% \small~ 6.80 / 1.4129  & \small~20.83 / 0.6295 & \small~20.88 / 0.6912 &  \small~\color{blue}{21.24}\color{black}{ / 0.6822} & \small~21.21 / 0.6357 & \small~\color{red}{21.75 } \color{black}{/} \color{blue}{0.5143} & \small~ 21.04 / \color{red}{0.3180}  & \small~PSNR↑ / LPIPS↓ 
 6.80 / 1.4129  & 20.83 / 0.6295  &  \color{blue}{21.24}\color{black}{ / 0.6822} & 21.21 / 0.6357 & 21.20 / 0.6746 & \color{red}{21.75 } \color{black}{/} \color{blue}{0.5143} &  21.04 / \color{red}{0.3180}  & PSNR↑ / LPIPS↓ 

\\
 %left bottom right top
\includegraphics[width=.122\textwidth,valign=t]{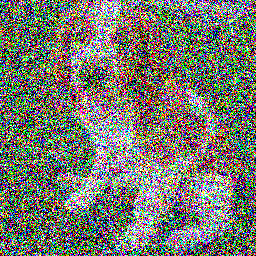} & 
\includegraphics[width=.122\textwidth,valign=t]{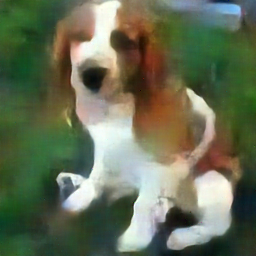} &  

\includegraphics[width=.122\textwidth,valign=t]{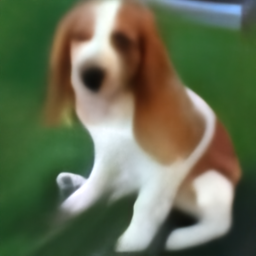} &  
\includegraphics[width=.122\textwidth,valign=t]{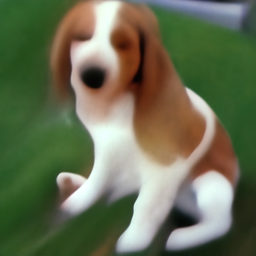} &  

\includegraphics[width=.122\textwidth,valign=t]{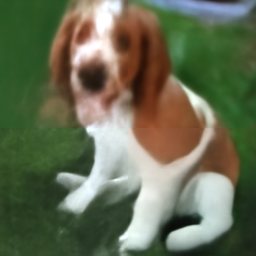} &

\includegraphics[width=.122\textwidth,valign=t]{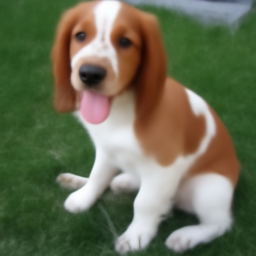} &  
\includegraphics[width=.122\textwidth,valign=t]{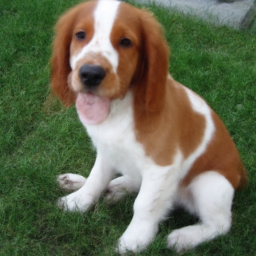}  
&   
\includegraphics[width=.122\textwidth,valign=t]{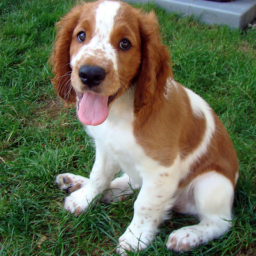} 
\\
 6.88 / 1.3411   & 21.26 / 0.4614  &  21.99 / 0.5033 & \color{blue}{22.21}\color{black}{ / 0.4903} & 22.10 / 0.4684 & \color{red}{22.71 }\color{black}{ /} \color{blue}{0.3659} &  22.06 / \color{red}{0.1908} & PSNR↑ / LPIPS↓
\\

\end{tabular}}
\end{center}
%\vspace*{-2mm}
\vspace{-1em}
\caption{Visual results on robust Gaussian image denoising. Our method can generate detailed texture, while other models even have severe chromatic aberration and blur.
}
\label{fig:arbitrary}
%\vspace{-1em}
\end{figure*}

\begin{table*}[t!]
\vspace{-1em}
 \centering
	\caption{Real-world image denoising. Our method achieves excellent performance across all metrics (PSNR↑ / SSIM↑ / LPIPS↓) when compared to both supervised (the top row) and unsupervised methods (the bottom row). The best and second-best methods are in {\color{red}red} and {\color{blue}blue}.} 
	\label{table:1-real}
	\renewcommand{\arraystretch}{1.2}
	\setlength{\tabcolsep}{8pt}
	\scalebox{1}{
		\begin{tabular}{l| c c c| c c c |c c c }
			\toprule
			
		{\makebox[0.13\textwidth][c]{}}
			& \multicolumn{3}{c|}{CC~\cite{2016A}}
			& \multicolumn{3}{c|}{PolyU~\cite{xu2018real}} 
			& \multicolumn{3}{c}{FMDD~\cite{zhang2018poisson}}
			
			\\ 
			\cline{2-10} 
			
			\multirow{-2}{*}{Method}  
			& {\makebox[0.06\textwidth][c]{PSNR}} 
			& {\makebox[0.06\textwidth][c]{SSIM}}
			& {\makebox[0.06\textwidth][c]{LPIPS}}
			& {\makebox[0.06\textwidth][c]{PSNR}} 
			& {\makebox[0.06\textwidth][c]{SSIM}} 
			& {\makebox[0.06\textwidth][c]{LPIPS}}
			& {\makebox[0.06\textwidth][c]{PSNR}} 
			& {\makebox[0.06\textwidth][c]{SSIM}} 
			& {\makebox[0.06\textwidth][c]{LPIPS}}
			\\ 
			\midrule
			
			DANet$_+$~\cite{ECCV2020_984}
			&35.91 &0.9816 &\color{blue}{0.073}
			&37.23 &0.9796 &0.088
			&31.59 &0.7962 &0.238
                % &39.47 &0.957- &-
			\\
			MIRNet~\cite{MIR}
			&36.04 &0.9797 &0.088
			&37.45 &0.9783 &0.101
			&31.83 &0.8106 &0.227
			% &39.72 &0.9579 &0.202
			\\
			MPRNet~\cite{zamir2021multi}
			&36.20 &0.9769 &0.094
			&37.43 &0.9759 &0.109
			&31.21 &0.7915 &0.292
                % &39.71 &0.9581 &0.203
			\\
			     DeamNet~\cite{ren2021adaptivedeamnet}
			&35.64 &0.9652 &0.089
			&32.46 &0.8615 &0.130
			&32.28 &0.8116 &0.261
			% &39.47 &0.957- &-
			\\
			NAFNet~\cite{chen2022simple} 
			&34.39 &0.9784 &\color{blue}{0.073}
			&36.04 &0.9607 &0.107
			&26.27 &0.6519 &0.340
			% &\color{red}{40.30} &\color{red}{0.9609} &\color{red}{0.188}
			\\
			Uformer~\cite{wang2022uformer}
			&36.31 &0.9795 &0.085
			&37.31 &0.9782 &0.096
			&31.81 &0.8118 &\color{red}{0.207}
                % &39.81 &0.9590 &0.200

			\\
			Restormer~\cite{zamir2022restormer}
			&36.27 &0.9810 &0.077
			&37.51 &0.9776 &0.102
			&31.81 &0.8046 & \color{blue}{0.211}
                % &\color{blue}{40.02} &\color{blue}{0.9598}  &\color{blue}{0.198}
			
			\\
			\midrule
                 
                \midrule
			N2V~\cite{noise2void}
			&32.27 & 0.862- & -
			&33.83 & 0.873- & -
			&- & - & -
                % &- & - & -

			\\
			N2S~\cite{noise2self}
			&33.38 & 0.846- & -
			&35.04 & 0.902- & -
			&- & - & -
                % &- & - & -
                 \\
                S2S~\cite{self2self}
%			 &35.27 & 0.143 & 67.8
			 &{\color{blue}37.52} & 0.951- & -
			 &38.37 & 0.962- & -
			 &30.76 & 0.695- & -
			 % &29.46 & 0.595- & -

			\\
			AP-BSN~\cite{lee2022apbsn}
			&34.86 &0.9744 &0.131
			&36.45 &0.9750 &0.099
			&32.40 &0.8461 &0.335
                %&35.91 &0.9278 &0.290
                % &34.95 &0.9169  & 0.360
			\\
            R2R~\cite{Recorrupted2Recorrupted}
			&33.43 &0.9564 &0.227
			&36.23 &0.9655 &0.151
			&27.17 &0.5250 &0.448
			% &			
			\\
   
			LG-BPN~\cite{wang2023lgbpn}
			&34.58 &0.9755 &0.135
			&36.59 &0.9763 &0.102
			&\color{blue}{33.12} &\color{blue}{0.8668} &0.283
                % &35.26 &0.9203 &0.361
			\\
			% \cline{2-13} 

			ZS-N2N~\cite{ZSNoise2Noise}

			&33.51 &0.9571 &0.224
			&35.99 &0.9587 &0.197
			&31.65 &0.7674 &0.222
                % &25.59 &0.5648 &0.657
			\\
                ScoreDVI~\cite{SocreDVI}
                &37.09 &0.945- & -
                &37.77 &0.959- & -
                &33.10 &0.865- & -
                % &34.75 &0.856- & -
                \\
			
			 \textbf{DMID-d (Ours)}
                %利用clean的结果：
			% & \color{red}{38.36} & \color{red}{0.9892}  & \color{black}{0.075}
			% &  \color{red}{38.93} & \color{red}{0.9861} & \color{blue}{0.063}
			% &  \color{red}{33.92} & \color{red}{0.8982} & 0.237
			%SURE的结果：
                & \color{red}{37.99} & \color{red}{0.9880}  & \color{black}{0.078}
			&  \color{red}{38.62} & \color{red}{0.9853} & \color{blue}{0.069}
			&  \color{red}{33.40} & \color{red}{0.8747} & 0.266
			\\
			\textbf{DMID-p (Ours)}
                %利用clean的结果：
			% &\color{blue}{38.12} & \color{blue}{0.9887}  & \color{red}{0.072}
			% &\color{blue}{38.79} & \color{blue}{0.9853}  & \color{red}{0.059}
			% &\color{blue}{33.65} & \color{blue}{0.8774}  & \color{black}{0.224}
			% &35.21 &0.9313 &0.257
                %SURE的结果：
                &\color{black}{37.09} & \color{blue}{0.9854}  & \color{red}{0.072}
			&\color{blue}{38.46} & \color{blue}{0.9843}  & \color{red}{0.067}
			&\color{black}{33.09} & \color{black}{0.8616}  & \color{black}{0.232}

			\\
			\bottomrule %[0.1em]
			
		\end{tabular}%
		
	}
\vspace{-1em}
\end{table*}
	%\vspace{-1em}\end{table*}

     \begin{figure*}[!t]
\footnotesize
\renewcommand{\arraystretch}{1.2}
\begin{center}
\setlength{\tabcolsep}{1.5pt}
\scalebox{0.97}{
\begin{tabular}[b]{c c c c c c c c}
% \hspace{-4mm}

\multirow{4}{*}{\includegraphics[trim={0 0 0 0},clip,width=.245\textwidth,valign=t]{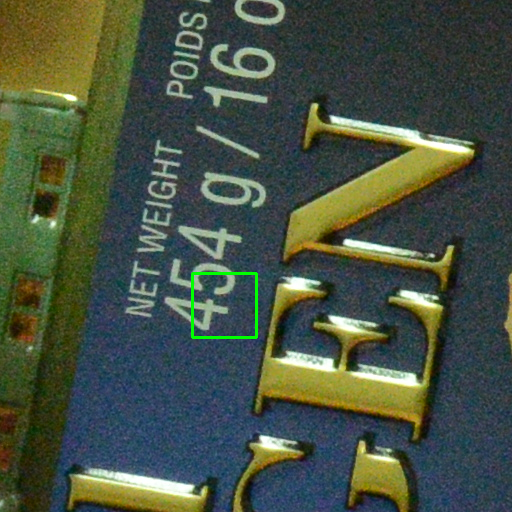}}&

\includegraphics[trim={192 175 256 273},clip,width=.1\textwidth,valign=t]{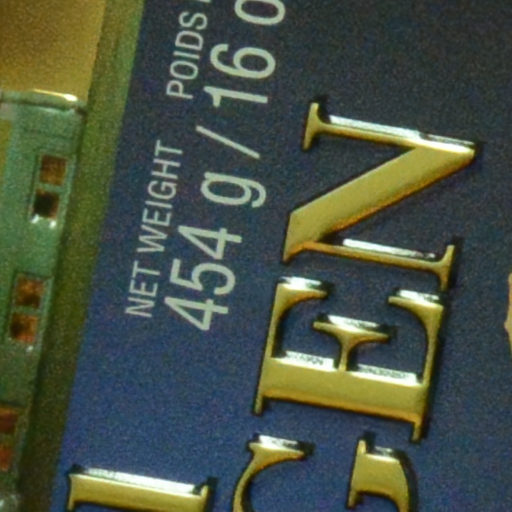} & 
\includegraphics[trim={192 175 256 273},clip,width=.1\textwidth,valign=t]{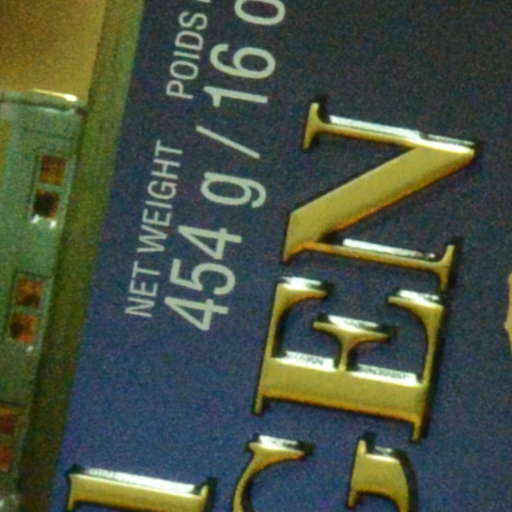} & 
\includegraphics[trim={192 175 256 273},clip,width=.1\textwidth,valign=t]{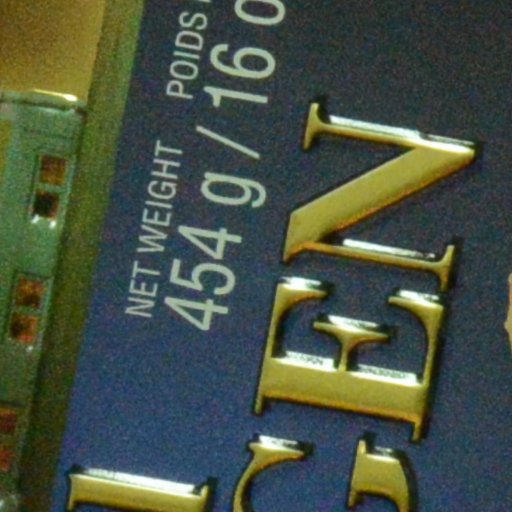} & 
\includegraphics[trim={192 175 256 273},clip,width=.1\textwidth,valign=t]{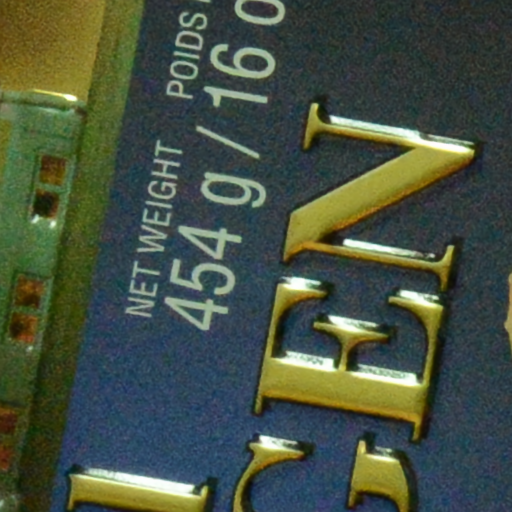} &  
\includegraphics[trim={192 175 256 273},clip,width=.1\textwidth,valign=t]{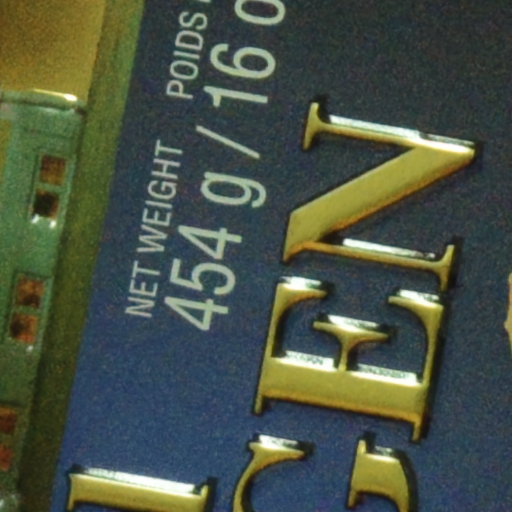} & 
\includegraphics[trim={192 175 256 273},clip,width=.1\textwidth,valign=t]{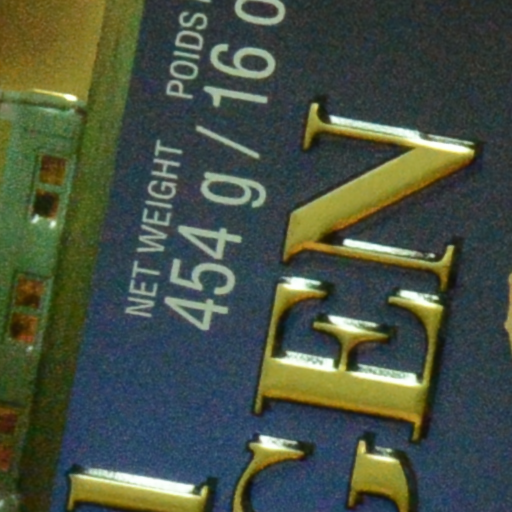} & 
\includegraphics[trim={192 175 256 273},clip,width=.1\textwidth,valign=t]{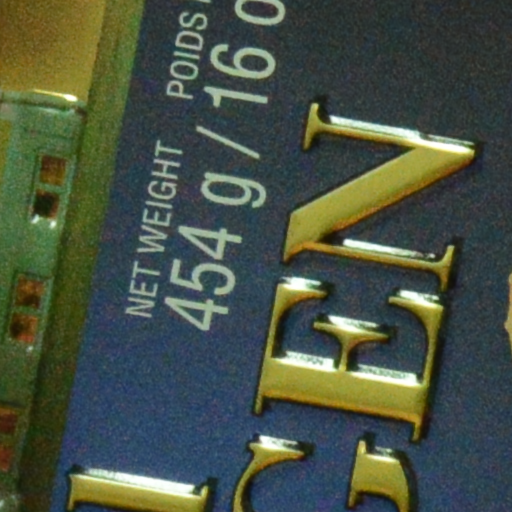} 
\\
& 32.38 / 0.1532
& 31.94 / 0.1885
& 31.94 / 0.1946
& 32.20 / 0.1917
& 31.66 / {\color{blue}0.1457}
& 32.21 / 0.1808
& 32.02 / 0.1752
\\

&DANet$_+$%~\cite{ECCV2020_984}
&MIRNet%~\cite{MIR}
&MPRNet%~\cite{zamir2021multi}
&DeamNet%~\cite{ren2021adaptivedeamnet}
&NAFNet%~\cite{chen2022simple} 
&Uformer%~\cite{wang2022uformer}
% &\small{Restormer$_g$~\cite{zamir2022restormer}}
&Restormer
\\
&
%left bottom right top

\includegraphics[trim={192 175 256 273},clip,width=.1\textwidth,valign=t]{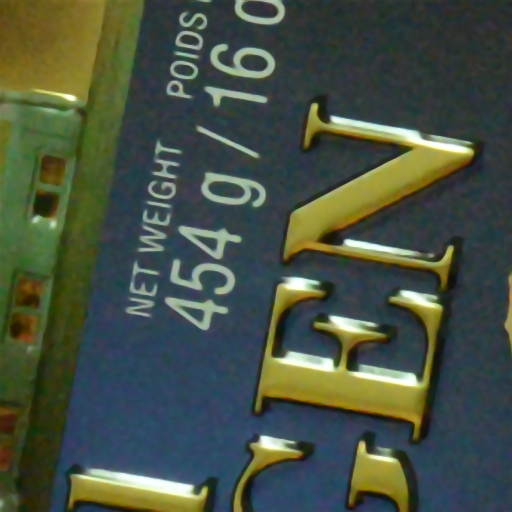} & 
 \includegraphics[trim={192 175 256 273},clip,width=.1\textwidth,valign=t]{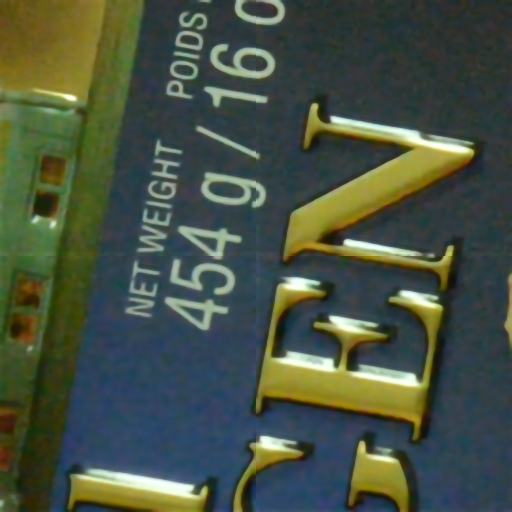} & 
\includegraphics[trim={192 175 256 273},clip,width=.1\textwidth,valign=t]{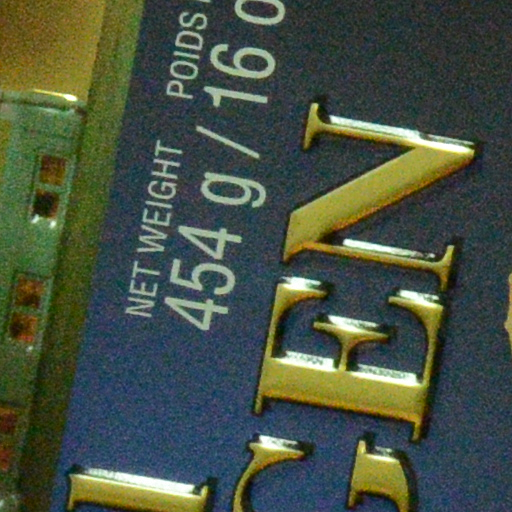} & 
\includegraphics[trim={192 175 256 273},clip,width=.1\textwidth,valign=t]{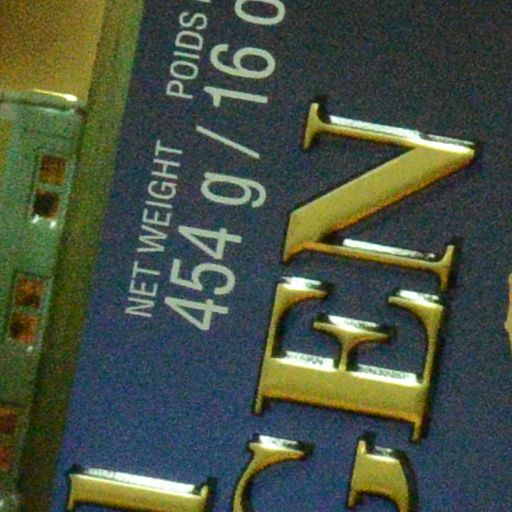} & 
\includegraphics[trim={192 175 256 273},clip,width=.1\textwidth,valign=t]{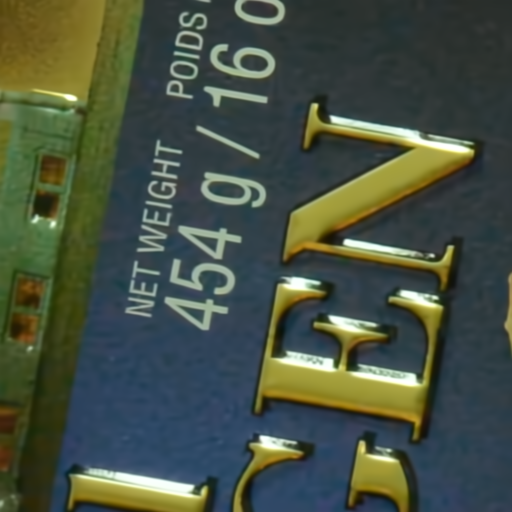} & 
\includegraphics[trim={192 175 256 273},clip,width=.1\textwidth,valign=t]{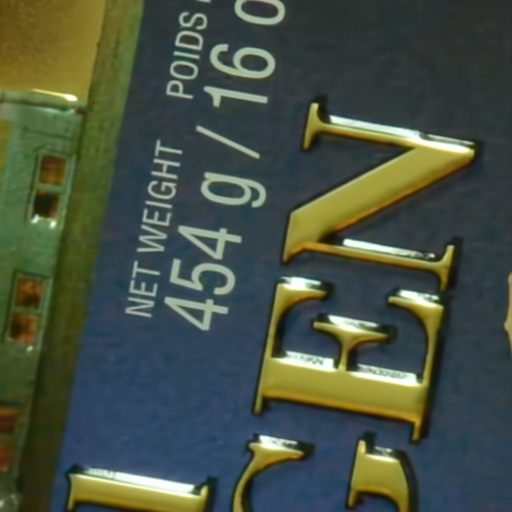} & 
\includegraphics[trim={192 175 256 273},clip,width=.1\textwidth,valign=t]{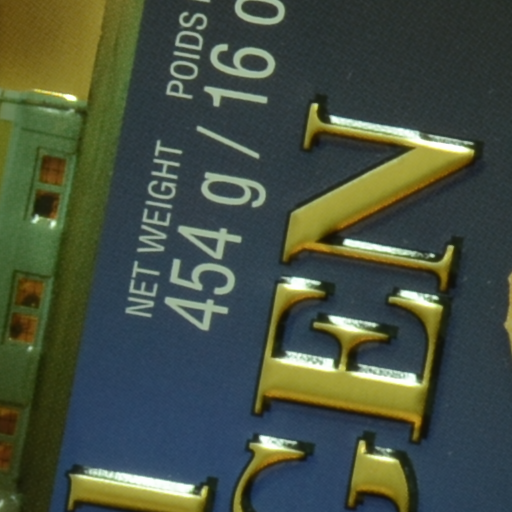} 
\\

29.63 / 0.2868
&32.02 / 0.2128
& 32.08 / 0.2169
&29.65 / 0.2871
&29.77 / 0.2789
&{\color{red}35.11 /}{ \color{black}0.1624}
&{\color{blue}34.84 /}{ \color{red} 0.1235}
&PSNR / LPIPS
\\
% &\small{Restormer$_r$~\cite{zamir2022restormer}}

Noisy

&AP-BSN%~\cite{lee2022apbsn}
&LG-BPN%~\cite{wang2023lgbpn}
&R2R%~\cite{Recorrupted2Recorrupted}
&ZS-N2N%~\cite{ZSNoise2Noise} 
% & \textbf{DMID-d (Ours)}
% & \textbf{DMID-p (Ours)} 
& \textbf{DMID-d}
& \textbf{DMID-p} 
&Clean
\\

%%%%%polyu3

\multirow{4}{*}{\includegraphics[trim={0 0 0 0},clip,width=.245\textwidth,valign=t]{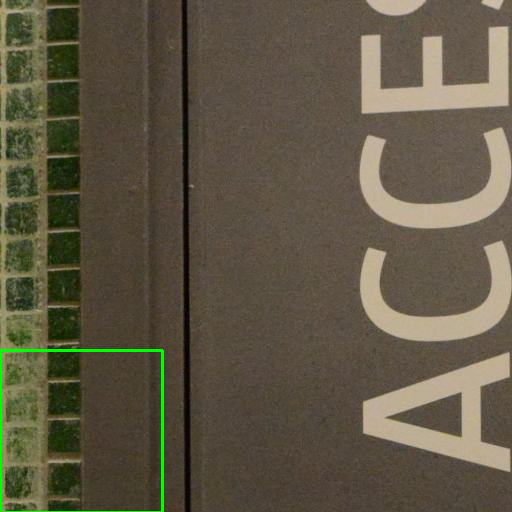}}&
\includegraphics[trim={0 0 350 350},clip,width=.1\textwidth,valign=t]{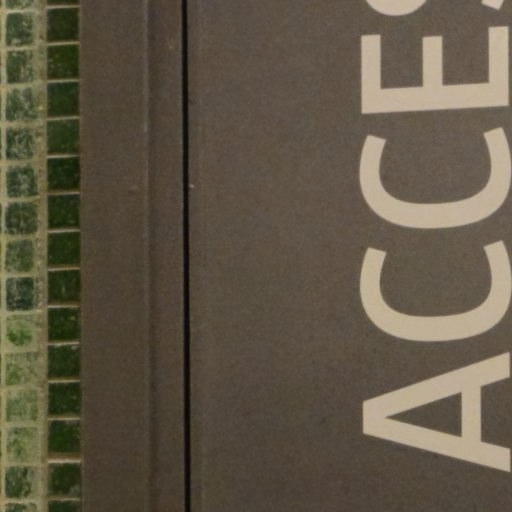} & 
\includegraphics[trim={0 0 350 350},clip,width=.1\textwidth,valign=t]{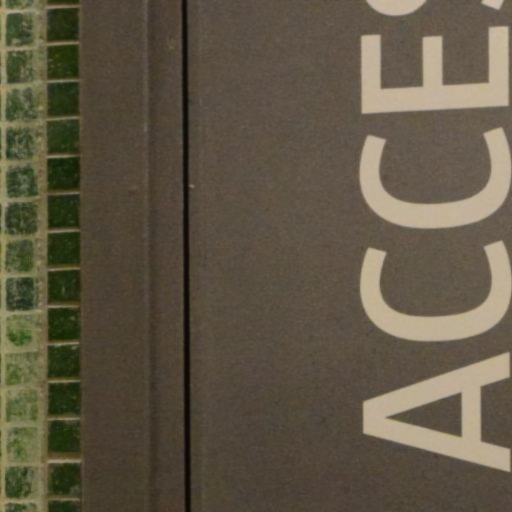} & 
\includegraphics[trim={0 0 350 350},clip,width=.1\textwidth,valign=t]{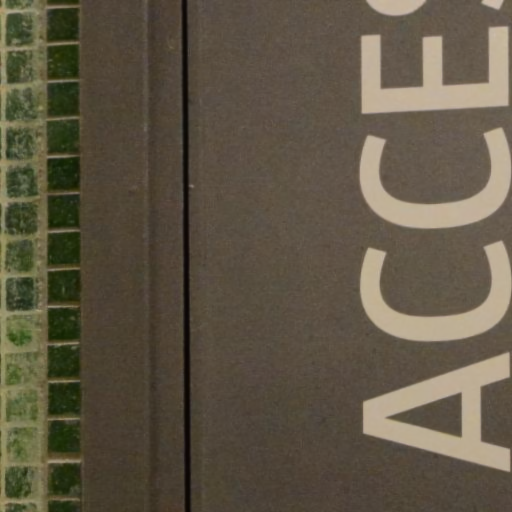} & 
\includegraphics[trim={0 0 350 350},clip,width=.1\textwidth,valign=t]{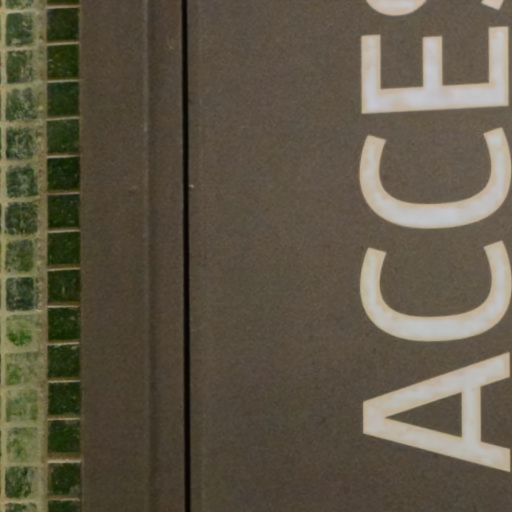} &  
\includegraphics[trim={0 0 350 350},clip,width=.1\textwidth,valign=t]{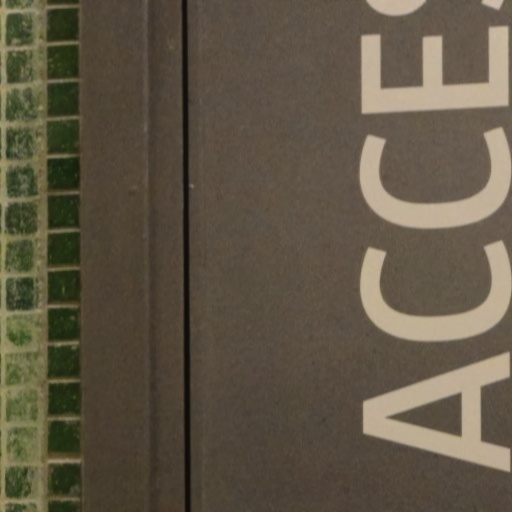} & 
\includegraphics[trim={0 0 350 350},clip,width=.1\textwidth,valign=t]{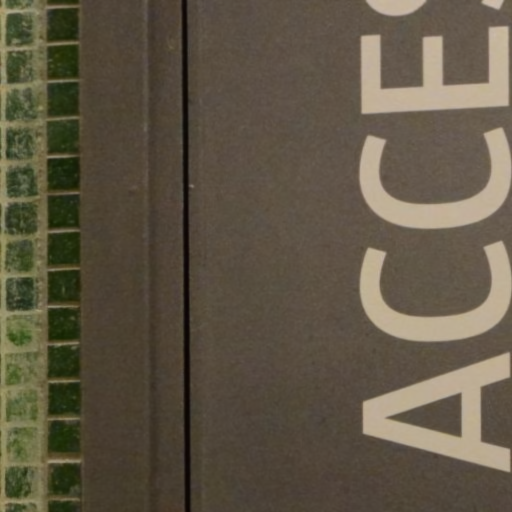} & 
\includegraphics[trim={0 0 350 350},clip,width=.1\textwidth,valign=t]{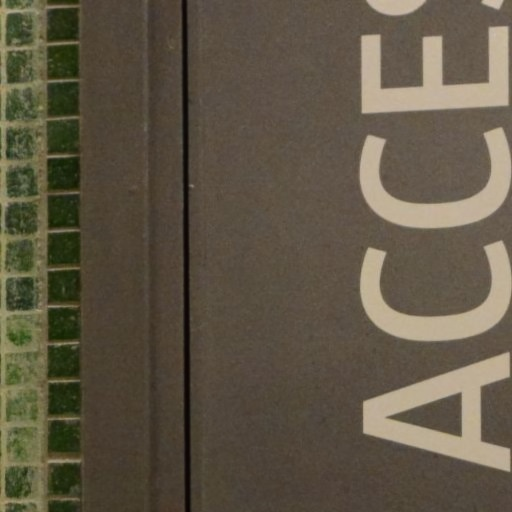} 
\\
& 37.04 / 0.1462
& 36.84 / 0.1666
& 36.99 / 0.1690
& 32.95 / 0.1677
& 36.28 / 0.1599
& 36.93 / 0.1564
& 36.78 / 0.1417

\\
&DANet$_+$%~\cite{ECCV2020_984}
&MIRNet%~\cite{MIR}
&MPRNet%~\cite{zamir2021multi}
&DeamNet%~\cite{ren2021adaptivedeamnet}
&NAFNet%~\cite{chen2022simple} 
&Uformer%~\cite{wang2022uformer}
% &\small{Restormer$_g$~\cite{zamir2022restormer}}
&Restormer
\\

&
%left bottom right top
\includegraphics[trim={0 0 350 350},clip,width=.1\textwidth,valign=t]{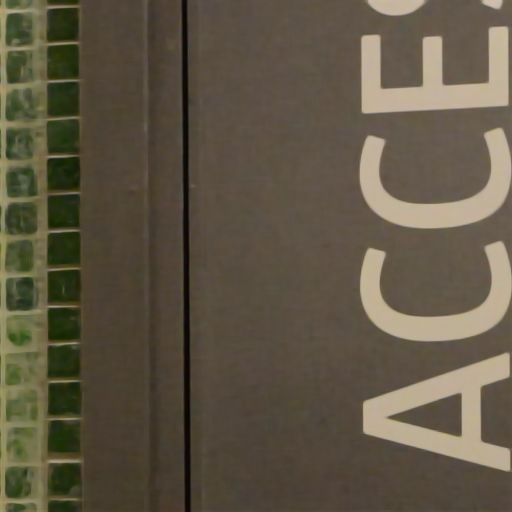} & 
 \includegraphics[trim={0 0 350 350},clip,width=.1\textwidth,valign=t]{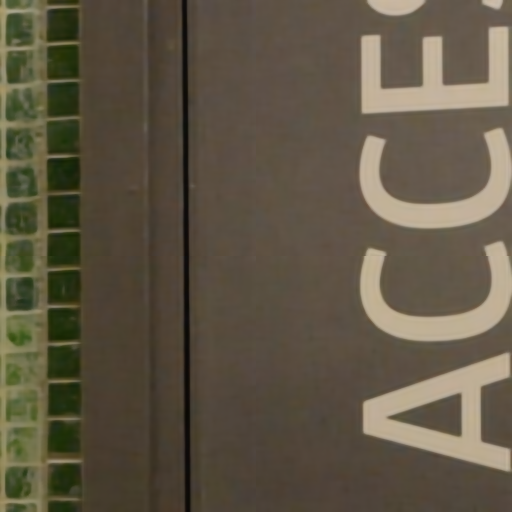} & 
\includegraphics[trim={0 0 350 350},clip,width=.1\textwidth,valign=t]{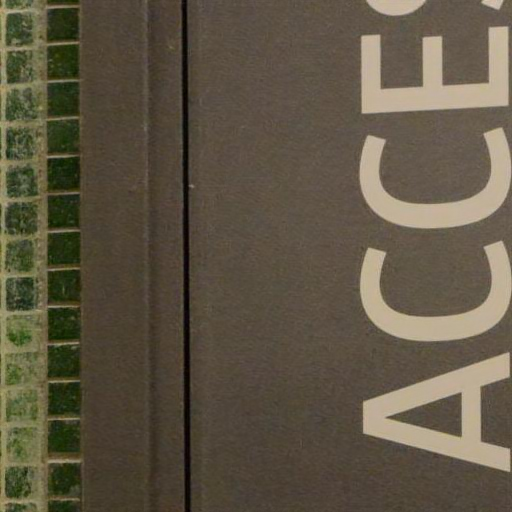} & 
\includegraphics[trim={0 0 350 350},clip,width=.1\textwidth,valign=t]{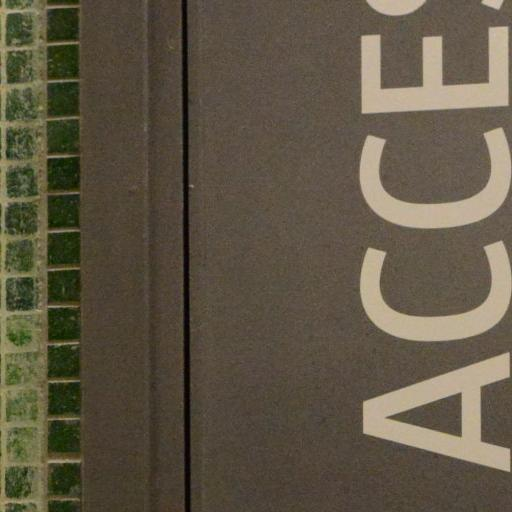} & 
\includegraphics[trim={0 0 350 350},clip,width=.1\textwidth,valign=t]{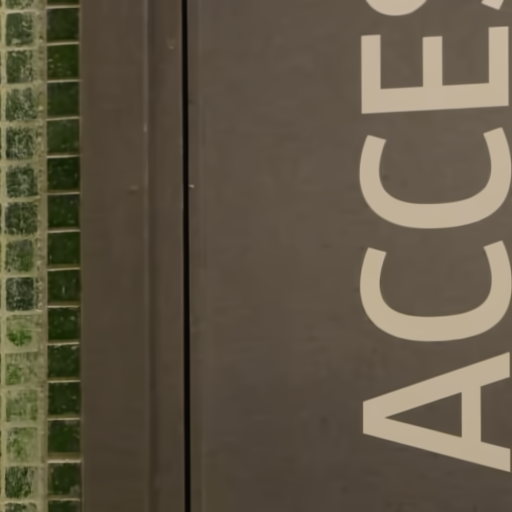} & 
\includegraphics[trim={0 0 350 350},clip,width=.1\textwidth,valign=t]{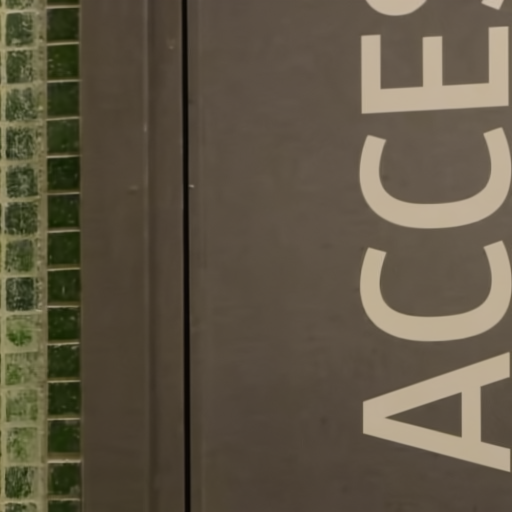} & 
\includegraphics[trim={0 0 350 350},clip,width=.1\textwidth,valign=t]{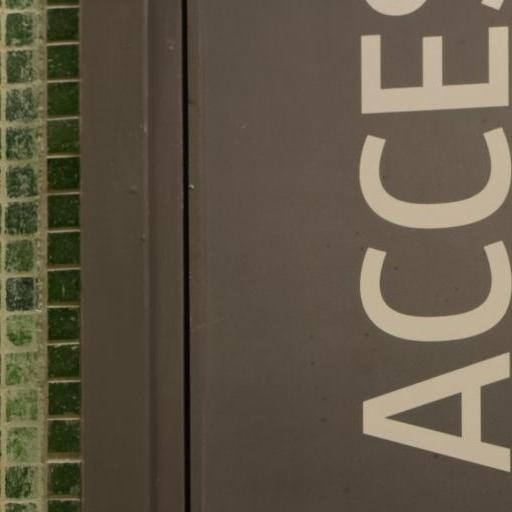} 
\\
35.36 / 0.2316
&36.56 / 0.1074
&36.07 / 0.1212
&35.78 / 0.1824
&35.37 / 0.2315
&{\color{red}39.54 /}{ \color{blue} 0.0658}
&{\color{blue}39.37 /}{ \color{red} 0.0607}
&PSNR / LPIPS

\\
Noisy
&AP-BSN%~\cite{lee2022apbsn}
&LG-BPN%~\cite{wang2023lgbpn}
&R2R%~\cite{Recorrupted2Recorrupted}
&ZS-N2N%~\cite{ZSNoise2Noise} 
& \textbf{DMID-d}
& \textbf{DMID-p} 
&Clean
\\

%left bottom right top
%%%%第4张图片 polyu2

\multirow{4}{*}{\includegraphics[trim={0 0 0 0},clip,width=.245\textwidth,valign=t]{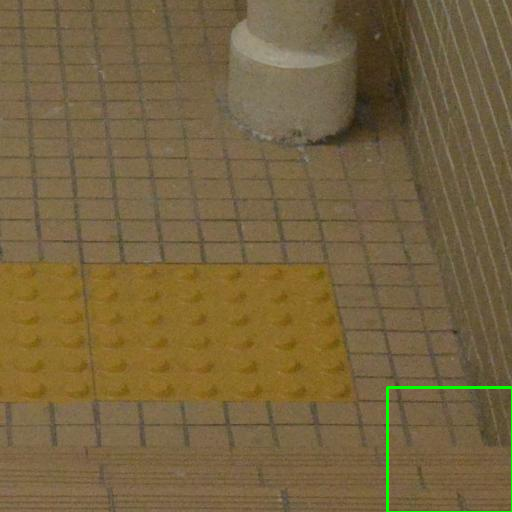}}&
\includegraphics[trim={387 0 0 387},clip,width=.1\textwidth,valign=t]{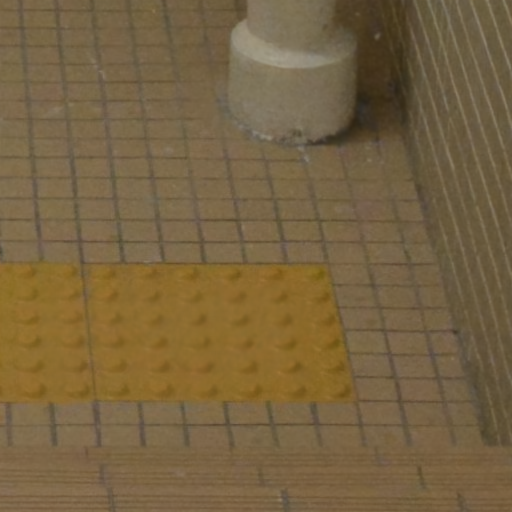} & 
\includegraphics[trim={387 0 0 387},clip,width=.1\textwidth,valign=t]{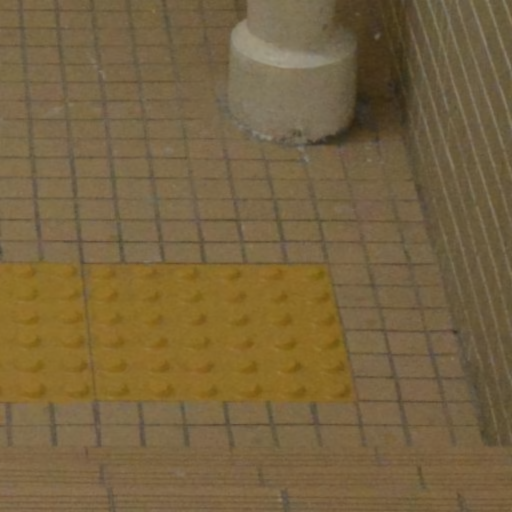} & 
\includegraphics[trim={387 0 0 387},clip,width=.1\textwidth,valign=t]{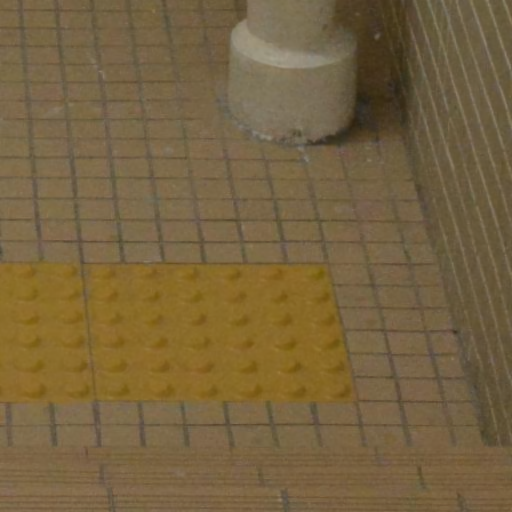} & 
\includegraphics[trim={387 0 0 387},clip,width=.1\textwidth,valign=t]{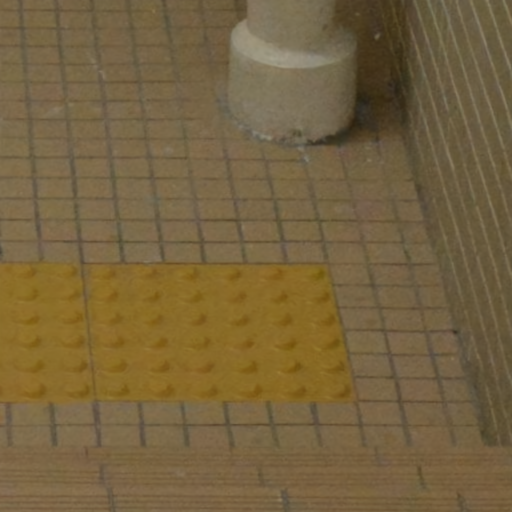} &  
\includegraphics[trim={387 0 0 387},clip,width=.1\textwidth,valign=t]{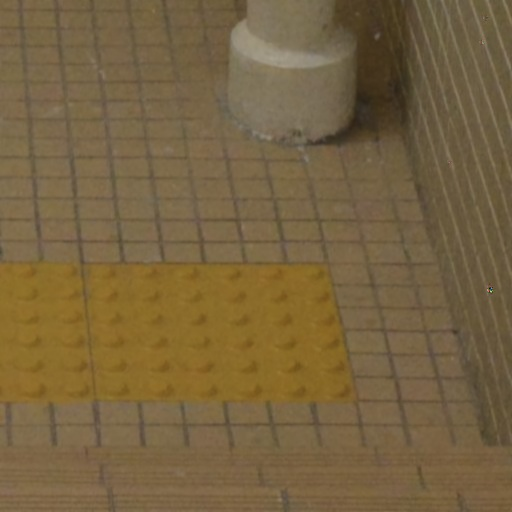} & 
\includegraphics[trim={387 0 0 387},clip,width=.1\textwidth,valign=t]{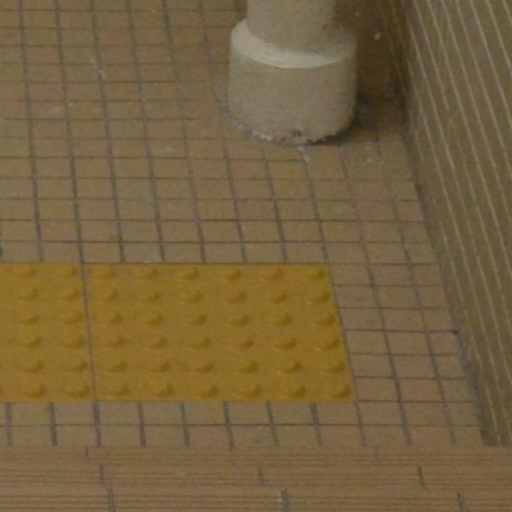} & 
\includegraphics[trim={387 0 0 387},clip,width=.1\textwidth,valign=t]{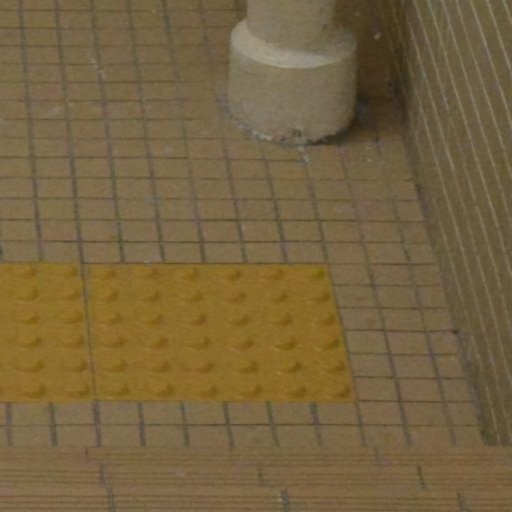} 
\\
& 37.07 / 0.1075
& 37.41 / 0.1155
& 37.03 / 0.1204
& 37.51 / 0.1156
& 37.15 / 0.1337
& 37.72 / 0.1057
& 36.92 / 0.1267

\\
&DANet$_+$%~\cite{ECCV2020_984}
&MIRNet%~\cite{MIR}
&MPRNet%~\cite{zamir2021multi}
&DeamNet%~\cite{ren2021adaptivedeamnet}
&NAFNet%~\cite{chen2022simple} 
&Uformer%~\cite{wang2022uformer}
% &\small{Restormer$_g$~\cite{zamir2022restormer}}
&Restormer
\\

&
%left bottom right top
%\includegraphics[trim={387 0 0 387},clip,width=.1\textwidth,valign=t]{Images/realworld/polyu2/restormer_r.png} & 
\includegraphics[trim={387 0 0 387},clip,width=.1\textwidth,valign=t]{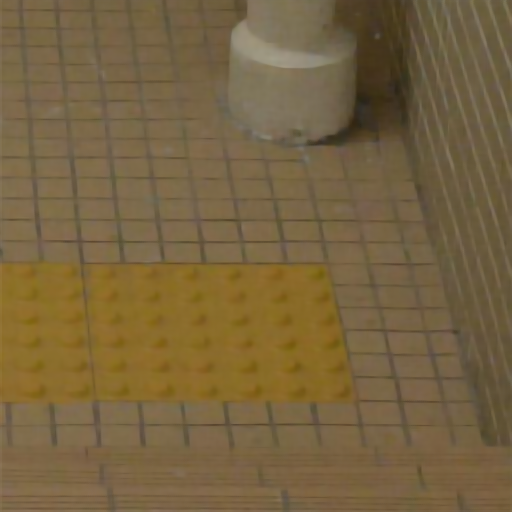} & 
 \includegraphics[trim={387 0 0 387},clip,width=.1\textwidth,valign=t]{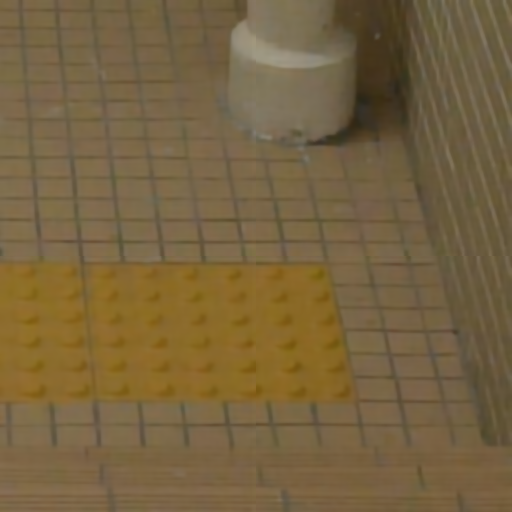} & 
\includegraphics[trim={387 0 0 387},clip,width=.1\textwidth,valign=t]{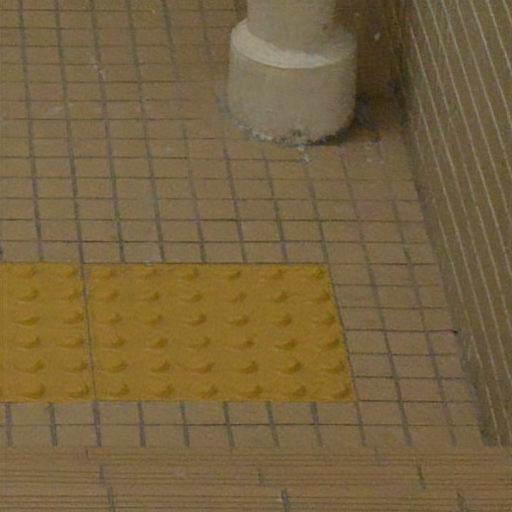} & 
\includegraphics[trim={387 0 0 387},clip,width=.1\textwidth,valign=t]{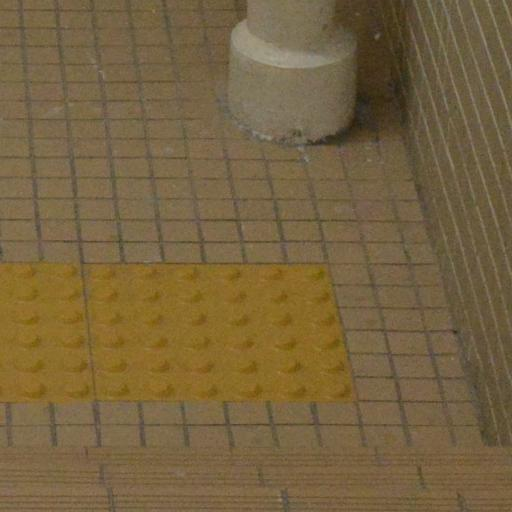} & 
\includegraphics[trim={387 0 0 387},clip,width=.1\textwidth,valign=t]{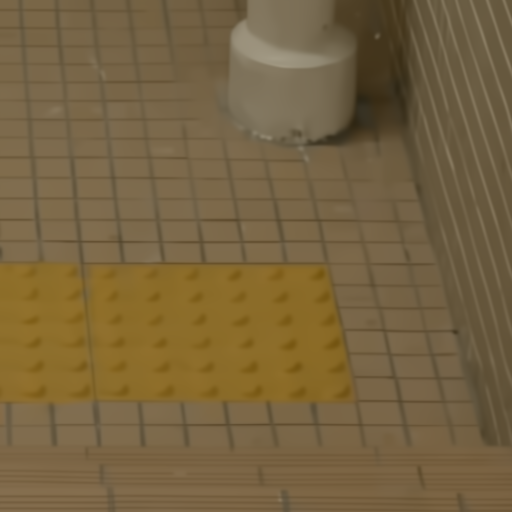} & 
\includegraphics[trim={387 0 0 387},clip,width=.1\textwidth,valign=t]{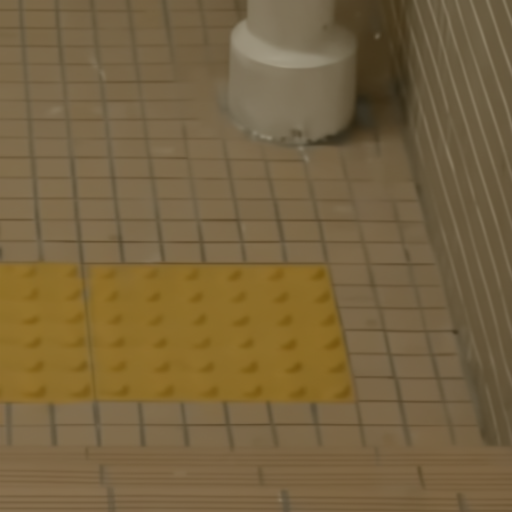} & 
\includegraphics[trim={387 0 0 387},clip,width=.1\textwidth,valign=t]{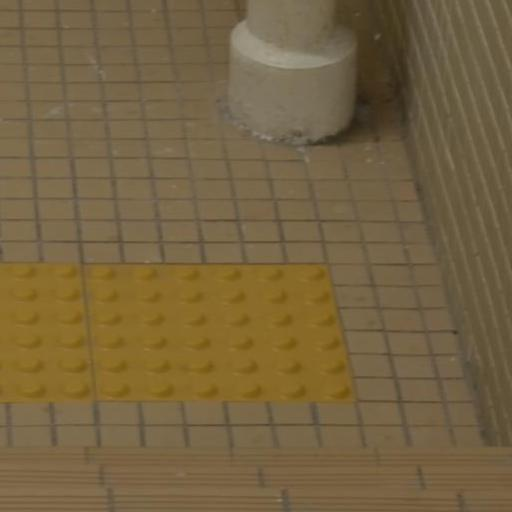} 
\\
34.86 / 0.2063

&37.57 / 0.1113
& 37.90 / 0.1322
&35.27 / 0.1546
&34.86 / 0.2063
% &{\color{red}38.79 /}{ \color{blue} 0.0987}
% &{\color{blue}38.31 /}{ \color{red}0.0896}
&{\color{red}39.33 /}{ \color{blue} 0.0993}
&{\color{blue}39.17 /}{ \color{red}0.0939}
&PSNR / LPIPS

\\
Noisy
%&Restormer$_r$
&AP-BSN%~\cite{lee2022apbsn}
&LG-BPN%~\cite{wang2023lgbpn}
&R2R%~\cite{Recorrupted2Recorrupted}
&ZS-N2N%~\cite{ZSNoise2Noise} 
% & \textbf{DMID-d (Ours)}
% & \textbf{DMID-p (Ours)} 
& \textbf{DMID-d}
& \textbf{DMID-p} 
&Clean
\\

\end{tabular}}
\end{center}
%\vspace*{-2mm}

\caption{Visual results on real-world image denoising. Our method achieves better denoising results and pictures restored by our method are more detailed and realistic. 
}
\label{fig:realworld}
% \vspace{-0.5em}
\end{figure*}
 
	As illustrated in Table \ref{table:2-arbitrary}, our method demonstrates robustness across various noise levels and achieves SOTA performance on all metrics. Under extreme conditions of the ImageNet~\cite{russakovsky2015imagenet} dataset, we outperform Restormer by more than 0.5dB in PSNR. Furthermore, we achieve substantial  improvements in perception-based metrics in all datasets. 
    % Such great improvement brings significant enhancement in perceptual quality, which can be clearly seen in Figure \ref{fig:arbitrary}. 
    This notable enhancement directly translates into improved perceptual quality, as vividly depicted in Figure \ref{fig:arbitrary}. 
    Whether dealing with irregular animal fur, intricate grass patterns, or structured circular designs, our method consistently exhibits superior performance when compared to alternative approaches. This experiment substantiates the robustness of our method across a spectrum of noise levels and evaluation metrics.

	\subsection{Real-world Image Denoising}
	\label{sec:realworld denoising}
	
	In this section, we conduct real-world image denoising experiments on real-world benchmark datasets.

    We compare our method with both supervised and unsupervised methods. As our method is essentially an unsupervised solution for real-world denoising, which does not require training on pairs of noisy-clean real-world images. Secifically, we evaluate different methods on three datasets: CC~\cite{2016A}, PolyU~\cite{xu2018real}, and FMDD~\cite{zhang2018poisson}, following~\cite{zheng2021unsupervised,SocreDVI,Recorrupted2Recorrupted,ZSNoise2Noise}. 
    % These datasets contain 16 paired sRGB images, 100 paired sRGB images, and 48 paired raw images, respectively.
    
    Subsequently, we will provide a detailed explanation of the comparative methods. 
    For supervised methods, 
    DANet$_+$~\cite{ECCV2020_984} is the only and the latest generative method that can be directly employed for denoising. In addition, 
    % Restormer$_g$ is pre-trained on Gaussian noise, while others are pre-trained on real-world dataset SIDD~\cite{SIDD}, and all of them are borrowed from their officially released versions. Restormer$_g$ first employs NN~\cite{zheng2021unsupervised} to convert real noisy images into a latent space where the Gaussian noise assumption remains valid, followed by denoising in that latent space. 
    all the supervised methods are pre-trained on real-world dataset SIDD~\cite{SIDD} and borrowed from their officially released versions.
    For unsupervised methods, the results of AP-BSN~\cite{lee2022apbsn}, LG-BPN~\cite{wang2023lgbpn}, ZS-N2N~\cite{ZSNoise2Noise}, R2R~\cite{Recorrupted2Recorrupted} are reproduced and evaluated by ourselves, since these methods are not evaluated on all the datasets and metrics we use in their original paper. The results of N2V~\cite{noise2void}, N2S~\cite{noise2self} and S2S~\cite{self2self} are reported from R2R~\cite{Recorrupted2Recorrupted} and ScoreDVI~\cite{SocreDVI}. 
    % Since we reproduce more advanced methods AP-BSN~\cite{lee2022apbsn} and LG-BPN~\cite{wang2023lgbpn} for N2V~\cite{noise2void} and N2S~\cite{noise2self}. In addition, S2S~\cite{self2self} requires too many iterations ($4.5\times10^{5}$) to produce a single image which takes around $4\sim5$ hours to process an image of size $512\times512$. %Therefore, we report the results of S2S 
    % For our method, the noise level of the latent noisy image after noise transformation is pre-defined as a constant for  CC~\cite{2016A}, PolyU~\cite{xu2018real}, and FMDD~\cite{zhang2018poisson}, following NN method~\cite{zheng2021unsupervised}. In practice, fine-tuning the noise level parameter could potentially lead to even better performance for our method.

	Table \ref{table:1-real} and Figure \ref{fig:realworld} demonstrate our significant advantages. 
    % The top row contains supervised methods while unsupervised methods lie in the bottom row in Table \ref{table:1-real}.
    The Table \ref{table:1-real} segregates supervised methods in the top row and unsupervised methods in the bottom row.
    %Supervised methods generally beat unsupervised methods except for our method. However, all these supervised methods are trained on the SIDD dataset. The supervised training leads to poor generalization and makes them fragile to out-of-distribution data. This is why the images restored by other methods either fail to denoise properly as seen in Figure \ref{fig:realworld}.
	In contrast to supervised methods, our unsupervised approach yields significantly higher PSNR and SSIM scores. Supervised methods, trained on the SIDD dataset~\cite{SIDD}, suffer from poor generalization, rendering them highly vulnerable to out-of-distribution data. As a result, their outcomes consistently exhibit significant residual noise.
	In contrast to unsupervised methods, our results demonstrate a substantial improvement. AP-BSN~\cite{lee2022apbsn} and LG-BPN~\cite{wang2023lgbpn} often yield overly blurred outputs, accompanied by texture distortion, whereas our results maintain remarkably clear textures. R2R~\cite{Recorrupted2Recorrupted} and ZS-N2N~\cite{ZSNoise2Noise} appear to grapple with complete noise removal, leaving residual noise in the final results.
	% In contrast to the generative method, our results surpass DaNet$_+$~\cite{ECCV2020_984} comprehensively across both metrics and visual performance. 
 	These experiments unequivocally underscore the adeptness of our method in handling denoising tasks, achieving high perceptual quality while introducing minimal distortion.

	The abnormal situation observed in the FMDD~\cite{zhang2018poisson} dataset concerning the LPIPS metric is worth discussing. It is notable that our method and DANet$_+$~\cite{ECCV2020_984}, both generative methods, perform well on LPIPS for CC~\cite{2016A} and PolyU~\cite{xu2018real} but fail for FMDD~\cite{zhang2018poisson}. Additionally, our method and LG-BPN~\cite{wang2023lgbpn} perform well on PSNR and SSIM for FMDD~\cite{zhang2018poisson} but not on LPIPS. This situation appears abnormal. 
	We hypothesize that it may be due to the perception-distortion curve phenomenon~\cite{8578750}. The perception-distortion curve implies a contradiction between distortion-based metrics and perception-based metrics. Therefore, outstanding performance on distortion-based metrics can not align with outstanding perception-based metrics. 
	This explanation is consistent with our results on CC~\cite{2016A}, where our DMID-d does not achieve the second-best results on LPIPS.
	Despite the presence of such an abnormal situation, our method still stands out among unsupervised methods. 
    Disregarding this abnormal situation, our method shines brightly even when compared to various supervised methods.

\begin{table}[t!]
	%\small
 \centering
	\caption{Results of the ablation study on the Embedding method with whether perform noise transformation.}
	\label{table:7-embedding}
	\setlength{\tabcolsep}{1.5pt}
	\renewcommand{\arraystretch}{1.2}
	\scalebox{0.99}{
		\begin{tabular}{c| c| c}
        \toprule
       {} &{CC} & {FMDD} 
        
        \\ 
        \cline{2-3} 
        \multirow{-2}{*}{Noise Transformation} 
        & PSNR / SSIM / LPIPS
        & PSNR / SSIM / LPIPS
    \\ 
    \midrule
       \XSolidBrush
       &\color{black}{35.97 / 0.9769 / 0.095}
       &\color{black}{29.47 / 0.6806 / 0.334}
      
       \\
       \Checkmark
       &\color{black}{37.09 / 0.9854  / 0.072}
       &\color{black}{31.97 / 0.7830 / 0.314}
       \\
       \bottomrule%[0.1em]

		\end{tabular}%
	}
\vspace*{-1em}
\end{table}
	\subsection{Ablation Study}
	\label{sec:ablation_study}
	In this section, we conduct ablation studies to evaluate our embedding method and ensembling method.

    \subsubsection{For embedding method}
	 
	Our embedding method first performs noise transformation and then converts the image to the intermediate state $x_N$.
	Therefore, for the embedding method, we first assess the impact of noise transformation. Then, we evaluate the impact of the choice of $N$, which is crucial for our method, as denoising cannot be performed without it for our DMID method.

    \begin{figure*}[t]
\vspace{2em}
 \includegraphics[width=0.99\textwidth]{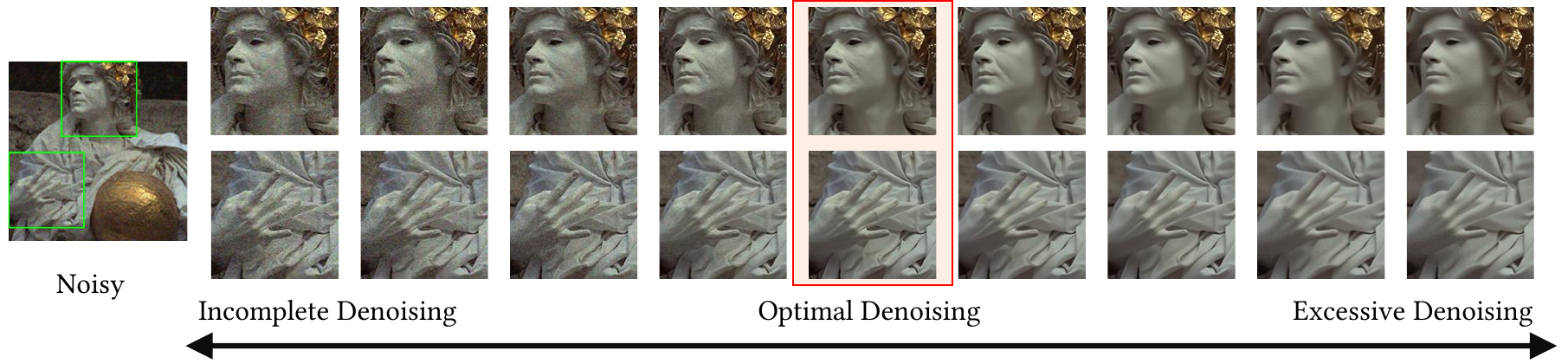}
 \caption{Visual results of ablation study on the embedding method.  The most optimal denoising outcomes are realized when the denoising ability of the diffusion model at timestep $N$ matches the noise level of the noisy image $x_N$. }

 \label{fig:ablation_N_2}

\end{figure*}
    % \begin{figure*}[t]
% \vspace{1em}
%  \includegraphics[width=0.99\textwidth]{Images/ablation/ablation_N2.pdf}
%  \caption{Results of Ablation Study of Embedding Method. The horizontal axis is $\frac{Denoising Ability}{Noise Level}$. The most optimal denoising outcomes are realized when the denoising ability of the diffusion model at timestep $N$ matches the noise level of the noisy image $x_N$. }
%  \Description{compare. }
%  \label{fig:ablation_N}

% \end{figure*}

\begin{figure}[t]
% \vspace{1em}
 \includegraphics[width=0.49\textwidth]{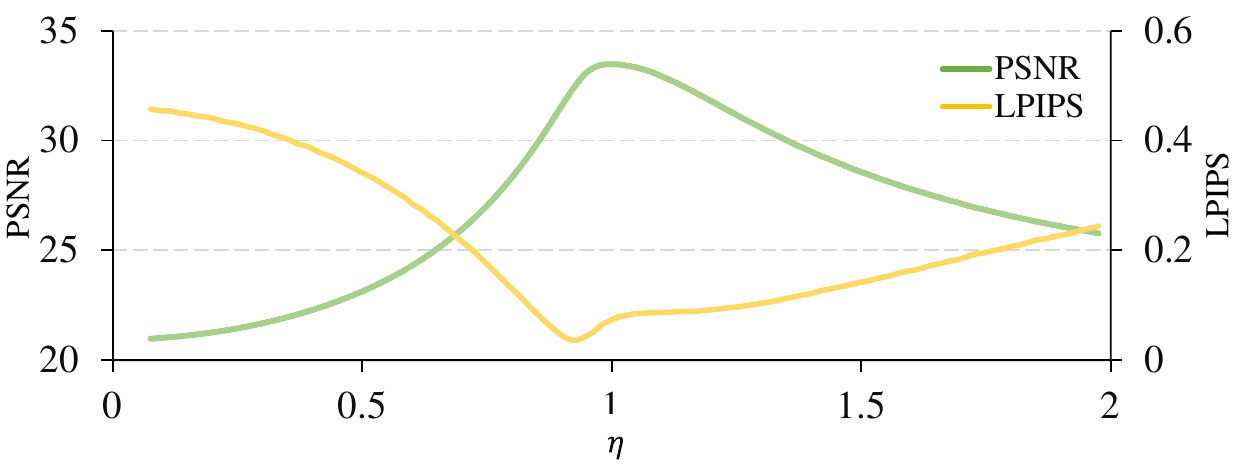}
% \vspace{-2em} 
 \caption{Results of the ablation study on the embedding method with varying timestep $N$. The most optimal denoising outcomes are realized when $\eta \approx 1$.
 %The horizontal axis is $\frac{Denoising \ Ability}{Noise \ Level}$.
 }

 \label{fig:ablation_N_1}

\end{figure}

    % \begin{figure*}[ht]

% \begin{tabular}[b]{c c}
%  \hspace{-4mm}

%  \subfloat[$S_t$]{\includegraphics[width=0.5\linewidth]{Images/ablation/S_t_2.pdf}
%  }
% \label{abliton_sub1}

% \subfloat[$K$]{\includegraphics[width=0.5\linewidth]{Images/ablation/K.pdf}
% \label{abliton_sub2}}
% \\
% %\small~$time_a$ & \small~$k$
% \end{tabular}

% \vspace*{-2mm}
% \caption{Results on \underline{Ablation Study}. Different parameters produce different effects, and the actual effect is very consistent with our theory.
% }
% \label{fig:ablation}
% \vspace{-0.5em}
% \end{figure*}

%\begin{figure*}[ht]
%%\vspace{-1em}
%
%\begin{tabular}[b]{c c}
% \hspace{-4mm}
%
%\subfloat[$S_t$]{\includegraphics[width=0.5\linewidth]{Images/ablation/S_t_3.pdf}
%\label{abliton_sub1}
% }
%
%
%\subfloat[$R_t$]{\includegraphics[width=0.5\linewidth]{Images/ablation/K_2.pdf}
%\label{abliton_sub2}
%}
%\\
%%\small~$time_a$ & \small~$k$
%\end{tabular}
%
%% \vspace*{-2mm}
%%\vspace{-1em}
%% \caption{Results on \underline{Ablation Study}. Different parameters produce different effects, and the actual effect is consistent with our theory.
%% }
%% \caption{Results on Ablation Study. Different parameters produce different effects, and the actual effect is consistent with our theory.
%\caption{Results on ablation study of ensembling method. Different parameters produce different effects which are consistent with our theory.
%}
%\label{fig:ablation}
%% \vspace{-0.5em}
%%\vspace{-1em}
%\end{figure*}

\begin{figure}[t]
    \includegraphics[width=.49\textwidth]{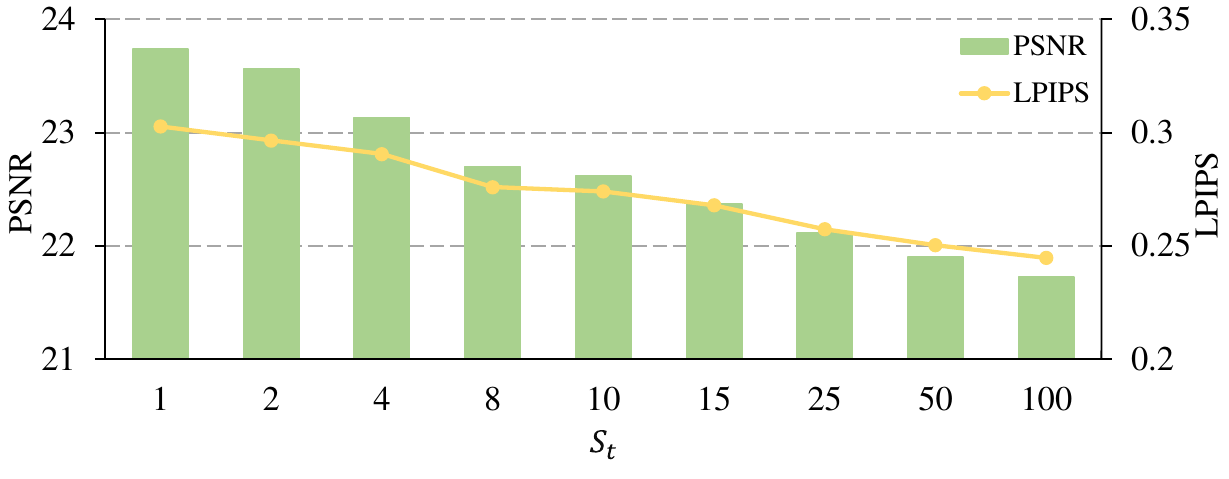}
    % \caption{Results on ablation study of ensembling method on sampling timesteps $S_t$. More sampling timesteps $S_t$ can result in more distortion, when $S_t>1$ for Gaussian denoising. 
    % }
\caption{
Results of the ablation study on the ensembling method with varying sampling times $S_t$.
Increased sampling times $S_t$ result in higher distortion, particularly when $S_t>1$ for Gaussian denoising.
}    
    \label{fig:ablation_s_t}
\end{figure}

% \begin{figure}[!t]
% 	%\vspace{-1em}
% 	\includegraphics[width=.49\textwidth]{Images/ablation/R_t.pdf}
% 	% \caption{Results on ablation study of ensembling method on repetition times $R_t$. More repetition times $R_t$ can result in less distortion.
% 	% }
% \caption{
% Results of the ablation study on the ensembling method with varying repetition times $R_t$.
% Increased repetition times $R_t$ result in less distortion, when $R_t>1$ for both Gaussian denoising and real-world denoising.
% }     
% 	\label{fig:ablation_r_t}
% \end{figure}

	% \paragraph{\textbf{Influence of noise transformation}}
    \textbf{(a). Influence of noise transformation.}
	We conduct experiments on CC~\cite{2016A} and FMDD~\cite{zhang2018poisson} with a fixed number of sampling times $S_t=1$ and repetition times $R_t=1$. 
 
	Without converting the image into the intermediate state $x_N$ of the diffusion model, our method cannot function. Therefore, we consistently convert the image into the intermediate state $x_N$, whether performing noise transformation or not. The results are presented in Table \ref{table:7-embedding}, and it is evident that noise transformation brings about significant improvement. This experiment demonstrates the importance of noise transformation for our method. 
    % With noise transformation, our approach becomes versatile in addressing various noise types.

	% \paragraph{\textbf{Influence of $N$}}
    \textbf{(b). Influence of $N$.}
	We conduct experiments on McMaster with noise level $\sigma=25$ using different choices of $N$ with a fixed number of sampling times $S_t=1$ and repetition times $R_t=1$. 
 
	As explained in Section \ref{sec:embedding}, the most optimal denoising outcomes are realized when the denoising ability of the diffusion model at timestep $N$ matches the noise level of the noisy image $x_N$. 
	In simpler terms, we define the ratio of the denoising ability and the noise level as $\eta$, and when $\eta \approx 1$ at timestep $N$, here we can get optimal denoising outcomes and the timestep $N$ corresponds to such a noise level.
	This concept is intuitively straightforward, and the denoising outcomes fluctuate based on alterations in the correlation with $\eta$ as illustrated in Figure \ref{fig:ablation_N_2}. The distinction between appropriate and inappropriate values of timestep $N$ can result in a significant difference in PSNR and LPIPS.

	Since LPIPS is more tolerant to noise than to blurring \cite{zhang2018perceptual}, the highest LPIPS score is frequently achieved when denoising is slightly incomplete. When $\eta \approx 1$ at timesetp $N$, this optimal timesetp $N$ is denoted as $N_o$.  As for our method, the optimal LPIPS is generally obtained within the range of $N_o\!\!-\!1$ to $N_o\!\!-\!5$ as shown in Figure \ref{fig:ablation_N_1}. In our experiments, we just set $N$ to be $N_o$, and further tuning could potentially yield even better LPIPS results. 
    %This phenomenon, where the optimal LPIPS is not attained at $N_{\text{optimal}}$, further substantiates the soundness of our approach. It underscores the significance of maintaining the appropriate noise level to achieve optimal denoising outcomes.
    This phenomenon, in which the optimal LPIPS score is not attained at $N_o$ while PSNR is achieved at $N_o$, further validates the credibility of our approach. It underscores the accuracy of our $N$ calculation.
	Furthermore, our method demonstrates robustness, showing favorable denoising outcomes when $\eta$ varies within the range of $1 \pm 10\%$ as shown in Figure~\ref{fig:ablation_N_1}.
	
    This experiment demonstrates the necessity of embedding into intermediate states and the appropriate choice of $N$ for our method. Without embedding into intermediate states, our method cannot perform denoising. With the appropriate choice of $N$, our denoising performance reaches its maximum potential.

    \subsubsection{For ensembling method}

    % \textbf{(a).For ensembling method.}
	Our ensembling method adjusts distortion and perception based on requirements by adjusting the sampling times and repeating the inference process.
	Therefore, for the ensembling method, we separately evaluate the impact of different values for sampling times $S_t$ and repetition times $R_t$.

	% \paragraph{\textbf{Influence of $S_t$}}
     \textbf{(a). Influence of $S_t$.}
	We conduct experiments on McMaster with noise level $\sigma=250$ using different sampling times $S_t$ with a fixed number of repetition times $R_t=1$ for Gaussian denoising. In addition, we conduct experiments on CC~\cite{2016A}, PolyU~\cite{xu2018real}, and FMDD~\cite{zhang2018poisson} for real-world denoising.

	Increasing sample timesteps $S_t$ can lead to more corruption and reconstruction times, and weaken the controllability of the noisy image $y$ introduced in each iteration. While this can improve perceptual quality, it may also increase the uncontrollability of picture details resulting in less similarity in PSNR. The experimental results are as shown in Figure \ref{fig:ablation_s_t} for Gaussian denoising and Table \ref{table:5-ensemble} for real-world denoising.

    The overall impact trend of sample timesteps $S_t$ is similar for both Gaussian denoising and real-world denoising. However, there are some noteworthy differences to consider. For Gaussian denoising, when the repetition times are set to $R_t=1$, the optimal distortion result (the highest PSNR and SSIM) is attained with a sampling time of $S_t=1$. However, with an increase in repetition times, results obtained within the range of $1<S_t<10$ can surpass the outcome obtained with a sampling time of $S_t=1$. For real-world, denoising, the optimal distortion result is always attained with sampling times of $S_t=2$ or $S_t=3$. The results are shown in Table \ref{table:5-ensemble}. This is because real-world denoising is more challenging, and 2-3 iterations tend to yield better results with relatively less stochasticity. Increasing the number of iterations introduces excessive stochasticity, and the performance improvement is not sufficient to compensate for the loss caused by increased stochasticity. 
    %This phenomenon highlights the effectiveness of our ensembling method and the validity of our explanations.

    This experiment illustrates the capacity of our method to adjust for distortion and perception.

\begin{table}[t!]
\centering
\caption{
Results of the ablation study on the ensembling method with varying sampling timesteps $S_t$.
Increased sampling timesteps $S_t$ result in higher distortion, particularly when $S_t>3$ for real-world denoising.
}
	\label{table:5-ensemble}
	\setlength{\tabcolsep}{1.5pt}
	\renewcommand{\arraystretch}{1.2}
	\scalebox{1}{
		\begin{tabular}{c |c |c |c|c }
			\toprule
			{\makebox[0.055\textwidth][c]{}} &
			{\makebox[0.055\textwidth][c]{} }&
			{\makebox[0.12\textwidth][c]{CC~\cite{2016A}}}  
			&{\makebox[0.12\textwidth][c]{PolyU~\cite{xu2018real}}} 
            &{\makebox[0.12\textwidth][c]{FMDD~\cite{zhang2018poisson}}} 
			
			\\ 
			\cline{3-5} 
			\multirow{-2}{*}{$S_t$}
            &\multirow{-2}{*}{$R_t$}
			& PSNR / SSIM
                & PSNR / SSIM
			& PSNR / SSIM

			\\ 
			\midrule
			
			$S_t=1$
                &$R_t=1$
			
			&37.09 / 0.9854
			&38.46 / 0.9843
			&31.97 / 0.7830 
			\\
                $S_t=2$
                &$R_t=1$
			
			&\color{red}{37.84 / 0.9877}
                &\color{red}{38.56 / 0.9849}  
                &33.09 / 0.8616 
			
			\\
                $S_t=3$
                &$R_t=1$
			
			&\color{blue}{37.82 / 0.9876}
                &\color{red}{38.56 / 0.9849} 
                &\color{red}{33.22 / 0.8693}
                
			\\
                $S_t=4$
                &$R_t=1$
			&37.73 / 0.9873  
                &\color{blue}{38.52 / 0.9847} 
                &\color{blue}{33.15 / 0.8671}
                
			\\
                $S_t=5$
                &$R_t=1$
			
			&37.67 / 0.9871  
                &38.51 / 0.9847 
                &33.11 / 0.8668 
                			
			\\
                $S_t=N$
                &$R_t=1$
                &37.49 / 0.9866 
                &38.50 / 0.9847 
                &32.85 / 0.8578 
                % \\

			\\
			\bottomrule%[0.1em]
		\end{tabular}%
	}
\end{table}

    \begin{figure}[!t]
	%\vspace{-1em}
	\includegraphics[width=.49\textwidth]{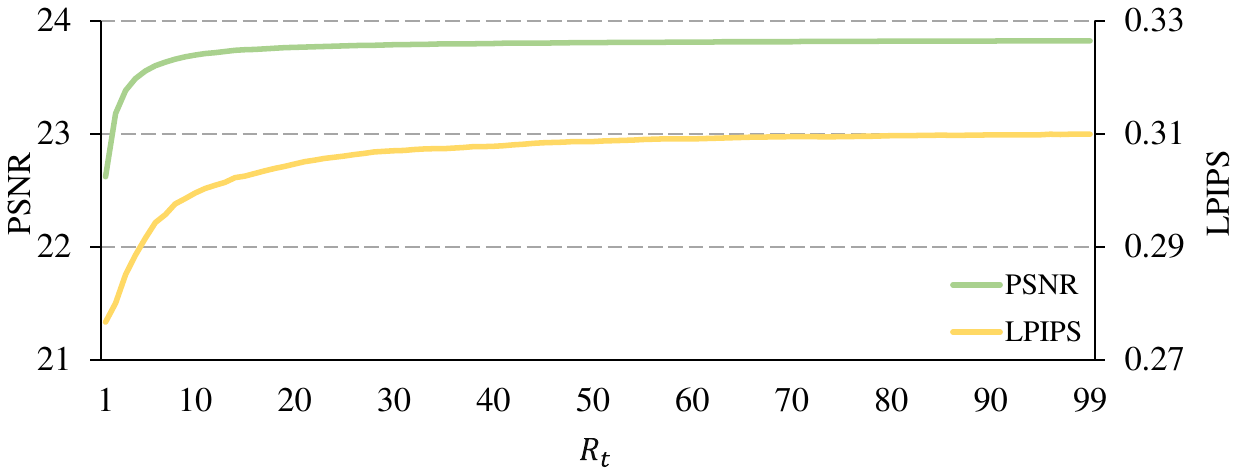}
	% \caption{Results on ablation study of ensembling method on repetition times $R_t$. More repetition times $R_t$ can result in less distortion.
	% }
\caption{
Results of the ablation study on the ensembling method with varying repetition times $R_t$.
Increased repetition times $R_t$ result in less distortion, when $R_t>1$ for both Gaussian denoising and real-world denoising.
}     
	\label{fig:ablation_r_t}
\end{figure}

\begin{table*}[t!]
\vspace{-1em}
% \vspace*{-5mm}
	%\small
 \centering
	\caption{Compared with other diffusion-based methods on real-world denoising.}
 % \vspace*{-3mm}
	\label{table:9-compare and extend}
	\renewcommand{\arraystretch}{1.2}
	\setlength{\tabcolsep}{1pt}
	\scalebox{1}{
		\begin{tabular}{c| c| c| c c c| c c c |c c c }
			\toprule
			{\makebox[0.15\textwidth][c]{}}
			%& \multicolumn{2}{c|}{Auxiliary tools}
                & \multicolumn{2}{c|}{Extension with our method}
			& \multicolumn{3}{c|}{CC~\cite{2016A}}
			& \multicolumn{3}{c|}{PolyU~\cite{xu2018real}} 
			& \multicolumn{3}{c}{FMDD~\cite{zhang2018poisson}}
			
			\\ 
			\cline{2-12} 
	    \multirow{-2}{*}{Method}  		
            % &\multirow{-2}{*}{Auxiliary tools}
            &{\makebox[0.12\textwidth][c]{Embedding}}
            &{\makebox[0.12\textwidth][c]{Ensembling}}

			& {\makebox[0.06\textwidth][c]{PSNR}} 
			& {\makebox[0.06\textwidth][c]{SSIM}}
			& {\makebox[0.06\textwidth][c]{LPIPS}}
   
			& {\makebox[0.06\textwidth][c]{PSNR}} 
			& {\makebox[0.06\textwidth][c]{SSIM}} 
			& {\makebox[0.06\textwidth][c]{LPIPS}}
   
			& {\makebox[0.06\textwidth][c]{PSNR}} 
			& {\makebox[0.06\textwidth][c]{SSIM}} 
			& {\makebox[0.06\textwidth][c]{LPIPS}}
			
			\\ 
			\midrule

		{DDRM}
            &\XSolidBrush
            &\XSolidBrush
            &33.52 & 0.9575 & 0.221
		&36.00 & 0.9590 & 0.196
		&27.23 & 0.5451 & 0.450
           \\
            DDRM+Clean
            &\XSolidBrush
            &\XSolidBrush
            &36.03 & 0.9788 & 0.081
            &38.07 & 0.9828 & {\color{red}0.057}
            &29.61 & 0.7156 & 0.291
		\\
           {DDNM}
            & \XSolidBrush
            & \XSolidBrush
            & 33.52 & 0.9575 & 0.222
            & 36.01 & 0.9590 & 0.196
            & 27.22 & 0.5442 & 0.450
            \\
            DDNM+Clean
            & \XSolidBrush
            & \XSolidBrush
            & 35.27 & 0.9723 & 0.126
            & 36.94 & 0.9700 & 0.133
            & 29.18 & 0.6790 & 0.328
            \\
            \midrule
            DM+Ours
            % DM+Embedding
            &\Checkmark
            &\XSolidBrush

            %clean:
		% &37.52 & 0.9859  & 0.085
            %&38.76 & 0.9854  & 0.056
            %&33.65 & 0.8936  & 0.236

            %sure
            
            % &36.86 &0.9837 &0.090
		&37.20 &0.9855 &\color{red}{0.072}
            &38.45 &0.9844 &\color{blue}{0.067}
            &30.99 &0.7721 &0.308
		\\
		DM+Ours
            % DM+Embedding+Ensembling
            %\multirow{-2}{*}{DM+Ours}
            &\Checkmark
		&\Checkmark
            %clean:
    	%&38.36 & 0.9892  & 0.070
              %&{\color{blue}38.34} & {\color{red}0.9892}  & {\color{red}0.070}
              %&{\color{blue}38.86} & {\color{blue}0.9858}  & 0.062
		% &{\color{blue}33.75} & {\color{blue}0.8937}  &        {\color{blue}0.235}	

            %sure
            &\color{blue}{37.97} &\color{blue}{0.9878} &0.079
            &\color{blue}{38.57} &\color{blue}{0.9851} &0.070
		&32.64 &0.8788 &\color{blue}{0.254}
		\\ 
           %\midrule
           {\textbf{DMID-d (Ours)}} 
           &\Checkmark
           &\Checkmark

           %利用clean的结果：
           % &{\color{red}38.36} & {\color{red}0.9892} & 0.075
           % &{\color{red}38.93} & {\color{red}0.9861} & 0.063
           % &{\color{red}33.92} & {\color{red}0.8982} & 0.237
            %SURE的结果：
            & \color{red}{37.99} & \color{red}{0.9880}  & \color{blue}{0.078}
            &  \color{red}{38.62} & \color{red}{0.9853} & \color{black}{0.069}
            &  \color{red}{33.40} & \color{red}{0.8747} & \color{black}{0.266}
           \\
           \textbf{DMID-p (Ours)}
           &\Checkmark
           &\Checkmark
           %利用clean的结果：
            % &38.12 & 0.9887 & {\color{blue}0.072}
            % &38.79 & 0.9853 & {\color{blue}0.059}
            % &33.65 & 0.8774 & {\color{red}0.224}
            
            %SURE的结果：
            &\color{black}{37.09} & \color{black}{0.9854}  & \color{red}{0.072}
            &\color{black}{38.46} & \color{black}{0.9843}  & \color{blue}{0.067}
            &\color{blue}{33.09} & \color{blue}{0.8616}  & \color{red}{0.232}
            \\
			\bottomrule %[0.1em]
			
		\end{tabular}%
		
	}
% \vspace*{-5mm}
\end{table*}

	% \paragraph{\textbf{Influence of $R_t$}}
	\textbf{(b). Influence of $R_t$}.
	We perform experiments on McMaster with noise level $\sigma=250$ using different repetition times $R_t$ and the same sampling times $S_t=10$. 

    \begin{figure}[t]
\centering
 \includegraphics[width=0.48\textwidth]{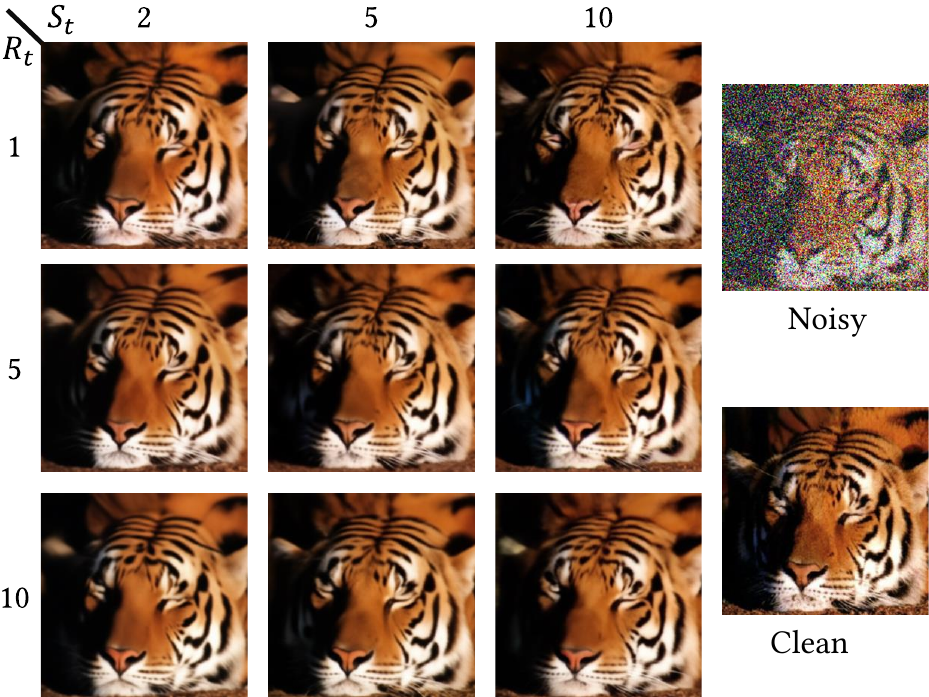}
 \caption{Visual results of ablation study on the ensembling method.
 (Best viewed with zoom-in)
 }
 \label{fig:visual_S_R}
\end{figure}

	More repetition times $R_t$ can result in denoised images that are more likely to match probability, resulting in less distortion. 
    % However, each candidate image obtained is slightly different, and taking the average may lose some details. 
    The increase brought about by the ensembling strategy has a severe marginal utility for distortion. The increase after 10 times is small, but the impact on perception lasts longer as shown in Figure \ref{fig:ablation_r_t}.
    While each candidate image obtained exhibits slight variations, averaging them could potentially lead to the loss of minor details. It is important to emphasize that critical details are, nonetheless, preserved. For instance, in Figure \ref{fig:visual_S_R}, we can observe a gradual enrichment of details as the number of sampling times $S_t$ increases. Furthermore, with an increase in repetition times $R_t$, we notice that most details tend to converge closer to the clean image, including fine features such as whiskers, which remain faithfully intact.

    This experiment validates the robustness of our analysis and further illustrates the capacity of our method to reduce distortion.

    \subsection{Comparative Analysis and Extension to Diffusion-Based Methods}
    \label{sec:compare and extend}

    In this section, we embark on a comprehensive analysis that involves comparing our proposed method with other diffusion-based image restoration methods. Additionally, we explore the adaptability of our method to address specific challenges and limitations encountered by existing diffusion-based methods.

    \begin{table}[t]
%\small
\caption{Compared with other diffusion-based methods on Gaussian denoising.
}
\label{table:4-diffusion}
\renewcommand{\arraystretch}{1.2}
\setlength{\tabcolsep}{1.5pt}
\scalebox{0.92}{
    \begin{tabular}{c| c |c| c }
    \toprule
    {\makebox[0.05\textwidth][c]{Noise}} &
    {\makebox[0.13\textwidth][c]{DDRM~\cite{kawar2022denoising}}} & 
    {\makebox[0.13\textwidth][c]{DDNM~\cite{wang2022zero}}} & 
    {\makebox[0.13\textwidth][c]{\textbf{DMID-d (Ours)}}} 

    \\ 
    \cline{2-4} 
    Level
    & PSNR / SSIM / LPIPS
    & PSNR / SSIM / LPIPS
    & PSNR / SSIM / LPIPS
    \\ 
    \midrule

%     $\sigma=50$

% & 28.76 / 0.8950 / 0.163
%     & {\color{blue}29.20 / 0.9030 / 0.158}
%     & \color{red}{29.92} / \color{red}{0.9155} / \color{red}{0.121} 
%     \\
%     $\sigma=100$

%     & 26.12 / 0.8395 / 0.261
%     & {\color{blue}26.40 / 0.8484 / 0.249}
%     & \color{red}{27.01} / \color{red}{0.8621} / \color{red}{0.211}
    
%     \\
    
%     $\sigma=150$

% 	& 24.70 / 0.8036 / 0.316
%     & {\color{blue}24.77 / 0.8085 / 0.308}
%     & \color{red}{25.39} / \color{red}{0.8233} / \color{red}{0.274}  
%     \\
    
%     $\sigma=200$

%     & {\color{blue}23.66} / 0.7737 / 0.362
%     &23.61 / {\color{blue}0.7763 / 0.352}
%     & \color{red}{24.28} / \color{red}{0.7931} / \color{red}{0.305} 

%     \\
%     $\sigma=250$ 

%     & {\color{blue}22.81} / 0.7465 / 0.405
%     &22.71 / {\color{blue}0.7487 / 0.387}
%     & \color{red}{23.44} / \color{red}{0.7666} / \color{red}{0.360} 
        $\sigma=50$

& 28.76 / 0.8950 / 0.163
    & {\color{blue}29.20 / 0.9030 / 0.158}
    & \color{red}{29.90} / \color{red}{0.9157} / \color{red}{0.114} 
    \\
    $\sigma=100$

    & 26.12 / 0.8395 / 0.261
    & {\color{blue}26.40 / 0.8484 / 0.249}
    & \color{red}{27.00} / \color{red}{0.8626} / \color{red}{0.201}
    
    \\
    
    $\sigma=150$

	& 24.70 / 0.8036 / 0.316
    & {\color{blue}24.77 / 0.8085 / 0.308}
    & \color{red}{25.39} / \color{red}{0.8236} / \color{red}{0.263}  
    \\
    
    $\sigma=200$

    & {\color{blue}23.66} / 0.7737 / 0.362
    &23.61 / {\color{blue}0.7763 / 0.352}
    & \color{red}{24.25} / \color{red}{0.7915} / \color{red}{0.295} 

    \\
    $\sigma=250$

    & {\color{blue}22.81} / 0.7465 / 0.405
    &22.71 / {\color{blue}0.7487 / 0.387}
    & \color{red}{23.44} / \color{red}{0.7675} / \color{red}{0.346} 
    
    \\
    \bottomrule %[0.1em]
    
    \end{tabular}%

}

\end{table}

    \begin{figure}[t]
    \begin{center}
    \includegraphics[width=1\linewidth]{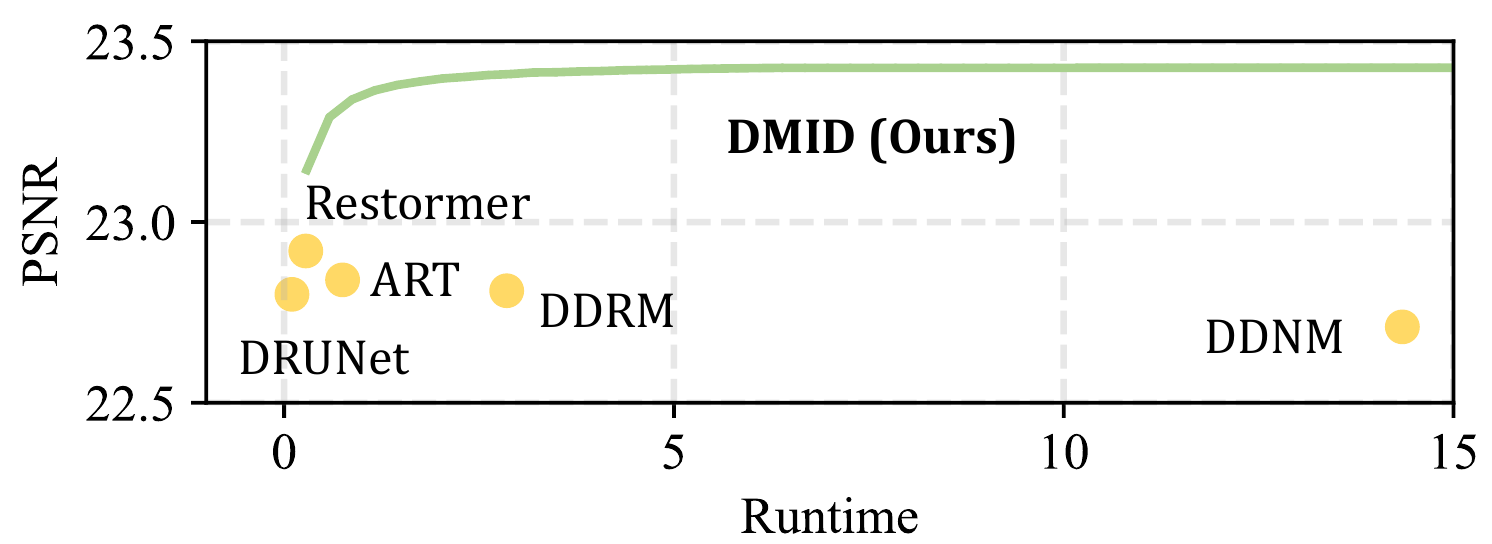}
    \end{center}
    \vspace{-1.5em}
    \caption{The comparison of runtime on the ImageNet dataset with noise level $\sigma=250$. Notably, our method consistently yields optimal results under the same runtime. }
    \label{fig:runtime}
\end{figure}

\begin{figure*}[!t]
\footnotesize
\renewcommand{\arraystretch}{1.2}

% \vspace{2em}
\begin{center}
\setlength{\tabcolsep}{1.5pt}
\scalebox{0.97}{
\begin{tabular}[b]{c c c c c c c c}

\multirow{3}{*}{\includegraphics[trim={0 0 0 0},clip,width=.22\textwidth,valign=t]{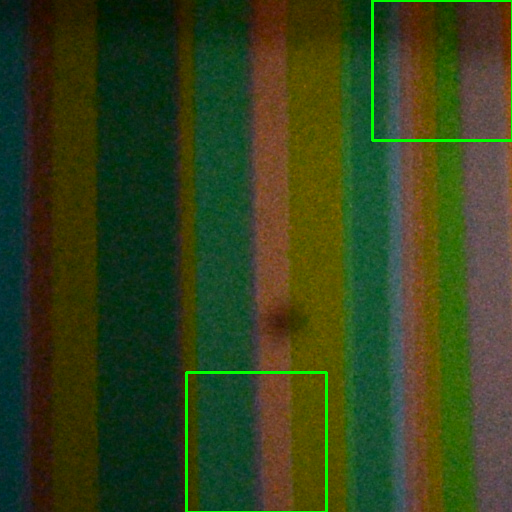}}&
 
 \includegraphics[trim={372 372 0 0},clip,width=.105\textwidth,valign=t]{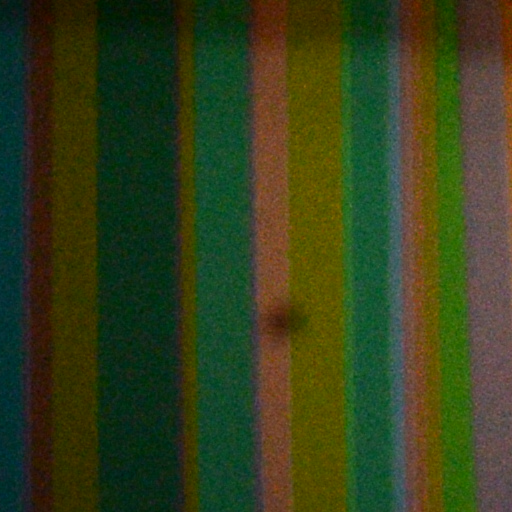} & 
  \includegraphics[trim={372 372 0 0},clip,width=.105\textwidth,valign=t]{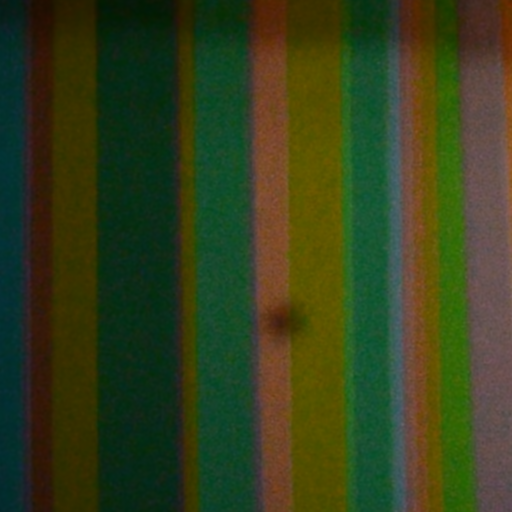} & 
\includegraphics[trim={372 372 0 0},clip,width=.105\textwidth,valign=t]{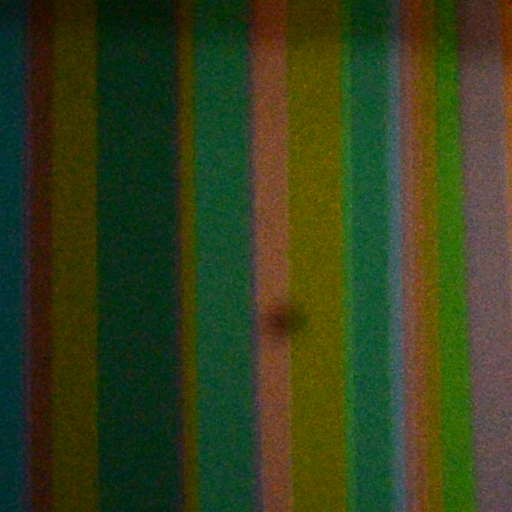} & 
\includegraphics[trim={372 372 0 0},clip,width=.105\textwidth,valign=t]{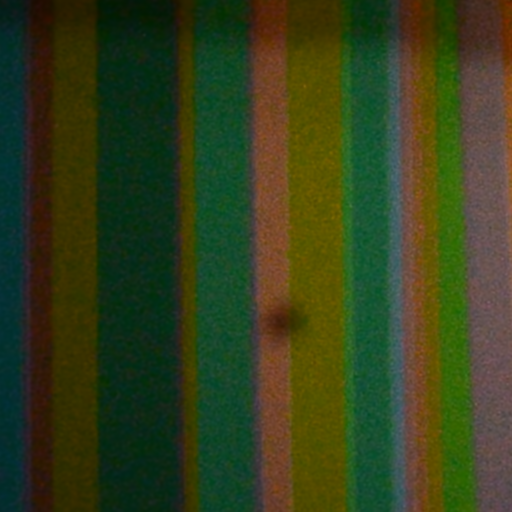} & 
\includegraphics[trim={372 372 0 0},clip,width=.105\textwidth,valign=t]{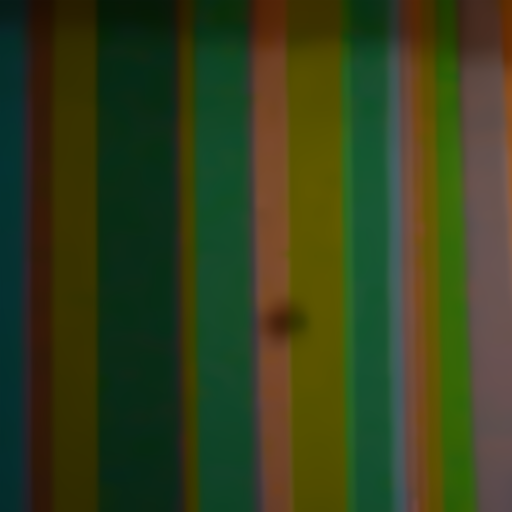}
&
\includegraphics[trim={372 372 0 0},clip,width=.105\textwidth,valign=t]{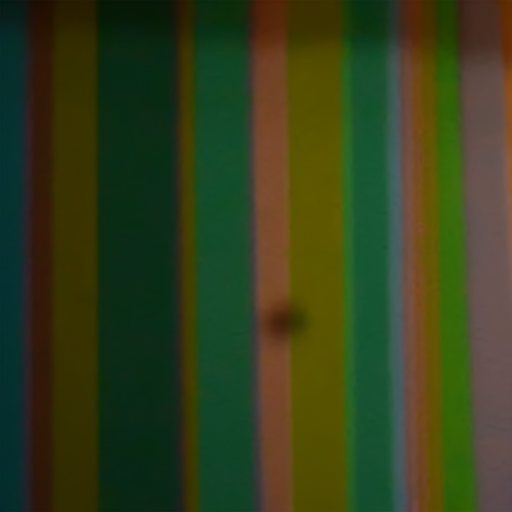}
&
\includegraphics[trim={372 372 0 0},clip,width=.105\textwidth,valign=t]{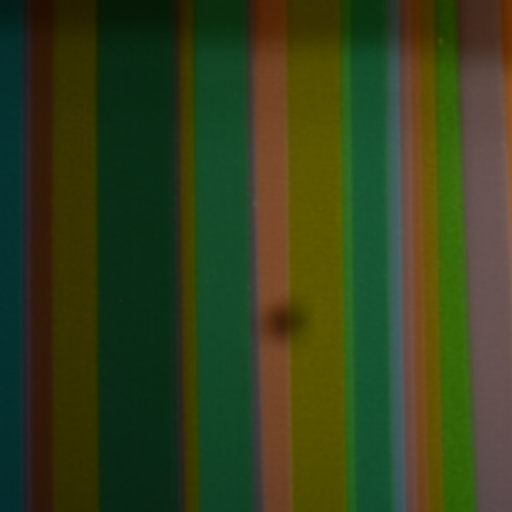}
\\
\addlinespace[2pt]
%& DDRM%~\cite{kawar2022denoising}
%& DDNM% ~\cite{wang2022zero}
%& \textbf{DMID-d} 
%& Clean
%\\
&
 \includegraphics[trim={186 0 186 372 },clip,width=.105\textwidth,valign=t]{Images/diffusion/cc3/DDRM_est.png} & 
  \includegraphics[trim={186 0 186 372 },clip,width=.105\textwidth,valign=t]{Images/diffusion/cc3/DDRM_exact.png} & 
\includegraphics[trim={186 0 186 372 },clip,width=.105\textwidth,valign=t]{Images/diffusion/cc3/DDNM_est.png} & 
\includegraphics[trim={186 0 186 372 },clip,width=.105\textwidth,valign=t]{Images/diffusion/cc3/DDNM_exact.png} & 
\includegraphics[trim={186 0 186 372 },clip,width=.105\textwidth,valign=t]{Images/diffusion/cc3/DMID-d.png}
&
\includegraphics[trim={186 0 186 372 },clip,width=.105\textwidth,valign=t]{Images/diffusion/cc3/DMID-p.png}
&

\includegraphics[trim={186 0 186 372 },clip,width=.105\textwidth,valign=t]{Images/diffusion/cc3/clean.png}
\\
% 32.91 / 0.4138 & 
% 33.01 / 0.4090 & 
% 37.96 / 0.0561 & 
% 33.01 / 0.4075 & 
% 39.09 / 0.1027
32.91 / 0.4138 & 
33.01 / 0.4090 & 
37.31 / 0.2011 & 
33.01 / 0.4075 & 
35.70 / 0.2865
% & {\color{red}41.84 / }{\color{blue}0.0089}
% & {\color{blue}41.75 / }{\color{red}0.0085}
& {\color{red}42.77 / }{\color{blue}0.0265}
& {\color{blue}41.74 / }{\color{red}0.0240}
& PSNR↑ / LPIPS↓ 

\\
Noisy 
& DDRM%~\cite{kawar2022denoising}
& DDRM+Clean
& DDNM% ~\cite{wang2022zero}
% & \textbf{DMID-d (Ours)} 
& DDNM+Clean
& \textbf{DMID-d} 
& \textbf{DMID-p} 
& Clean

\\

%第4张
%left bottom right top
\multirow{3}{*}{\includegraphics[trim={0 0 0 0},clip,width=.22\textwidth,valign=t]{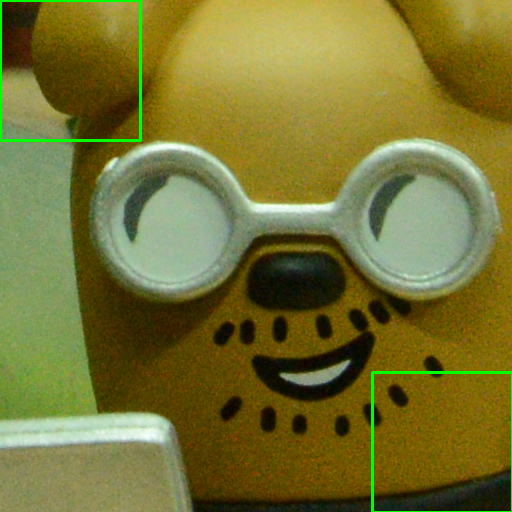}}&
 
 \includegraphics[trim={0 372 372 0},clip,width=.105\textwidth,valign=t]{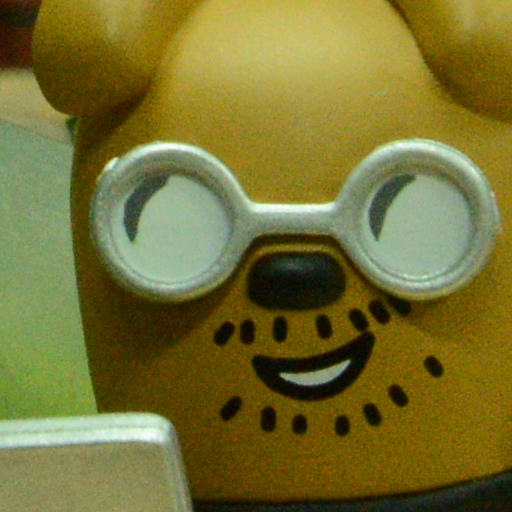} & 
  \includegraphics[trim={0 372 372 0},clip,width=.105\textwidth,valign=t]{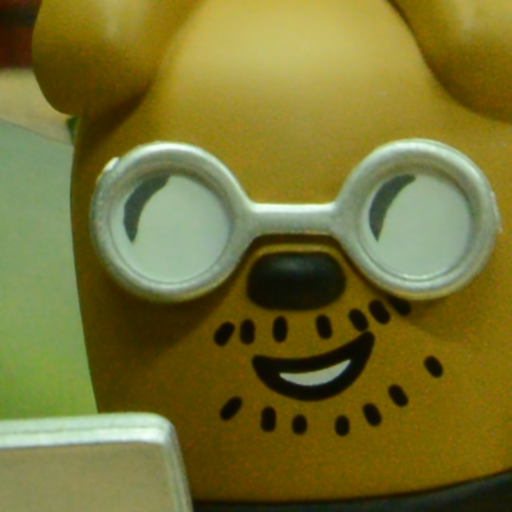} & 
\includegraphics[trim={0 372 372 0},clip,width=.105\textwidth,valign=t]{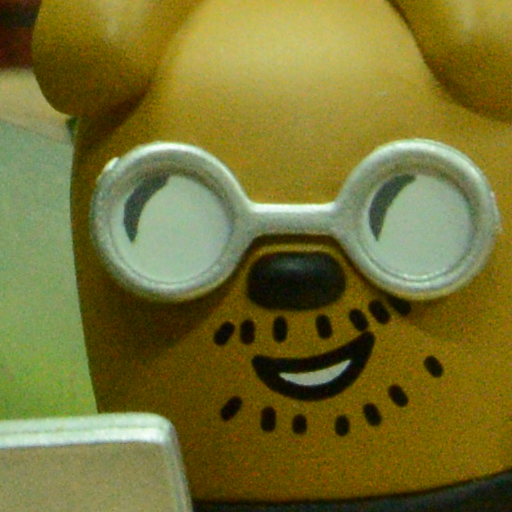} & 
\includegraphics[trim={0 372 372 0},clip,width=.105\textwidth,valign=t]{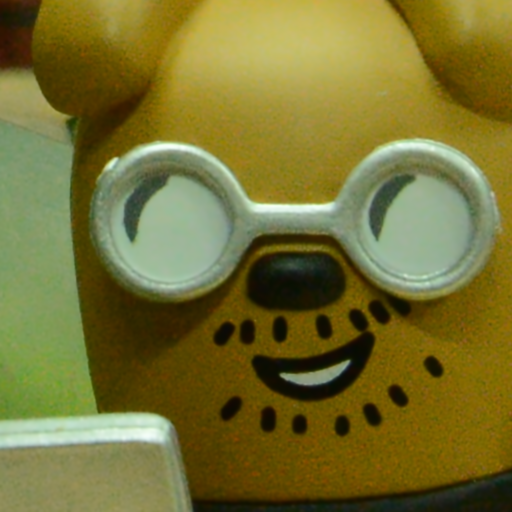} & 
\includegraphics[trim={0 372 372 0},clip,width=.105\textwidth,valign=t]{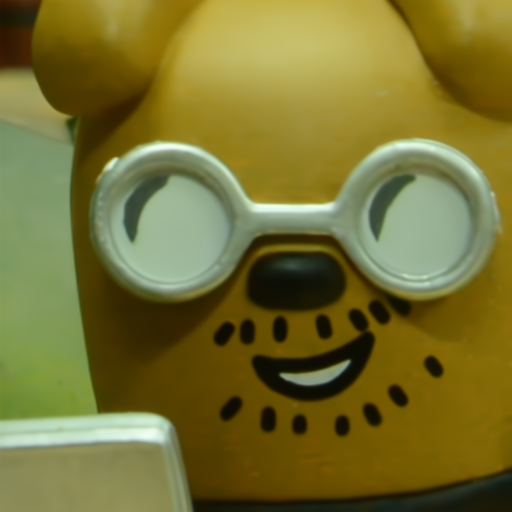}
&
\includegraphics[trim={0 372 372 0},clip,width=.105\textwidth,valign=t]{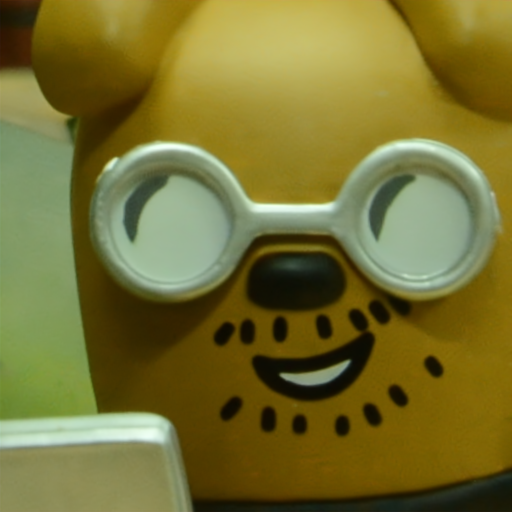}
&
\includegraphics[trim={0 372 372 0},clip,width=.105\textwidth,valign=t]{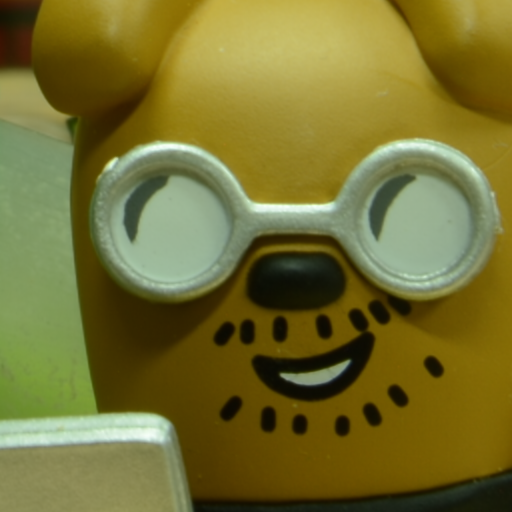}
\\
%& DDRM%~\cite{kawar2022denoising}
%& DDNM% ~\cite{wang2022zero}
%& \textbf{DMID-d} 
%& Clean
%\\
\addlinespace[2pt]
&
%left bottom right top
 \includegraphics[trim={372 0 0 372},clip,width=.105\textwidth,valign=t]{Images/diffusion/cc2/DDRM_est.png} & 
  \includegraphics[trim={372 0 0 372},clip,width=.105\textwidth,valign=t]{Images/diffusion/cc2/DDRM_exact.png} & 
\includegraphics[trim={372 0 0 372},clip,width=.105\textwidth,valign=t]{Images/diffusion/cc2/DDNM_est.png} & 
\includegraphics[trim={372 0 0 372},clip,width=.105\textwidth,valign=t]{Images/diffusion/cc2/DDNM_exact.png} & 
\includegraphics[trim={372 0 0 372},clip,width=.105\textwidth,valign=t]{Images/diffusion/cc2/DMID-d.png}
&
\includegraphics[trim={372 0 0 372},clip,width=.105\textwidth,valign=t]{Images/diffusion/cc2/DMID-p.png}
&

\includegraphics[trim={372 0 0 372},clip,width=.105\textwidth,valign=t]{Images/diffusion/cc2/clean.png}
\\
% 33.26 / 0.2976 & 
% 33.38 / 0.2918 & 
% 37.27 / 0.1000 & 
% 33.38 / 0.2923 & 
% 35.98 / 0.1648
33.26 / 0.2976 & 
33.38 / 0.2918 & 
37.27 / 0.1000 & 
33.38 / 0.2923 & 
35.98 / 0.1648
% & {\color{red}40.38 / }{\color{blue}0.0494}
% & {\color{blue}40.22 / }{\color{red}0.0489}
& {\color{red}40.32 / }{\color{blue}0.0465}
& {\color{blue}39.31 / }{\color{red}0.0342}
& PSNR↑ / LPIPS↓ 
\\
Noisy 
& DDRM%~\cite{kawar2022denoising}
& DDRM+Clean
& DDNM% ~\cite{wang2022zero}
% & \textbf{DMID-d (Ours)} 
& DDNM+Clean
& \textbf{DMID-d} 
& \textbf{DMID-p} 
& Clean

\\
\end{tabular}}
\end{center}
\vspace{-1em}

\caption{Visual results compared with other diffusion-baesd methods. 
Other methods can not deal with real-world noise. 
%The top row other
% In the top row, DDRM and DDNM denoise excessively, leading to over-smooth detail. In the bottom row, DDRM and DDNM denoise incompletely, leading to leftover noise.
% (Best viewed with zoom-in.)
}
\vspace{-1em}
\label{fig:diffusion-compared}
%\vspace{-0.5em}
\end{figure*}

    \begin{figure}[t]
\centering
 \includegraphics[width=0.48\textwidth]{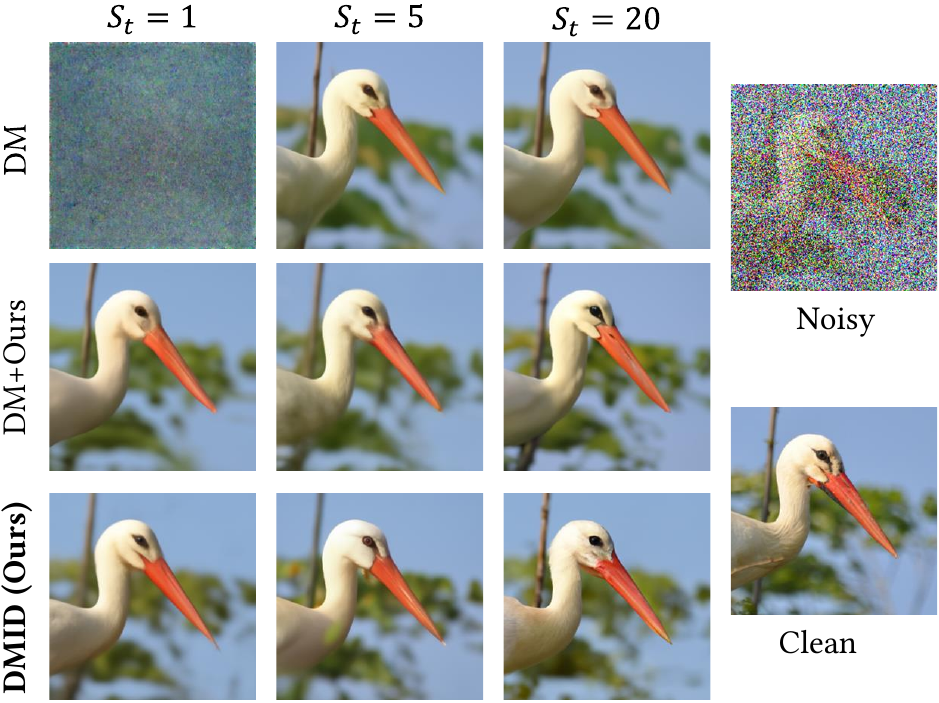}
 \caption{Visual results of the diffusion-based method and extension with our methods.}
 \label{fig:extend}
% \vspace{-1em}
\end{figure}

    Firstly, we conduct experiments compared with other diffusion-based image restoration methods. DDRM~\cite{kawar2022denoising} and DDNM~\cite{wang2022zero} are the representative diffusion-based image restoration methods. Tables \ref{table:4-diffusion} and \ref{table:9-compare and extend} showcase the results of Gaussian denoising on ImageNet and real-world denoising on CC~\cite{2016A}, PolyU~\cite{xu2018real}, and FMDD~\cite{zhang2018poisson}.
    Our method consistently yields optimal results under the same runtime as shown in Figure \ref{fig:runtime}.
    As DDRM~\cite{kawar2022denoising} and DDNM~\cite{wang2022zero} do not focus on image denoising, we do not compare them in the previous Section \ref{sec:Gaussian denoising} and Section \ref{sec:realworld denoising}, and the results in the table are produced by ourselves. The hyperparameters of noise level used in DDRM~\cite{kawar2022denoising} and DDNM~\cite{wang2022zero} are estimated using the method \cite{chen2015efficient}. Other hyperparameters are employed their recommended settings. 
    % Due to their poor performance on real-world denoising, we upgrade DDRM and DDNM to “DDRM+Clean” and “DDNM+Clean”, respectively. Here, for “DDRM+Clean” and “DDNM+Clean”, we introduce the clean image as auxiliary information to calculate the noise level for the corresponding noisy image. 
    Due to their poor performance on real-world denoising, we introduce the clean image as auxiliary information to calculate the noise level and upgrade DDRM and DDNM to “DDRM+Clean” and “DDNM+Clean”, respectively. 
    %Here, for “DDRM+Clean” and “DDNM+Clean”, we introduce the clean image as auxiliary information to calculate the noise level for the corresponding noisy image. 
    However, the performance is still limited.

    As depicted in Figure \ref{fig:diffusion-compared}, both DDRM~\cite{kawar2022denoising} and DDNM~\cite{wang2022zero} struggle to handle image denoising tasks. That is because these methods are not well-suited for image denoising tasks. As discussed in Section \ref{sec:diffusion model}, they can not circumvent the input inconsistency and overlook the content inconsistency. 
    % In contrast, our method is designed to tackle the two problems. 
    % In addition, our DMID method represents a general strategy for denoising within diffusion models, without imposing restrictions on the sampling strategy, as explained in Section \ref{sec:overview}. 
    % %And the sampling strategy can be arbitrary. 
    % Thus, our method can extend to other strategies and help them avoid input inconsistency and content inconsistency.

    Next, we will extend our method to DDRM~\cite{kawar2022denoising} and DDNM~\cite{wang2022zero}. The sampling strategy in our method can be arbitrary. However, practical applications require some adjustments when applying different sampling strategies. This is because various sampling methods have unique designs and cannot be directly integrated. Both DDRM~\cite{kawar2022denoising} and DDNM~\cite{wang2022zero} utilize the same sampling strategy for noise handling, with the primary difference being that DDRM~\cite{kawar2022denoising} involves a specific initialization step. This initialization, however, conflicts with our embedding method and is therefore replaced. Thus, DDRM~\cite{kawar2022denoising} and DDNM~\cite{wang2022zero} are the same for our method, and we denote them as DM.

    The results are presented in Table \ref{table:9-compare and extend}. 
    %DDRM and DDNM are the same as discussed in Table \ref{table:9-compare and extend}.  	
	For the embedding method, we apply the same noise transformation as in our method. In addition, we convert the image to the intermediate state $x_N$ and start sampling from $x_N$. 
	For the ensembling method, we set the sampling times to be $S_t=3$ for CC~\cite{2016A} and PolyU~\cite{xu2018real}, and $S_t=11$ for FMDD~\cite{zhang2018poisson}.

	The visual results are presented in Figure \ref{fig:extend}. 
    The versions featuring our embedding method and ensembling method (referred to "DM +Ours"), outperform their original counterparts DDRM~\cite{kawar2022denoising} and DDNM~\cite{wang2022zero} easily. 
    Additionally, "DM +Ours" employs significantly fewer sampling times compared to DDRM~\cite{kawar2022denoising} and DDNM~\cite{wang2022zero}, respectively. 
    For example, the sampling times $S_t$ are 3 for "DM +Ours", whereas DDRM~\cite{kawar2022denoising} and DDNM~\cite{wang2022zero} require 20 and 100 sampling times, respectively, on the PolyU~\cite{xu2018real} dataset. 
    % {\color{red} Specifically, the denoising runtime for an single image of $512 \times 512 \times 1$ from FMDD dataset is present in Table \ref{table:9_time}.} 
    DDRM and DDNM require more runtime because DDRM~\cite{kawar2022denoising} and DDNM~\cite{wang2022zero} start sampling from {
    % \color{red} 
    standard} Gaussian noise for image denoising. 
    {
    % \color{red}
    Moreover, when DDRM and DDNM produce the intermediate image $x_N$ with a noise level comparable to that of the original input noisy image $y$, the content and information of $x_N$ do not align with those of the original input noisy image $y$.} 
    Consequently, {
    % \color{red}
    DDRM~\cite{kawar2022denoising} and DDNM~\cite{wang2022zero} } not only exhibit poor performance but also demand a larger number of sampling times. 
    This experiment further emphasizes the superiority, effectiveness, and scalability of our method.
    
    \section{Discussion}
    \label{sec:final_discussion}
    %%%%%%%%%%%%%%%%%%%%%%%%%%%%%%%%%%%%%%%%%%%%%%%%%%%%%%%%%%
    % In this paper, we propose a method to employ the unconditional diffusion model pre-trained for image synthesis. 
    % Currently, there is no paradigm that effectively stimulates the diffusion model for image denoising. 
    % The pros and cons of trainable and pre-trained solutions present a very interesting question that has not been discussed before. In the following paragraphs, we will delve into the advantages and disadvantages of the two various approaches and strive to arrive at a conclusion.
    In this paper, we introduce a method that utilizes the diffusion model pre-trained for image synthesis. At present, there is no established paradigm for effectively leveraging the diffusion model for image denoising. The comparison between trainable and pre-trained solutions raises an intriguing question that has yet to be thoroughly explored. In the subsequent paragraphs, we will analyze the merits and drawbacks of both approaches and aim to draw a conclusion.

    The pre-trained solution employs the same pre-trained diffusion model for universal image denoising. Utilizing pre-trained diffusion models eliminates the need for training and provides extensive image priors, enhancing denoising capabilities. Despite these advantages, challenges related to input inconsistency and content inconsistency are required to be corrected.
    
    % The trainable solution trains customized diffusion models for different noise types and datasets, highly targeted but not efficient. The original diffusion models add Gaussian noise during the sampling process. For the customized diffusion model tailored to a specific noise type, the sampling process will include that particular noise. Thus, the trainable solution enables better handling in processing the same type of noise as the training noise. However, complexities arise from the unknown distribution of real noise and the sensor-specific nature of noise, leading to potential inefficiencies in training. Additionally, diffusion models are delicate and achieve optimal results only when Gaussian noise is added in the sampling process. Changing the noise distribution added in the sampling process leads to suboptimal performance.
    The trainable solution customizes diffusion models for various noise types and datasets~\cite{xie2023diffusion}, improving the handling of specific noise. However, complexities arise from the unknown distribution of real noise and the sensor-specific nature of noise~\cite{PMN}, leading to improbability in customizing specific models for real noise and inefficiencies in training different models for various noise. In addition, diffusion models are delicate and only excel with Gaussian noise~\cite{bansal2024cold,jolicoeur2023diffusion}. Customizing diffusion models and altering the noise distribution leads to suboptimal performance~\cite{bansal2024cold,jolicoeur2023diffusion}.

    In summary, our solution which employs pre-trained diffusion models offers a flexible solution for handling various types of noise. However, some modifications are required to adapt the pre-trained diffusion model. On the other hand, trainable diffusion models provide a solution for customizing the training of different diffusion models based on distinct noise profiles. Nevertheless, challenges arise in customizing the training process and optimizing the denoising capabilities.
    
	%%%%%%%%%%%%%%%%%%%%%%%%%%%%%%%%%%%%%%%%%%%%%%%%%%%%%%%%%%
	\section{Conclusion}
	
	\label{sec:conclusion}
	In this paper, we present a novel strategy to stimulate the diffusion model for image denoising. Specifically, we revisit diffusion models from the denoising perspective. Furthermore, we propose an adaptive embedding method to perform denoising and an adaptive ensembling method to reduce distortion. Our method achieves SOTA performance on both distortion-based and perception-based metrics, for both Gaussian and real-world image denoising. 
    %And we solely rely on a diffusion model pre-trained for image synthesis, without any additional supervised training.
    % In future research endeavors, we intend to broaden the applicability and further enhance the effectiveness of our method.
    In future research endeavors, we intend to stimulate the diffusion model for multiple other restoration tasks.

    %  \section*{Acknowledgments}
    % The authors would like to thank the reviewers for their valuable time and great efforts in reviewing this paper. This work is supported by the National Natural Science Foundation of China under Grants 62322204, 62131003, and 62072038.

\ifCLASSOPTIONcompsoc

\ifCLASSOPTIONcaptionsoff
\newpage
\fi

% \normalem
\bibliographystyle{IEEEtran}
\bibliography{Reference}

\begin{IEEEbiography}[{\includegraphics[width=1in,
height=1.25in,
clip,keepaspectratio]{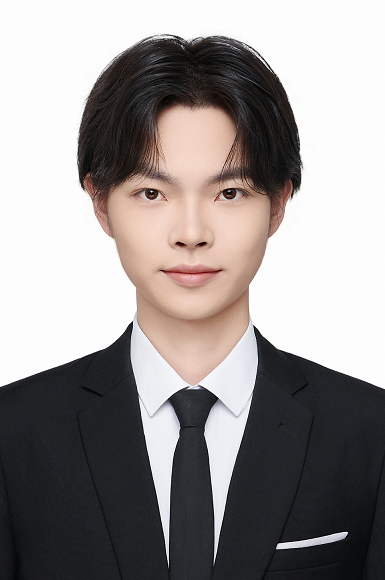}}]{Tong Li}
 received the BS degree from Beijing Institute of Technology, China, in 2023. He is currently a Master student with the School of Computer Science and Technology at Beijing Institute of Technology. His research interests include computational photography and image processing.
\end{IEEEbiography}
\vspace*{-1.25cm}
\begin{IEEEbiography}[{\includegraphics[width=1in,
height=1.25in,
clip,keepaspectratio]{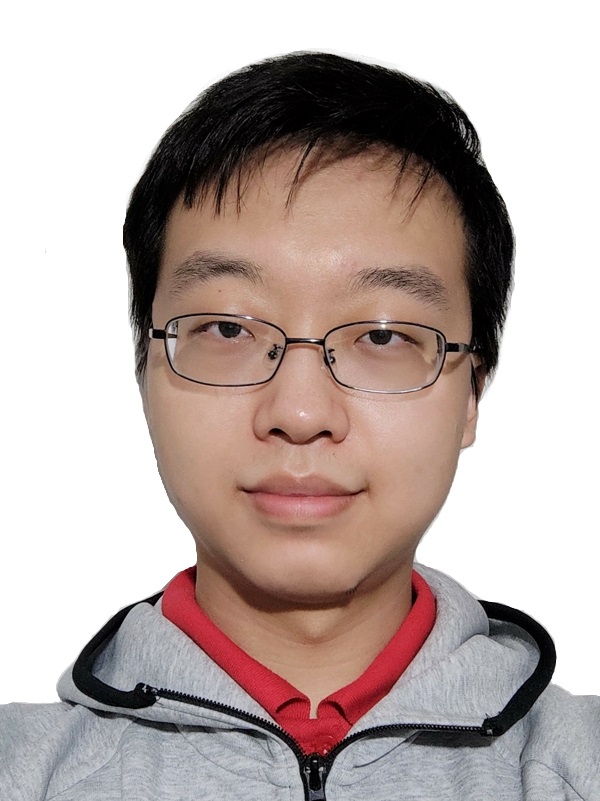}}]{Hansen Feng}
received the BS degree from the University of Science and Technology Beijing, China, in 2020. He is currently a Ph.D. student with the School of Computer Science and Technology, Beijing Institute of Technology. His research interests include computational photography and image processing. He received the Best Paper Runner-Up Award of ACM MM 2022.
\end{IEEEbiography}
\vspace*{-1.25cm}

\begin{IEEEbiography}[{\includegraphics[width=1in,
height=1.25in,
clip,keepaspectratio]{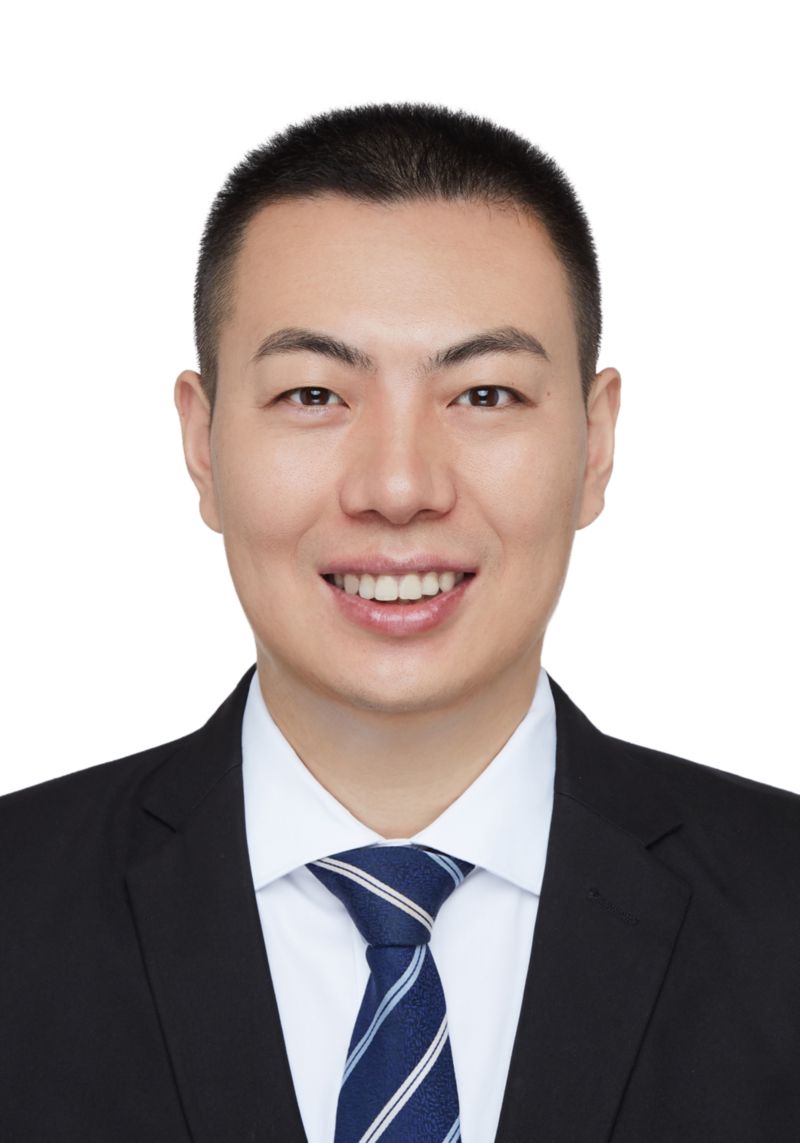}}]{Lizhi Wang}
%(M’17) 
(Member, IEEE))
received the B.S. and Ph.D. degrees from Xidian University, Xi’an, China, in 2011 and 2016, respectively. He is currently a professor with the School of Artificial Intelligence, Beijing Normal University. His research interests include computational photography and image processing. He is serving as an associate editor of IEEE Transactions on Image Processing. He received the Best Paper Runner-up Award of ACM MM2022and Best Paper Award of IEEE VCIP 2016.
\end{IEEEbiography}
\vspace*{-1.25cm}

\begin{IEEEbiography}[{\includegraphics[width=1in,height=1.25in,clip,keepaspectratio]{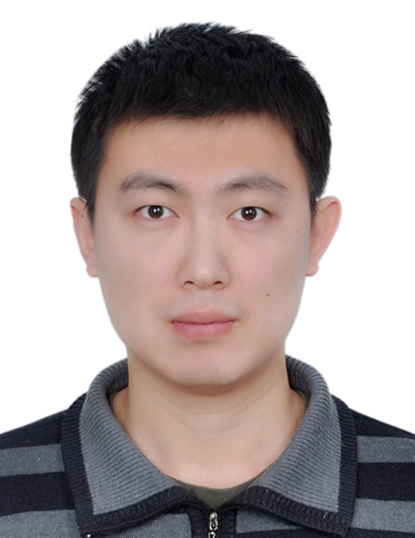}}] {Lin Zhu}
(Member, IEEE)
received the B.S. degree in computer science from the Northwestern Polytechnical University, China, in 2014, the M.S. degree in computer science from the North Automatic Control Technology Institute, China, in 2018, and the Ph.D. degree with the School of Electronics Engineering and Computer Science, Peking University, China, in 2022. He is currently an assistant professor with the School of Computer Science, Beijing Institute of Technology, China. His current research interests include image processing, computer vision, neuromorphic computing, and spiking neural network. 
\end{IEEEbiography}

\vspace*{-1.25cm}
\begin{IEEEbiography}
[{\includegraphics[width=1in,
height=1.25in,
clip,keepaspectratio]{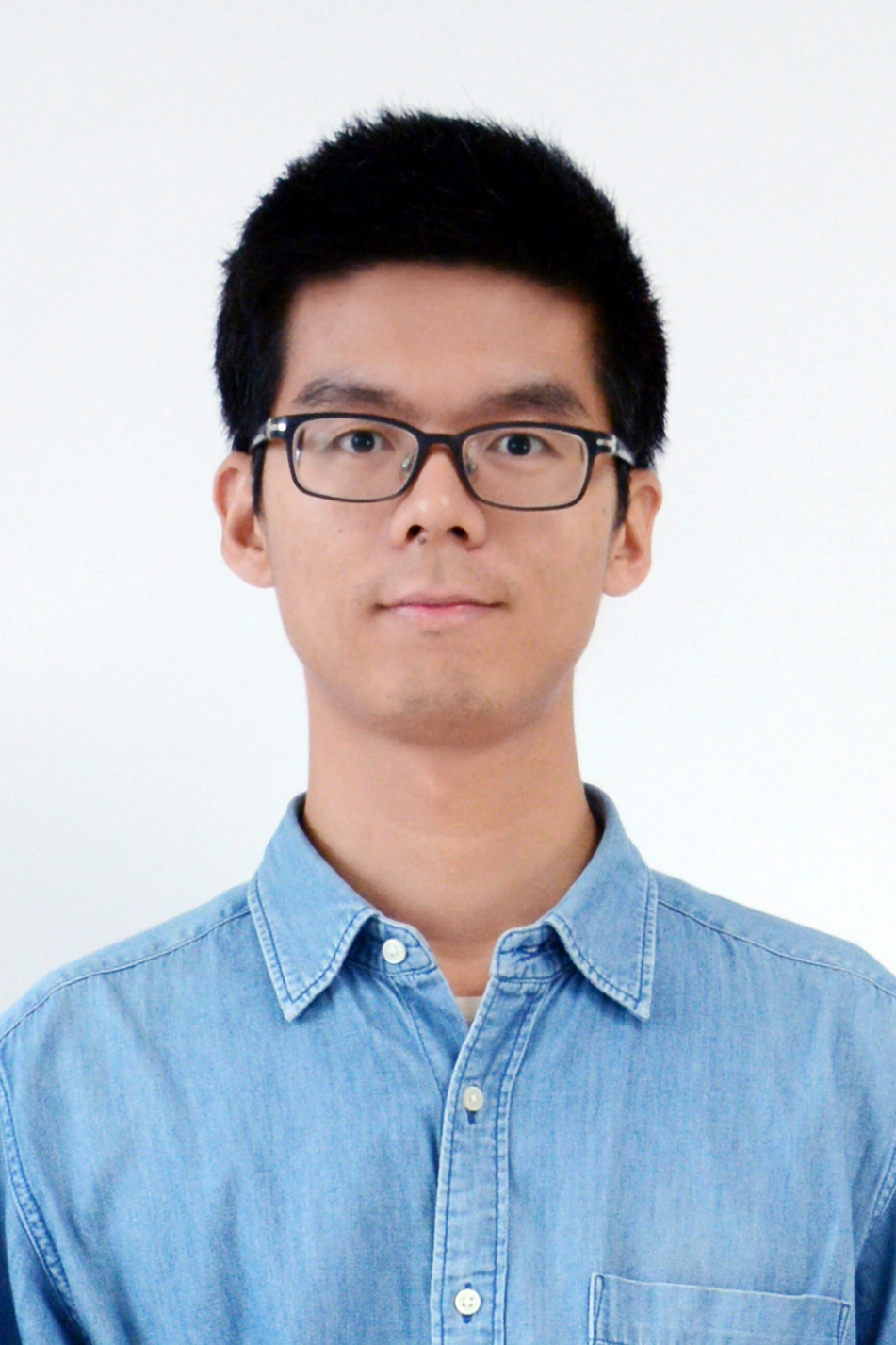}}]{Zhiwei Xiong}
(Member, IEEE)
received the BS and
PhD degrees in electronic engineering from the University of Science and Technology of China (USTC), in 2006 and 2011, respectively. He is currently a professor with USTC. Before that, he was a researcher with Microsoft Research Asia (MSRA).
His research interests include computational photography, 3D vision, and biomedical image analysis.
He has authored or coauthored more than 100
papers in premium journals and conferences. He
received the Best Paper Award of IEEE VCIP 2016
and MSRA Fellowship 2009. He and his students were winners of 8 technical
challenges held in CVPR / ICCV / ECCV / MM / ICME / ISBI.
\end{IEEEbiography}
 \vspace*{-1.25cm}
\begin{IEEEbiography}
[{\includegraphics[width=1in,
height=1.25in,
clip,keepaspectratio]{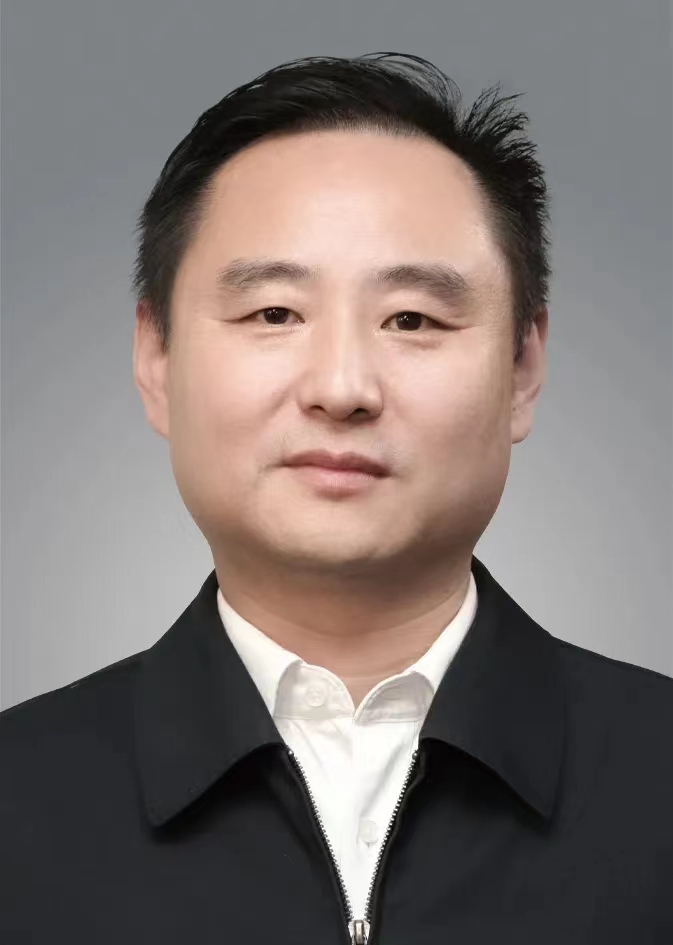}}]{Hua Huang}
(Senior Member, IEEE)
%(SM’19) 
received the B.S. and Ph.D.	degrees from Xi’an Jiaotong University, in 1996	and 2006, respectively. He is currently a professor in the School of Artificial Intelligence, Beijing Normal University. He is also an adjunct professor at Xi’an Jiaotong University and Beijing Institute of Technology. His main research interests include image and video processing, computational photography, and computer graphics. He received the Best Paper Award of ICML2020 / EURASIP2020 / PRCV2019 / ChinaMM2017.
\end{IEEEbiography}
\end{document}